\documentclass{article} 
\usepackage{iclr2027_conference,times}

\usepackage{amsmath,amsfonts,bm}

\def\eqref#1{equation~\ref{#1}}

\def\1{\bm{1}}

\DeclareMathAlphabet{\mathsfit}{\encodingdefault}{\sfdefault}{m}{sl}
\SetMathAlphabet{\mathsfit}{bold}{\encodingdefault}{\sfdefault}{bx}{n}

\usepackage{hyperref}
\usepackage{url}
\usepackage{mathtools}

\usepackage{amssymb}
\usepackage{mathrsfs}
\usepackage{amsthm}
\usepackage{enumitem}   
\usepackage{graphicx}
\usepackage{wrapfig}    
\usepackage{lipsum}     
\usepackage{float}
\usepackage{booktabs}
\usepackage{amsmath}
\usepackage{amsfonts}
\usepackage{nicefrac}
\usepackage{microtype}
\usepackage{xspace}
\usepackage{multirow}
\usepackage{cleveref}
\usepackage{etoc}       
\usepackage{times}
\usepackage{epsfig}
\usepackage{adjustbox}
\usepackage{xcolor,colortbl}
\usepackage{cuted}
\usepackage{bm}

\newcommand{\eg}{\emph{e.g.}\xspace}

\newcommand{\methodname}{Gen4P\xspace}

\newcommand{\heading}[1]{\vspace{2pt}\noindent\textbf{#1.}}

\usepackage{xcolor}
\usepackage{caption}
\usepackage{fontawesome5}
\usepackage{pgfplots}
\pgfplotsset{compat=1.18}
\usepackage{subcaption}
\usepackage{tabularx}
\usepackage{wrapfig}
\crefname{figure}{Fig.}{Figs.}
\Crefname{figure}{Fig.}{Figs.}

\crefname{table}{Tab.}{Tabs.}
\Crefname{table}{Tab.}{Tabs.}

\crefname{section}{Sec.}{Secs.}
\Crefname{section}{Sec.}{Secs.}

\crefname{subsection}{Sec.}{Secs.}
\Crefname{subsection}{Sec.}{Secs.}

\crefname{equation}{Eq.}{Eqs.}
\Crefname{equation}{Eq.}{Eqs.}

\crefname{algorithm}{Alg.}{Algs.}
\Crefname{algorithm}{Alg.}{Algs.}

\crefname{appendix}{App.}{Apps.}
\Crefname{appendix}{App.}{Apps.}

\newcommand{\frozen}{\ifmmode\text{\faIcon{snowflake}}\else\faIcon{snowflake}\fi}

\newcommand{\tsrc}{t_{\mathrm{src}}}
\newcommand{\ttgt}{t_{\mathrm{tgt}}}
\newcommand{\tcam}{t_{\mathrm{cam}}}

\definecolor{bestblue}{RGB}{65, 145, 200}
\definecolor{secondblue}{RGB}{125, 185, 225}
\definecolor{thirdblue}{RGB}{185, 225, 250}

\definecolor{plotblue}{RGB}{70, 140, 195}

\newcommand{\besthl}[1]{%
    \begingroup
    \setlength{\fboxsep}{1pt}%
    \colorbox{bestblue}{\textbf{#1}}%
    \endgroup
}

\newcommand{\secondbesthl}[1]{%
    \begingroup
    \setlength{\fboxsep}{1pt}%
    \colorbox{secondblue}{#1}%
    \endgroup
}

\newcommand{\thirdbesthl}[1]{%
    \begingroup
    \setlength{\fboxsep}{1pt}%
    \colorbox{thirdblue}{#1}%
    \endgroup
}

\newcommand\rurl[1]{%
  \href{https://#1}{\nolinkurl{#1}}%
}

\title{\methodname: Video Generators Are 4D Perceivers}

\iclrfinalcopy
\author{%
Yitong Xia$^{1}$ \quad
Zeren Jiang$^{2}$ \quad
\textbf{Chuanxia Zheng}$^{1}$\\[0.5em]
\normalfont $^{1}$PVG, Nanyang Technological University \quad $^{2}$VGG, University of Oxford \quad
\\[0.3em]
}

\begin{document}

\maketitle

\vspace{-15pt}

\begin{figure*}[htbp]
    \centering
    \includegraphics[width=\linewidth]{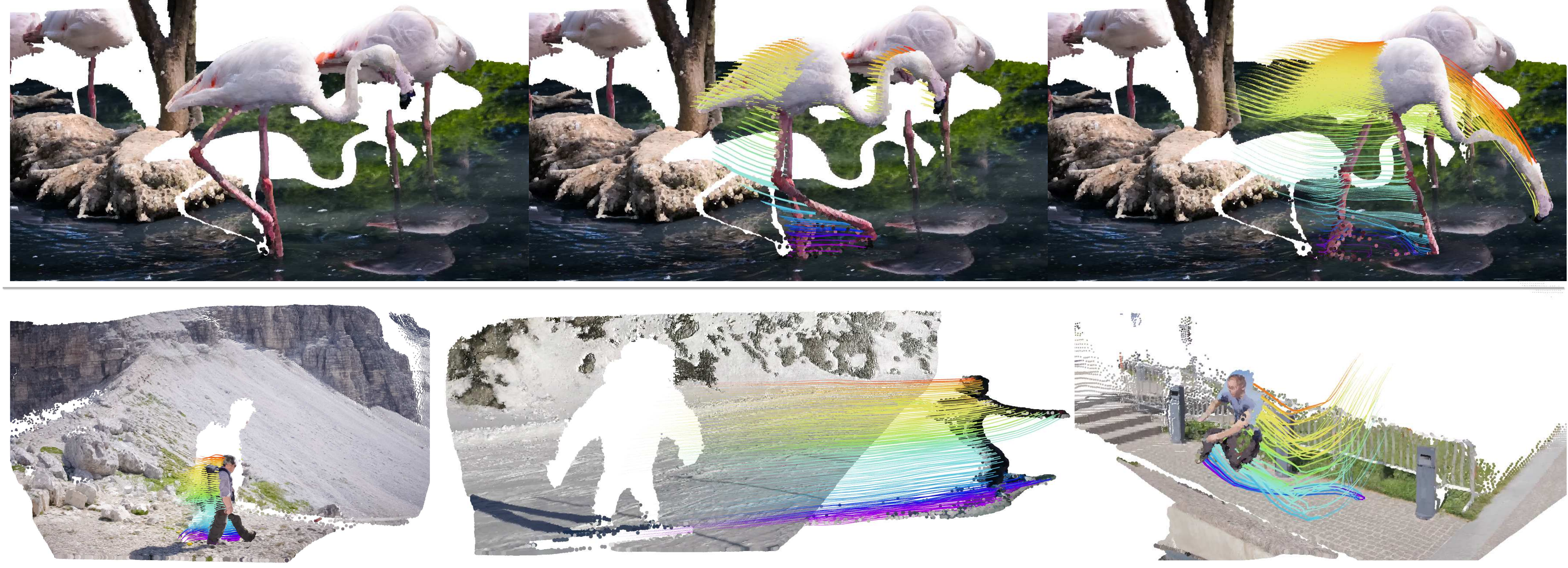}
    \caption{
        \textbf{\methodname} repurposes a multi-step Video \textbf{\underline{Gen}}erator into a single-step \textbf{\underline 4}D \textbf{\underline P}erceiver,
        enabling unified feed-forward prediction of dense 3D tracks, point maps, depth maps, and camera poses from a video input.
        \methodname is able to track complex motion while maintaining the delicate structure of in-the-wild content, like the thin legs of the flamingo and long-range skiing motion.
    }
    \label{fig:teaser}
\end{figure*}

\etocdepthtag.toc{mainpaper}

\begin{abstract}


We introduce \textbf{\methodname}, a unified framework for recovering 4D geometry and motion from monocular video.
We show that a pretrained multi-step video generator can be repurposed into a feed-forward 4D perceiver without iterative denoising.
Building on on-demand querying, \methodname processes clean video latents in a single encoder pass and predicts the requested 4D attributes through a lightweight cross-attention decoder.
The on-demand query formulation enables flexible spatial sampling, and our proposed continuous-time queries further provide temporal flexibility.
To strengthen geometric knowledge, we propose a straightforward plug-in solution by incorporating a pretrained video-depth LoRA as a frozen geometry prior, improving overall performance without adding extra trainable parameters.
\methodname achieves state-of-the-art or competitive performance on tasks of 3D point tracking, video depth and camera pose estimation. Qualitative results on in-the-wild videos illustrate its capability of tracking complex motion while maintaining the delicate structure.
Together, these results demonstrate the potential of video generative pretraining for unified 4D perception. 
See \url{https://gen4p.github.io/} for more visual results.

\end{abstract}
\section{Introduction}
\label{sec:intro}

We consider the problem of \emph{feed-forward full 4D reconstruction} from monocular video.
A unified 4D representation of scene geometry, depth, camera poses, and 3D point trajectories provides a foundation for simulating and interacting with the physical environment.
However,
recovering such a representation remains challenging:
it requires accurately capturing both the \emph{static} geometry and the \emph{dynamic} motion of the scene \emph{across viewpoints and over time}.

Recent advances in feed-forward reconstruction demonstrate the value of pretrained foundational vision models.
Geometry Foundation Models (GFMs),
including VGGT~\citep{wang2025vggt}, VGGT-$\Omega$~\citep{wang2026vggt}, and DA3~\citep{lin2025da3}, 
build on pretrained vision models~\citep{oquabdinov2,siméoni2025dinov3},
adapting them through large-scale geometric supervision.
These GFMs then provide strong geometry \emph{priors} for subsequent 4D reconstruction: V-DPM~\citep{sucar2026vdpm} and 4RC~\citep{luo20264rc},
for instance, extend VGGT and DA3 to jointly recover geometry and motion.
However, geometry priors alone do not explicitly model how points should evolve over time in dynamic scenes.

Meanwhile,
foundational generative models offer a complementary source of priors over appearance, geometry, and motion,
by learning to synthesize images and videos~\citep{wan2025,bytedance2026seedance}.
Recent studies have exploited generative priors for
2D correspondence~\citep{hedlin2023unsupervised},
depth estimation~\citep{ke2024repurposing,hu2025depthcrafter},
3D geometry recovery~\citep{zhang2024world,lu2025matrix3d,jiang2025geo4d}, motion tracking~\citep{zhu2026motioncrafter,nam2026trackcraft3r},
and general perception~\citep{wang2026genception}.
However, they still address 4D perception through task-specific approaches.
Besides,
they often require dense, frame-aligned supervision, limiting their flexibility in handling arbitrary spatiotemporal video inputs.

In this work,
we introduce \textbf{\methodname},
a 4D perceiver that leverages rich priors from large-scale pretrained video generators for full 4D reconstruction.
Specifically,
\methodname{} tames a \emph{multi-step} video generator into a \emph{single-step} feed-forward 4D perceiver.
To this end, the objective is formulated as learning a direct \emph{regression} network, instead of the \emph{iterative denoising} for geometric reconstruction in prior works~\citep{ke2024repurposing,jiang2025geo4d,lu2025matrix3d}.
Such a formulation requires \emph{only one} encoder forward pass per input clip. The resulting scene tokens are then shared across queries, reducing the encoding cost over repeated 4D predictions.

We obtain the full 4D attributes by querying the scene tokens using a lightweight cross-attention decoder.
It decouples the output sampling from pre-extracted feature grids,
allowing \emph{spatial flexibility} in the output resolution.
Additionally, we introduce a \emph{continuous-time query} mechanism to handle variable-length video sequences effectively.
Furthermore, we discover that a pretrained video-depth LoRA can further be incorporated as a frozen plug-in \emph{geometric} prior, strengthening geometry without introducing additional trainable parameters.

Our key contributions are summarized as follows:
(i) we repurpose a video generator into a feed-forward 4D perceiver under sparse query supervision, offering an alternative to GFM-based methods;
(ii) we introduce a \emph{continuous-time query mechanism} together with on-demand decoding to provide temporal and spatial flexibility;
(iii) we incorporate a frozen depth LoRA as a \emph{plug-in geometry prior}, improving overall 4D reconstruction without adding trainable parameters; and
(iv) we provide a comprehensive evaluation of our method on a set of 4D reconstruction tasks, complemented by qualitative results on in-the-wild videos, demonstrating its effectiveness and generalization.

\section{Related Work}
\label{sec:related}



\noindent\textbf{Feed-forward GFMs} directly regress pixel-aligned geometry, from pairwise point maps~\citep{wang2024dust3r} to sequence models that use spatial memory~\citep{wang2025spann3r}, recurrent scene states~\citep{wang2025continuous}, or joint multi-view attention with multiple prediction heads~\citep{wang2025vggt}.
Later models remove the reference view~\citep{wang2025pi}, accept optional geometric inputs~\citep{keetha2026mapanything}, or learn unified depth--ray targets~\citep{lin2025da3}.

\noindent\textbf{Point Tracking in 2D and 3D.}
2D point tracking has been extensively studied in both sparse and dense settings~\citep{doersch2022tap,karaev2024cotracker,karaev2025cotracker3,lai2026cowtracker}.
3D point tracking methods lift pixels with monocular depth and iterative refinement~\citep{xiao2024spatialtracker},
improve dense long-range estimation through optimized surrogate queries~\citep{ngo2025delta, ngo2025deltav2}, or jointly estimate geometry, camera, and object motion~\citep{xiao2025spatialtrackerv2}.
GFM-based 4D perceivers represent dynamic 3D scenes through per-frame point maps~\citep{zhang2025monst3r}, dynamic point maps~\citep{sucar2026vdpm}, conditional frame-wise decoding~\citep{luo20264rc}, or motion bases~\citep{lin2026sm4rt}.
Other approaches include common-time point-map pairs~\citep{feng2025st4rtrack}, parametric trajectories~\citep{liu2026trace}, and scene flow~\citep{zhu2026motioncrafter,karhade2026any4d,qian2026flow4r}.
Most rely on frame-aligned grid decoding, coupling output sampling and cost to spatial resolution, whereas D4RT~\citep{zhang2026d4rt} decouples prediction from predefined grids through on-demand queries. 

\noindent\textbf{Generative Priors for Geometry and Motion.}
Large pretrained generators encode geometry and motion priors transferable to perception tasks~\citep{zhan2024does, huang2026vidfm}.
Image adaptations estimate depth through iterative denoising~\citep{ke2024repurposing} or single-step regression~\citep{he2025lotus}.
Video adaptations predict depth~\citep{hu2025depthcrafter}, point maps~\citep{xu2025geometrycrafter}, fused point--ray maps~\citep{jiang2025geo4d}, or scene flow~\citep{zhu2026motioncrafter}.
Feature-based trackers instead exploit internal temporal correspondences for 2D tracking~\citep{nam2025emergenttemporalcorrespondencesvideo,son2025ditracker}.
Recent feed-forward adaptations repurpose video generators for direct video depth regression~\citep{zhang2026dvd}, dense 3D tracking~\citep{nam2026trackcraft3r}, or multimodal unified perception~\citep{wang2026genception}.


\section{Methods}
\label{sec:methods}

\begin{figure*}
    \centering
    \includegraphics[width=\linewidth]{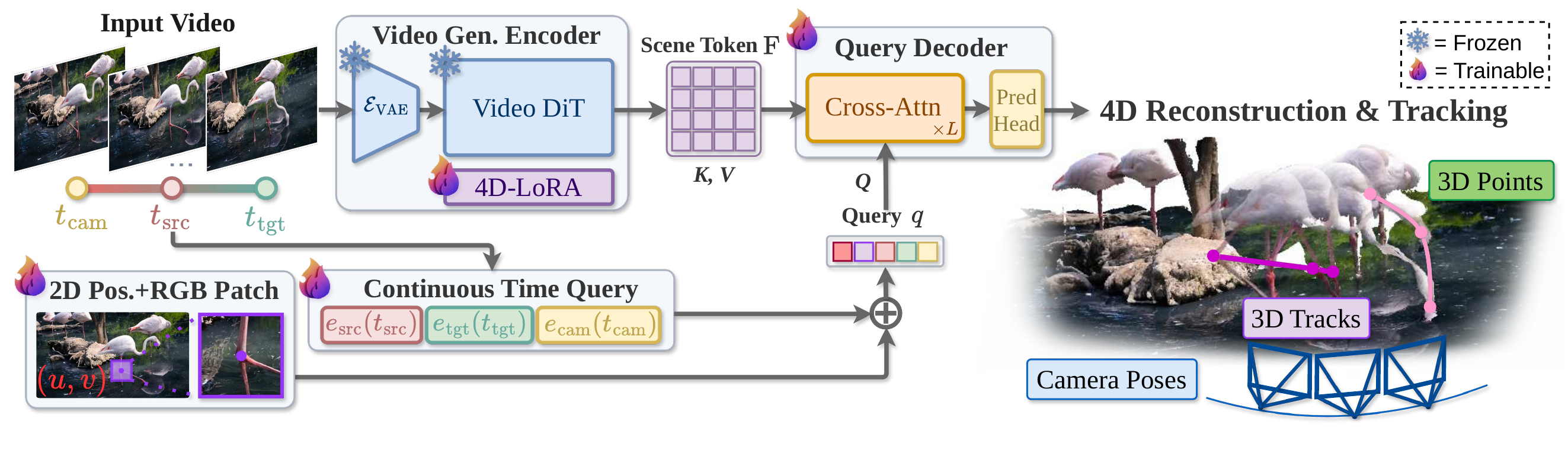}
    \captionsetup{skip=-2pt}
    \caption{
        \textbf{Overview of \methodname.} The input video is encoded into scene tokens $\mathbf{F}$ \emph {in a single pass} by the \textit{Video Generative Encoder}, which consists of a frozen VAE encoder and DiT, an optional frozen Depth-LoRA, and a trainable 4D-LoRA. 
        The cross-attention \textit{Query Decoder} is queried independently for each query $\mathbf{q}(u,v,\tsrc,\ttgt,\tcam)$, where
        the query timestamps are parameterized as continuous time embeddings.
        The decoder outputs the corresponding 3D and 2D point attributes, enabling the full construction of 3D tracks, depth maps and camera poses.
    }
    \label{fig:pipeline}
\vspace{-10pt}
\end{figure*}



Given an input video $\bm{V}=\{\bm{I}_{t}\}_{t=1}^{T}\in\mathbb{R}^{T\times H\times W\times 3}$,
our goal is to learn a neural network
$p_{\bm{\theta}}$
to predict a full set of 4D attributes,
including 3D tracking, 3D point clouds, video depths, and camera parameters.
Instead of training a large-scale 4D geometric network from scratch, in \methodname we repurpose a pretrained video generator $p_{\mathrm{pre}}$ as a feed-forward 4D perceiver.

As illustrated in \Cref{fig:pipeline}, \methodname adopts an \textit{on-demand} query-based formulation inspired by D4RT (\Cref{sec:preliminary}).
A lightweight decoder queries spatiotemporal features from the pretrained video generator (\Cref{sec:encoder}) to predict 4D attributes at specified locations and times.
Since real-world videos vary in duration, a practical 4D reconstruction method should accommodate different sequence lengths. Unlike D4RT, we introduce a \emph{continuous time query} (\Cref{sec:decoder}) to enhance flexibility. We further introduce a \emph{plug-in} geometry prior (\Cref{sec:geometry_prior}) to enhance geometric performance.
%
The training objective is described in \Cref{sec:loss}.
We detail each component below.

\subsection{Preliminaries: D4RT}
\label{sec:preliminary}

D4RT~\citep{zhang2026d4rt} formulates full 4D reconstruction as \textit{on-demand} point querying.
Given a \textit{fixed-length} input video $\bm{V}\in\mathbb{R}^{48 \times H \times W \times 3}$,
a VideoMAE-initialized~\citep{wang2023videomaev2} encoder
$\mathcal{E}$
first encodes it into a global scene representation
$\mathbf{F}$ 
that is shared by all subsequent point queries:
\begin{equation}
    \label{eq:encoder}
    \mathbf{F} = \mathcal{E}(\bm{V}) \in \mathbb{R}^{N_e\times C}.
\end{equation}
Then $\mathbf{F}$ is decoded to produce 3D points and auxiliary 2D attributes,
including positions, normals, visibility, and confidence,
through parallelized independent point queries.
Each input query token 
is parameterized as
\begin{equation}
    \begin{aligned}
        \mathbf{q}(u,v,\tsrc,\ttgt,\tcam)
        &= \mathrm{MLP}_q\big(
        \phi_{\mathrm{uv}}(u,v)
        + \phi_{\mathrm{patch}}(\mathbf{p}_{u,v}) \\
        &+ \mathbf{e}_{\mathrm{src}}(\tsrc)
        + \mathbf{e}_{\mathrm{tgt}}(\ttgt)
        + \mathbf{e}_{\mathrm{cam}}(\tcam)
        \big)
        \in \mathbb{R}^{C},
    \end{aligned}
    \label{eq:query}
\end{equation}
where $\phi_{\mathrm{uv}}$ is the encoded query coordinate,
$\phi_{\mathrm{patch}}(\mathbf{p}_{u,v})$ denotes the MLP-encoded local $9\times9$
RGB patch centered at $(u,v)$,
and $\mathbf{e}_{\mathrm{src}}$, $\mathbf{e}_{\mathrm{tgt}}$, and $\mathbf{e}_{\mathrm{cam}}$ represent the temporal embeddings.
The query $\mathbf{q}(u,v,\tsrc,\ttgt,\tcam)$ asks \textit{``what are the 3D position and the other attributes of the 3D point corresponding to the pixel at $(u, v)$ in source frame $\tsrc$, at time $\ttgt$, expressed in the camera coordinate system of $\tcam$''}.
Specifically, each temporal embedding is implemented as a discrete lookup table with 48 independently learnable entries, one for each temporal position~\citep{jiang2026d4rtpytorch}.

$\mathbf{q}$ is decoded by attending to $\mathbf{F}$ via a multi-layer cross-attention decoder $\mathcal{D}$:
\begin{equation}
    \label{eq:decoder}
    \mathbf{z}_{q} = \mathcal{D}(\mathbf{q}, \mathbf{F}) \in \mathbb{R}^{C}.
\end{equation}
The feature
$\mathbf{z}_{q}$
is then fed into an MLP prediction head $\mathcal{H}$ to produce the desired 3D and 2D attributes:
\begin{equation}
    \label{eq:output}
    \left(
    \hat{\mathbf{X}}^{3D},
    \hat{\mathbf{x}}^{2D},
    \hat{\mathbf{n}},
    \hat{\mathbf{d}},
    \hat{v},
    \hat{c}
    \right)
    =
    \mathcal{H}(\mathbf{z}_{q}).
\end{equation}
For each queried point $(u,v,\tsrc)$,
$\mathcal{H}$ outputs its 3D position $\hat{\mathbf{X}}^{3D}$ and surface normal $\hat{\mathbf{n}}$ at $\ttgt$,
as well as its dynamic 3D displacement $\hat{\mathbf{d}}$ from $\tsrc$ to $\ttgt$,
all expressed in the $\tcam$ camera coordinate system.
Besides,
$\mathcal{H}$ also outputs the 2D image-plane position $\hat{\mathbf{x}}^{2D}$ and visibility $\hat{v}$ of the queried point.
Finally, a confidence score $\hat{c}$ is predicted, indicating the network's certainty about its estimates.
The attributes are defined by corresponding losses and preprocessed targets.
For more details, see \citet{zhang2026d4rt} and \citet{li2026opend4rt}.

\subsection{Video Generators as 4D Perceivers}
\label{sec:encoder_decoder}
\methodname repurposes a video generator into a single-pass feed-forward 4D perceiver,
without multi-step denoising~\citep{ke2024repurposing,jiang2025geo4d}
or noise injection~\citep{zhan2024does,huang2026vidfm} to read out diffusion features.
This formulation enables direct feature extraction while avoiding the overhead of iterative denoising.
Specifically, \methodname consists of a video-generator-based encoder with a LoRA-adapted video DiT backbone, and a cross-attention query decoder, as described next.

\subsubsection{Video Generative Encoder}
\label{sec:encoder}

\heading{Feed-forward Encoder with LoRA}
We adopt Wan2.1-1.3B~\citep{wan2025} as our encoder backbone.
To repurpose a \emph{multi-step} diffusion model into a \emph{single-pass feed-forward encoder}, we first investigate how the amount of noise corruption applied to the input video latent affects the quality of the extracted DiT features. Here, the noise level $\sigma$ denotes the corruption strength applied to the input video latents.

\par

\begingroup
\setlength{\columnsep}{0pt}
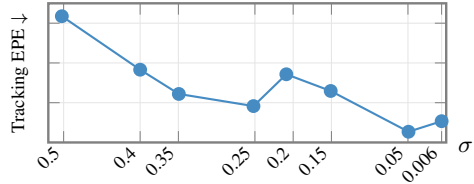
\begin{wrapfigure}{r}{0.49\columnwidth}
    \vspace{-8pt}
    \centering

    \begin{tikzpicture}
        \begin{axis}[
            width=\linewidth,
            height=0.5\linewidth,
            xlabel={$\sigma$},
            xlabel style={
                at={(axis description cs:1.005,-0.035)},
                anchor=west,
                font=\small
            },
            ylabel={Tracking EPE $\downarrow$},
            ylabel style={
                at={(axis description cs:-0.025,0.5)},
                anchor=south,
                font=\scriptsize
            },
            x dir=reverse,
            xmin=0.0,
            xmax=0.52,
            ymin=0.30,
            ymax=0.37,
            xtick={
                0.5,0.4,0.35,0.25,
                0.2,0.15,0.05,0.006
            },
            xticklabels={
                0.5,0.4,0.35,0.25,
                0.2,0.15,0.05,0.006
            },
            ytick={
                0.30,0.32,0.34,0.36
            },
            yticklabels={},
            tick label style={
                font=\scriptsize
            },
            x tick label style={
                rotate=45,
                anchor=east
            },
            grid=major,
            grid style={
                gray!20
            },
            axis line style={
                draw=gray
            },
            line width=0.9pt,
        ]
            \addplot[
                color=plotblue,
                mark=*,
                mark size=2.2pt,
                mark options={
                    fill=plotblue,
                    draw=plotblue
                },
                thick,
            ]
            coordinates {
                (0.502327,0.3635)
                (0.400001,0.3366)
                (0.349427,0.3244)
                (0.251872,0.3183)
                (0.208944,0.3343)
                (0.150946,0.3259)
                (0.049278,0.3054)
                (0.005973,0.3107)
            };
        \end{axis}
    \end{tikzpicture}

    \captionsetup{
        skip=-4pt,
        belowskip=-8pt
    }
    \caption{\textbf{Noise Level -- Tracking.}}
    \label{fig:feature_probing}
    \vspace{-10pt}
\end{wrapfigure}
We start from $\sigma=0.5$, a noise level used by a prior Wan2.1-based perception method~\citep{huang2026vidfm}.
We train identical models at decreasing noise levels and report the tracking End-Point Errors (EPEs).
As shown in \Cref{fig:feature_probing}, low-noise features
(\eg, $\sigma=0.05$) in general outperform the relatively high-noise features
(\eg, $\sigma=0.35\sim0.5$).
This observation is consistent with recent feed-forward perception
methods~\citep{zhu2026motioncrafter,wang2026genception}.
\WFclear
\endgroup

\noindent
Motivated by the stronger performance of low-noise features, and to further achieve a simple deterministic feed-forward formulation,
we directly pass clean video latents to the DiT without noise corruption, and use its final-block features before the velocity prediction head for query decoding.
Specifically, we process each input video $\bm{V}\in\mathbb{R}^{T\times H\times W\times 3}$ through the Wan VAE encoder $\mathcal{E}_{\mathrm{VAE}}$ and DiT blocks $\mathcal{F}_{\mathrm{DiT}}$,
obtaining the scene tokens $\mathbf{F}$ as:
\begin{equation}
    \label{eq:encoder_ours}
    \begin{aligned}
        \mathbf{z}_0 &= \mathcal{E}_{\mathrm{VAE}}(\bm{V}) \in \mathbb{R}^{(\frac{T-1}{4}+1) \times h \times w \times 16}, \\
        \mathbf{F} &= \mathcal{F}_\mathrm{DiT}(\mathrm{Patchify}(\mathbf{z}_0);\tau_\textrm{last},\mathbf{c}_{\emptyset}) \in \mathbb{R}^{N_\mathrm{F}\times 1536},
    \end{aligned}
\end{equation}
where $\tau_\textrm{last}$ denotes the final scheduler step time condition, $\mathbf{c}_{\emptyset}$ denotes the empty text prompt condition, and $N_\mathrm{F} = (\frac{T-1}{4}+1) \cdot \frac{h}{2} \cdot \frac{w}{2}$ is the number of scene tokens obtained by flattening the spatiotemporal dimensions.
During training, we freeze both $\mathcal{E}_{\mathrm{VAE}}$ and $\mathcal{F}_{\mathrm{DiT}}$, and adapt the DiT using a rank-512 LoRA~\citep{hu2021lora}, referred to as 4D-LoRA ($\Delta W_\mathrm{4D}^{512}$).

\subsubsection{Query Decoder}
\label{sec:decoder}

\textbf{Continuous Time Query} is a key component that allows the model to handle real-world videos of variable lengths.
Rather than using discrete lookup tables for a fixed 48-frame horizon~\citep{zhang2026d4rt},
we introduce a continuous temporal parameterization that allows the model to operate on variable-length sequences during both training and inference.
Taking source time $t_{\textrm{src}}$ as an example, we define:
\begin{equation}
    \label{eq:continuous_time}
    \begin{aligned}
        \alpha(\tsrc) &= \frac{\tsrc}{T-1},\; \tsrc \in \{0,\dots,T-1\}, \\
        \mathbf{e}_{\mathrm{src}}(\tsrc) &\coloneqq  
        (1-\alpha(\tsrc))\,\mathbf{e}_\mathrm{src, 0} + \alpha(\tsrc)\,\mathbf{e}_\mathrm{src, 1},
    \end{aligned}
\end{equation}
where $\mathbf{e}_{\mathrm{src},0}$ and $\mathbf{e}_{\mathrm{src},1}$ are two learnable temporal embeddings representing the start and end of the video clip, respectively.
This parameterization represents each timestamp by its normalized position within the clip, making the time query independent of the absolute sequence length $T$ and thereby enhancing the model's temporal flexibility.

\heading{Cross-Attention Decoder}
The decoder consists of multiple cross-attention blocks, where the query attends to the scene tokens $\mathbf{F}$ at each layer.
We construct the query embedding $\mathbf{q}$ from the continuous time query embeddings $\mathbf{e}_{\mathrm{src}}$, $\mathbf{e}_{\mathrm{tgt}}$, and $\mathbf{e}_{\mathrm{cam}}$,
the 2D position embedding $\phi_{\mathrm{uv}}(u,v)$ and the RGB patch embedding $\phi_{\mathrm{patch}}(\mathbf{p}_{u,v})$. 
Then we feed $\mathbf{q}$ and $\mathbf{F}$ into a cross-attention decoder $\mathcal{D}$ to obtain the query feature $\mathbf{z}_q$, which is further mapped by an MLP head $\mathcal{H}$ to the complete set of 4D attributes, as defined in \Cref{eq:decoder,eq:output}.


\subsection{Plug-in Geometry Prior}
\label{sec:geometry_prior}

Recent GFM-finetuned 4D perceivers~\citep{sucar2026vdpm, luo20264rc, lin2026sm4rt}
benefit from geometry priors, which are learned through large-scale dense 3D pretraining.
In contrast, \methodname adapts a pretrained video generator to exploit its motion prior, while its sparse query-based supervision alone provides a weaker signal for
learning geometry than the GFM-finetuned methods.
To narrow this gap, we propose a natural solution within our framework: by \emph{plugging in} a pretrained video-depth adapter as a frozen geometry prior, we are able to reuse knowledge learned from dense depth estimation without a separate dense 3D pretraining stage.

\heading{Double-LoRA Architecture}
To incorporate pretrained dense geometric knowledge,
we adopt a plug-in geometry prior from a pretrained video depth estimation model,
DVD~\citep{zhang2026dvd}, which is also built upon Wan2.1.
We combine the pretrained Depth-LoRA ($\Delta W_\mathrm{Depth}$) with our 4D-LoRA ($\Delta W_\mathrm{4D}$) (Sec.~\ref{sec:encoder}), forming a \emph{double-LoRA} architecture.
For each adapted layer, the effective weight is
\begin{equation}
    W = W_{\mathrm{DiT}} + \Delta W_{\mathrm{Depth}}^{512} + \Delta W_{\mathrm{4D}}^{512},
    \label{eq:geometry_prior_composition}
\end{equation}
where $W_{\mathrm{DiT}}$ denotes the DiT weight of the pretrained video generator, and $\Delta W_{\mathrm{Depth}}^{512}$ and $\Delta W_{\mathrm{4D}}^{512}$ are the depth and 4D weight updates with a rank of 512, respectively.

\heading{Lightweight Training}
Following prior work on retaining pretrained adapters during task adaptation~\citep{yu2025progressivelora}, we freeze the VAE encoder, $W_{\mathrm{DiT}}$ and $\Delta W_{\mathrm{Depth}}$,
and jointly optimize $\Delta W_{\mathrm{4D}}$, the query embeddings, the query decoder, and the prediction head. Thus, the pretrained depth update is retained while a separate update is learned for 4D perception under the task objective. 
We point out that the double-LoRA variant requires no additional dense prediction head and has the \emph{same number of trainable parameters} as the variant using 4D-LoRA alone. 
In Sec.~\ref{sec:flexibility}, we compare several adaptation strategies,
providing empirical support for our design.

\subsection{Training Objective}
\label{sec:loss}
Our method predicts a full set of 4D attributes and is trained with a composite loss:
\begin{equation}
\label{eq:loss}
    \mathcal{L} = 
    \lambda_{3D}\mathcal{L}_{3D}
    +
    \lambda_{\mathrm{conf}}\mathcal{L}_{\mathrm{conf}}
    +
    \lambda_{2D}\mathcal{L}_{2D}
    +
    \lambda_{\mathrm{vis}}\mathcal{L}_{\mathrm{vis}}
    +
    \lambda_{\mathrm{disp}}\mathcal{L}_{\mathrm{disp}}
    +
    \lambda_{\mathrm{normal}}\mathcal{L}_{\mathrm{normal}},
\end{equation}
where $\mathcal{L}_{3D}$, $\mathcal{L}_{\mathrm{conf}}$, $\mathcal{L}_{2D}$, $\mathcal{L}_{\mathrm{vis}}$, $\mathcal{L}_{\mathrm{disp}}$, and $\mathcal{L}_{\mathrm{normal}}$ denote the 3D position, confidence, 2D position, visibility, 3D displacement, and surface-normal losses, respectively. See \Cref{sec:appendix_losses} for more details.




\section{Experiments}
\label{sec:exp}

\subsection{Training Setups}
\heading{Training Datasets}
We use approximately 200K video clips from 11 public datasets for training:
Syn4D~\citep{jiang2026syn4d},
Dynamic Replica~\citep{karaev23dynamicstereo},
PointOdyssey~\citep{zheng2023pointodyssey},
Kubric CoTracker~\citep{karaev2025cotracker3},
DL3DV~\citep{ling2024dl3dv},
ScanNet++~\citep{yeshwanthliu2023scannetpp},
Virtual KITTI 2~\citep{cabon2020virtualkitti2},
MVS-Synth~\citep{huang2018deepmvs},
Spring~\citep{mehl2023spring},
Waymo~\citep{sun2020waymo},
and Aria Digital Twin~\citep{pan2023aria}.

\textbf{Data Usage.}
We distinguish two types of input frames: \emph{encoder frames}, which are input to the encoder to extract scene tokens, and \emph{query frames}, which provide local RGB patches for decoder queries. 
They share the same aspect ratio but may differ in resolution.
Encoder frames are maximally center-cropped to one of five predefined aspect ratios and uniformly resized to the largest resolution within a $480\times832$ bound,
while query frames generally retain their original resolution.
To speed up training,
we use $\mathcal{E}_\text{VAE}$ to pre-encode all video clips into latents and discard the VAE during training. 

We also apply temporal augmentation to handle varying motion speeds.
To this end,
each raw video is subsampled into 49-frame clips with temporal intervals ranging from 1 to 4. During training, we randomly slice the first $21\sim49$ frames in steps of 4, and head-truncate the video latent accordingly.
We also apply downsizing-only spatial augmentation to the query frames, so that the model sees a variety of query-patch fields of view.
See \Cref{sec:appendix_dataset,sec:appendix_preprocess_and_sampling} for more details about datasets and data preprocessing \& sampling, respectively.

\heading{Architecture}
We evaluate two adaptation variants.
The \emph{basic} variant,
denoted as `Ours',
leverages \emph{motion} priors from a large-scale video generator,
with a trainable rank-512 4D-LoRA, $\Delta W_{\mathrm{4D}}^{512}$.
The \emph{double-LoRA} variant,
denoted as `Ours-\textbf{II}',
additionally incorporates \emph{geometry} priors through the frozen rank-512 Depth-LoRA,
$\Delta W_{\mathrm{Depth}}^{512}$ (\Cref{sec:geometry_prior}).
Both variants have $0.70$B trainable encoder parameters.
We use an 8-block cross-attention decoder, which has $0.18$B trainable parameters.

\heading{Optimization}
We train for 75K optimization steps using AdamW~\citep{loshchilov2017adamw} on 8 NVIDIA H200 GPUs.
Gradient accumulation yields an effective batch size of 32 video clips.
The learning rate is linearly warmed up to a peak of $10^{-4}$ over the first 2.5K steps and then follows a cosine decay schedule to $10^{-6}$.
We apply a maximum gradient norm clipping of 10.

\subsection{World-coordinate Tracking}
\label{sec:exp_world_coord_tracking}

\begin{figure*}[tb!]
    \centering
    \includegraphics[width=\textwidth]{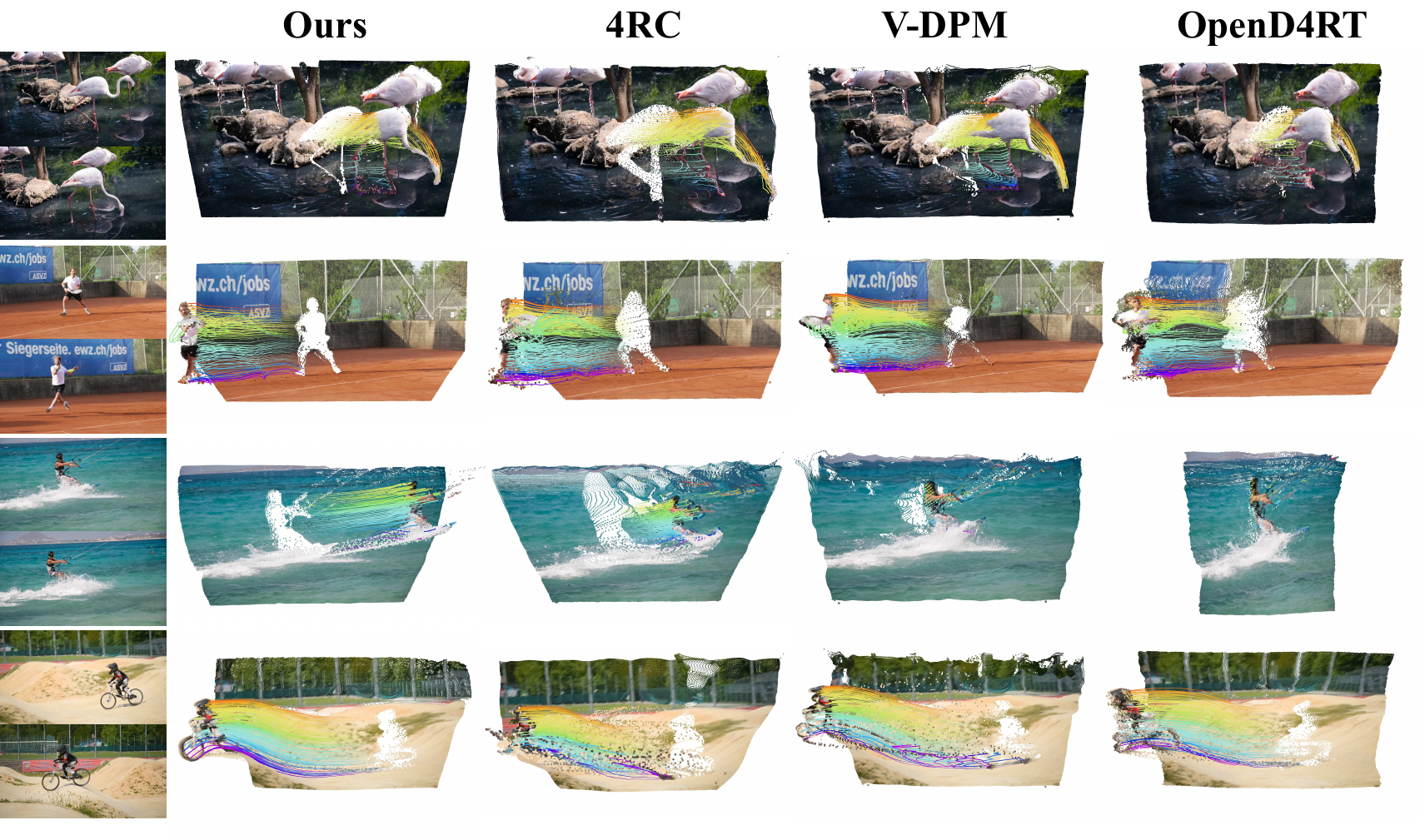}
    \captionsetup{skip=2pt}
    \caption{
        \textbf{Qualitative Results of 3D Tracking} across different methods, consisting of both static and dynamic content. We colorize the dynamic foreground tracklines.
    }
    \label{fig:track_visualization}
\vspace{-15pt}
\end{figure*}

\begin{table*}[tb!]
\centering
\captionsetup{skip=2pt}
\caption{\textbf{Quantitative Results on World-Coordinate Tracking.} We report the standard Average Percent of Points within Delta (APD$~\uparrow$) and End-Point Error (EPE$~\downarrow$) for \underline{All} points and only \underline{Dynamic} points in separate groups.
Each group is aligned to the GT tracks using a global median alignment over 64 frames.
We highlight the \besthl{best}, \secondbesthl{second} and \thirdbesthl{third} best results. }
\label{tab:worldcoord-tracking}
\setlength{\tabcolsep}{2pt}
\renewcommand{\arraystretch}{1.1}
\resizebox{\textwidth}{!}{%
    \begin{tabular}{@{}lcccccccc@{}}
    \toprule
    \cmidrule(lr){2-9}
    \multirow{3}{*}{\textbf{Method}}
    & \multicolumn{2}{c}{\textbf{ADT}}
    & \multicolumn{2}{c}{\textbf{DR}}
    & \multicolumn{2}{c}{\textbf{PO}}
    & \multicolumn{2}{c}{\textbf{PStudio}} \\

    \cmidrule(lr){2-3}
    \cmidrule(lr){4-5}
    \cmidrule(lr){6-7} 
    \cmidrule(lr){8-9} 
    
    & {APD}$\uparrow$ & {EPE}$\downarrow$
    & {APD}$\uparrow$ & {EPE}$\downarrow$
    & {APD}$\uparrow$ & {EPE}$\downarrow$
    & {APD}$\uparrow$ & {EPE}$\downarrow$ \\
    
    & {All / Dyn.}
    & {All / Dyn.}
    & {All / Dyn.}
    & {All / Dyn.}
    & {All / Dyn.}
    & {All / Dyn.}
    & {All}
    & {All} \\
    \midrule
    
    MonST3R~\citep{zhang2025monst3r}
    & 73.58 / 82.17
    & 0.326 / 0.176
    & 53.79 / 52.87
    & 0.514 / 0.547
    & 32.95 / 35.02
    & 0.902 / 1.095
    & 58.14
    & 0.454 \\
    
    STv2~\citep{xiao2025spatialtrackerv2}
    & \secondbesthl{91.89} / \thirdbesthl{84.44}
    & \besthl{0.095} / \besthl{0.143}
    & 70.71 / 68.52
    & 0.287 / 0.333
    & 69.48 / 65.35
    & 0.388 / 0.500
    & 73.52
    & 0.237 \\
    
    St4RTrack~\citep{feng2025st4rtrack}
    & 73.02 / 78.01
    & 0.294 / 0.209
    & 74.34 / 77.35
    & 0.260 / 0.207
    & 67.34 / 72.03
    & 0.317 / 0.247
    & 70.67
    & 0.249 \\
    
    Any4D~\citep{karhade2026any4d}
    & 58.47 / 69.73
    & 0.408 / 0.271
    & 79.14 / 70.31
    & 0.217 / 0.309
    & 67.28 / 68.69
    & 0.347 / 0.366
    & 64.31
    & 0.296 \\
    
    TraceAnything~\citep{liu2026trace}
    & 75.67 / 73.85
    & 0.251 / 0.259
    & 60.62 / 60.44
    & 0.570 / 0.463
    & 39.47 / 47.03
    & 1.053 / 0.744
    & 71.41
    & 0.272 \\
    
    V-DPM~\citep{sucar2026vdpm}
    & 85.74 / 74.46
    & 0.170 / 0.271
    & 75.33 / 78.35
    & 0.249 / 0.220
    & \besthl{81.27} / \thirdbesthl{82.54}
    & \besthl{0.189} / \thirdbesthl{0.176}
    & 78.75
    & 0.179 \\
    
    4RC~\citep{luo20264rc}
    & 84.20 / 72.94
    & 0.176 / 0.270
    & \thirdbesthl{82.73} / 73.82
    & \thirdbesthl{0.190} / 0.289
    & \secondbesthl{80.28} / 76.11
    & 0.268 / 0.360
    & 69.01
    & 0.261 \\
    
    OpenD4RT~\citep{li2026opend4rt}
    & 72.18 / 74.22
    & 0.275 / 0.309
    & 76.68 / \thirdbesthl{82.06}
    & 0.243 / \thirdbesthl{0.185}
    & 67.98 / 74.25
    & 0.317 / 0.258
    & \thirdbesthl{79.61}
    & \thirdbesthl{0.175} \\

    
    \midrule
    


    Ours
    & \thirdbesthl{91.37} / \secondbesthl{86.43}
    & \thirdbesthl{0.121} / \thirdbesthl{0.170}
    & \secondbesthl{85.78} / \secondbesthl{91.69}
    & \secondbesthl{0.146} / \secondbesthl{0.095}
    & 74.49 / \secondbesthl{82.76}
    & \thirdbesthl{0.230} / \secondbesthl{0.166}
    & \secondbesthl{81.92}
    & \secondbesthl{0.157} \\
    
    Ours-\textbf{II}
    & \besthl{91.93} / \besthl{86.82}
    & \secondbesthl{0.112} / \secondbesthl{0.160}
    & \besthl{88.18} / \besthl{92.52}
    & \besthl{0.130} / \besthl{0.090}
    & \thirdbesthl{78.11} / \besthl{84.92}
    & \secondbesthl{0.196} / \besthl{0.138}
    & \besthl{82.11}
    & \besthl{0.149} \\
    
    \bottomrule
    \end{tabular}%
}
\end{table*}
\vspace{-6pt}

\heading{Evaluation Protocol}
We evaluate the performance of 3D tracking on WorldTrack~\citep{feng2025st4rtrack},
which includes Aria Digital Twin (ADT),
Dynamic Replica (DR),
PointOdyssey (PO),
and Panoptic Studio (PS)~\citep{joo2015pstudio}.
We report results for \emph{All} and \emph{Dynamic} points on ADT, DR, and PO, and for all points on PS, whose evaluated tracks are entirely dynamic.
Dynamic points are identified by accumulated ground-truth displacements exceeding a certain threshold,
default set to $5\times10^{-3}$\,m.
We report the standard 3D tracking metrics:
\emph{Average percent of points within delta error} (APD) and \emph{End-Point Error} (EPE).
Following MonST3R~\citep{zhang2025monst3r}, 
predictions are aligned to ground truth with a single median scale factor per evaluation group over all frames.
To satisfy the VAE's temporal constraint,
we process 65 frames and evaluate the first 64.
See \Cref{sec:appendix_world_coord_tracking} for more details. 

\heading{Quantitative Results}
As shown in \Cref{tab:worldcoord-tracking},
\methodname outperforms prior methods by a clear margin on DR and PS, where PS is a challenging, fully dynamic human--object benchmark on which no method has been trained with comparable data.
Meanwhile,
our method remains competitive on ADT and PO,
especially for the dynamic points.
Note that the basic variant `Ours' already surpasses many prior methods,
highlighting the effectiveness of motion priors from a pretrained video generator.
The depth prior further improves every reported tracking metric over the basic variant.
This provides a new strategy for leveraging motion and geometry priors in 4D tasks.

\heading{Qualitative Results}
We visualize the 3D tracking results of our method on in-the-wild DAVIS~\citep{perazzi2016davis},
along with 4RC, V-DPM and OpenD4RT in \Cref{fig:track_visualization,fig:appendix_additional_vis}.
As shown in \Cref{fig:track_visualization},
our method produces more visually plausible motion tracking (\eg, the articulated body and arm motion in \textit{Tennis}, the body and board motion in \textit{Kite-surf}) and better preserves delicate structures (\eg, the thin legs in \textit{Flamingo}, the front wheel in \textit{Bmx-bumps}). 
For more qualitative results, please see \Cref{sec:append_qualitative_results} and the video page either at \url{https://gen4p.github.io/} or in supplementary material. 

\begin{table*}[tb!]
\centering
\captionsetup{skip=2pt}
\caption{\textbf{Quantitative Results on Video Depth Estimation.}
Results are reported for three unseen datasets: Bonn, Sintel, and KITTI. 
We highlight the \besthl{best}, \secondbesthl{second} and \thirdbesthl{third} best results. }
\label{tab:video_depth}
\resizebox{\textwidth}{!}{%
\begin{tabular}{@{}l cc cc cc@{}}
\toprule
\multirow{2}{*}{\textbf{Method}}
& \multicolumn{2}{c}{\textbf{Bonn}}
& \multicolumn{2}{c}{\textbf{Sintel}}
& \multicolumn{2}{c}{\textbf{KITTI}} \\
\cmidrule(lr){2-3}
\cmidrule(lr){4-5}
\cmidrule(lr){6-7}
& AbsRel $\downarrow$
& $\delta_{<1.25}$ $\uparrow$
& AbsRel $\downarrow$
& $\delta_{<1.25}$ $\uparrow$
& AbsRel $\downarrow$
& $\delta_{<1.25}$ $\uparrow$ \\
\midrule

MonST3R~\citep{zhang2025monst3r}
& 0.062 & 96.36
& 0.300 & 58.01
& 0.089 & 91.11 \\

Spann3R~\citep{wang2025spann3r}
& 0.139 & 83.21
& 0.306 & 68.14
& 0.114 & 89.49 \\

VGGT~\citep{wang2025vggt}
& \besthl{0.049} & \besthl{97.40}
& 0.211 & \thirdbesthl{72.55}
& 0.059 & 96.15 \\

DepthAnythingv3~\citep{lin2025da3}
& \besthl{0.049} & 97.28
& \thirdbesthl{0.200} & 70.80
& \besthl{0.043} & \besthl{97.98} \\

MapAnything~\citep{keetha2026mapanything}
& \thirdbesthl{0.058} & 96.00
& 0.293 & 69.70
& \thirdbesthl{0.057} & \thirdbesthl{97.28} \\

Any4D~\citep{karhade2026any4d}
& \secondbesthl{0.051} & \thirdbesthl{97.31}
& 0.218 & 69.42
& 0.063 & 96.95 \\


V-DPM~\citep{sucar2026vdpm}
& 0.072 & 95.87
& 0.243 & 67.60
& 0.076 & 94.88 \\

4RC~\citep{luo20264rc}
& \secondbesthl{0.051} & 97.10
& 0.258 & 66.61
& \secondbesthl{0.054} & 95.71 \\

OpenD4RT~\citep{li2026opend4rt}
& \thirdbesthl{0.058} & 97.23
& 0.302 & 60.39
& 0.089 & 93.62 \\



\midrule





Ours
& 0.074 & 96.49
& \besthl{0.180} & \besthl{76.13}
& 0.062 & 96.68 \\

Ours-\textbf{II}
& 0.059 & \secondbesthl{97.33}
& \secondbesthl{0.182} & \secondbesthl{74.99}
& 0.059 & \secondbesthl{97.32} \\

\bottomrule
\end{tabular}%
}
\vspace{-10pt}
\end{table*}

\subsection{Video Depth Estimation}
\label{sec:exp_video_depth}

\heading{Evaluation Protocol}
To demonstrate the out-of-domain generalization ability,
we directly evaluate video depth estimation on three unseen datasets:
Bonn~\citep{palazzolo2019bonn}, Sintel~\citep{butler2012sintel}, and KITTI~\citep{geiger2013kitti},
covering realistic indoor scenes with moving people,
synthetic scenes with substantial camera and character motion,
and \emph{real} driving scenes captured with LiDAR sensors, respectively.
We report the standard Absolute Relative Error (AbsRel$~\downarrow$) and Inlier Ratio at a threshold of $1.25$ ($\delta_{<1.25}\uparrow$). 
Following MonST3R~\citep{zhang2025monst3r},
we evaluate up to 110 frames per sequence. 
We infer our method using 49-frame chunks with a 24-frame overlap.
For fair comparison, our query frames are resized to a maximum side length of 512 pixels while preserving the aspect ratio, and the encoder frames are resized to the predefined resolutions.
Regarding the evaluation protocols and inference setups, see \Cref{sec:appendix_video_depth} for more details. 

\heading{Results}
As shown in \Cref{tab:video_depth},
our method outperforms others on Sintel, 
which is the most challenging dataset due to its motion complexity~\citep{zhang2026d4rt}, 
while remaining competitive on Bonn and KITTI,
second best on Bonn and KITTI for $\delta_{<1.25}$,
two realistic datasets of body motion and driving scenes, respectively.
These results demonstrate our method's generalization to both realistic and synthetic scenes with strong camera and foreground motion. 


\begin{table*}[tb!]
\centering
\captionsetup{skip=2pt}
\caption{\textbf{Quantitative Results on Camera Pose Estimation.} We report the standard Absolute Translation Error ($\mathrm{ATE}\downarrow$), Relative Pose Error-Translation ($\mathrm{RPE_T}\downarrow$), and Relative Pose Error-Rotation ($\mathrm{RPE_R}\downarrow$). 
We highlight the \besthl{best} and \secondbesthl{second} best results.}
\label{tab:camera_pose}
\resizebox{\textwidth}{!}{%
\begin{tabular}{@{}l ccc ccc ccc@{}}
\toprule
\multirow{2}{*}{\textbf{Method}}
& \multicolumn{3}{c}{\textbf{Sintel}}
& \multicolumn{3}{c}{\textbf{TUM-Dynamics}}
& \multicolumn{3}{c}{\textbf{ScanNetv2}} \\
\cmidrule(lr){2-4}
\cmidrule(lr){5-7}
\cmidrule(lr){8-10}
& ATE $\downarrow$
& $\mathrm{RPE_T}\downarrow$
& $\mathrm{RPE_R}\downarrow$
& ATE $\downarrow$
& $\mathrm{RPE_T}\downarrow$
& $\mathrm{RPE_R}\downarrow$
& ATE $\downarrow$
& $\mathrm{RPE_T}\downarrow$
& $\mathrm{RPE_R}\downarrow$ \\
\midrule

VGGT~\citep{wang2025vggt}
& 0.172 & 0.061 & 0.471
& \secondbesthl{0.012} & 0.010 & \besthl{0.309}
& 0.034 & 0.015 & \secondbesthl{0.376} \\

DepthAnythingv3~\citep{lin2025da3}
& 0.122 & 0.043 & 0.348
& \secondbesthl{0.012} & \secondbesthl{0.008} & \secondbesthl{0.310}
& \secondbesthl{0.032} & \besthl{0.012} & \besthl{0.355} \\

MapAnything~\citep{keetha2026mapanything}
& 0.218 & 0.075 & 0.655
& 0.028 & 0.023 & 0.372
& 0.051 & 0.024 & 0.719 \\

Any4D~\citep{karhade2026any4d}
& 0.212 & 0.121 & 2.650
& 0.044 & 0.023 & 0.439
& 0.063 & 0.028 & 1.061 \\

V-DPM~\citep{sucar2026vdpm}
& 0.139 & 0.053 & 0.454
& 0.023 & 0.010 & 0.321
& 0.035 & 0.014 & 0.394 \\

4RC~\citep{luo20264rc}
& 0.138 & 0.052 & 0.429
& \besthl{0.011} & 0.009 & 0.312
& \besthl{0.031} & \besthl{0.012} & 0.384 \\

OpenD4RT~\citep{li2026opend4rt}
& 0.131 & 0.040 & 0.654
& 0.038 & 0.009 & 0.600
& 0.063 & 0.015 & 0.570 \\

\midrule



Ours
& \secondbesthl{0.114} & \secondbesthl{0.024} & \secondbesthl{0.310}
& 0.026 & \besthl{0.007} & 0.516
& 0.060 & 0.015 & 0.577 \\

Ours-\textbf{II}
& \besthl{0.104} & \besthl{0.023} & \besthl{0.306}
& 0.024 & \besthl{0.007} & 0.497
& 0.042 & \secondbesthl{0.013} & 0.527 \\

\bottomrule
\end{tabular}%
}
\end{table*}
\vspace{-3pt}

\subsection{Camera Pose Estimation}
\label{sec:exp_camera_pose}

\heading{Evaluation Protocol}
We evaluate camera pose estimation on Sintel, TUM-Dynamics~\citep{sturm2012tumdynamic} and ScanNetv2~\citep{dai2017scannet}. 
Following MonST3R~\citep{zhang2025monst3r},
we evaluate on the 14-sequence subset of Sintel,
and the first 90-frame evaluation setup with a temporal stride of three for TUM-Dynamics and ScanNetv2.
Our model processes 49-frame chunks with 34 overlapping frames.
Regarding the evaluation protocols and inference setups, see \Cref{sec:appendix_camera_pose} for more details.

\heading{Results}
Ours-II achieves the lowest errors on all three Sintel metrics among the compared methods (\Cref{tab:camera_pose}).
Both variants also achieve the lowest relative translation error on TUM-Dynamics, at 0.007, and remain competitive on ScanNetv2.

\begin{figure}[t]
    \centering
    \captionsetup{skip=-2pt}
    \caption{
        \textbf{Spatial and Temporal Flexibility} of our method, with 4RC as baseline.
        (a) Video depth estimation across different input resolutions on Sintel, with 4RC evaluated on the closest valid resolution.
        (b) Point tracking across different sequence lengths on DR.
    }

    \begin{subfigure}[t]{0.43\linewidth}
        \centering
        \begin{tikzpicture}

            \begin{axis}[
                width=0.84\linewidth,
                height=3.2cm,
                symbolic x coords={
                    218$\times$512,
                    327$\times$768,
                    436$\times$1024
                },
                xtick=data,
                xticklabel style={font=\small},
                ylabel={AbsRel $\downarrow$},
                ylabel style={text=black!70,font=\small},
                yticklabel style={
                    text=black!70,
                    font=\small,
                    /pgf/number format/fixed,
                    /pgf/number format/precision=2,
                    /pgf/number format/fixed zerofill
                },
                ymin=0.17,
                ymax=0.36,
                ytick={0.18,0.24,0.30,0.36},
                scaled y ticks=false,
                yticklabel={
                    \pgfmathprintnumber[fixed,precision=2]{\tick}
                },
                axis y line*=left,
                axis x line*=bottom,
                grid=major,
                enlargelimits=false,
                after end axis/.code={
                    \draw[line width=0.4pt]
                        (rel axis cs:0,1) -- (rel axis cs:1,1);
                },
                legend to name=flexibilitylegend,
                legend columns=1,
                legend style={
                    draw=black!60,
                    fill=white,
                    font=\tiny,
                    inner xsep=2pt,
                    inner ysep=1.5pt,
                    column sep=2pt
                },
                legend cell align=left
            ]

            \addplot[
                color=plotblue,
                mark=*,
                thick
            ] coordinates {
                (218$\times$512, 0.182)
                (327$\times$768, 0.180)
                (436$\times$1024, 0.179)
            };
            \addlegendentry{Ours}

            \addplot[
                color=black!70,
                mark=x,
                mark size=3pt
            ] coordinates {
                (218$\times$512, 0.259)
                (327$\times$768, 0.326)
                (436$\times$1024, 0.338)
            };
            \addlegendentry{4RC}

            \end{axis}

        \end{tikzpicture}
        \caption{Query Resolutions -- Video Depth.}
        \label{fig:flexibility_resolution_generalization}
    \end{subfigure}%
    \hfill%
    \raisebox{1.35cm}[0pt][0pt]{%
        \makebox[0pt][c]{%
            \hspace*{-4mm}%
            \makebox[0pt][c]{%
                \pgfplotslegendfromname{flexibilitylegend}%
            }%
        }%
    }%
    \hfill%
    \hfill%
    \begin{subfigure}[t]{0.56\linewidth}
        \centering
        \begin{tikzpicture}

            \begin{axis}[
                width=0.84\linewidth,
                height=3.2cm,
                xmin=41,
                xmax=105,
                xtick={41,49,57,65,73,81,89,97,105},
                xticklabel style={font=\small},
                ylabel={EPE $\downarrow$},
                ylabel style={text=black!70,font=\small},
                yticklabel style={
                    text=black!70,
                    font=\small,
                    /pgf/number format/fixed,
                    /pgf/number format/precision=3,
                    /pgf/number format/fixed zerofill
                },
                ymin=0.12,
                ymax=0.20,
                ytick={0.12,0.14,0.16,0.18,0.20},
                scaled y ticks=false,
                yticklabel={
                    \pgfmathprintnumber[fixed,precision=3]{\tick}
                },
                axis y line*=left,
                axis x line*=bottom,
                grid=major,
                enlargelimits=false,
                after end axis/.code={
                    \draw[line width=0.4pt]
                        (rel axis cs:0,1) -- (rel axis cs:1,1);
                }
            ]

            \addplot[
                color=plotblue,
                mark=*,
                thick
            ] coordinates {
                (41, 0.133)
                (49, 0.131)
                (57, 0.130)
                (65, 0.130)
                (73, 0.132)
                (81, 0.136)
                (89, 0.140)
                (97, 0.143)
                (105, 0.148)
            };

            \addplot[
                color=black!70,
                mark=x,
                mark size=3pt
            ] coordinates {
                (41, 0.191)
                (49, 0.190)
                (57, 0.190)
                (65, 0.189)
                (73, 0.188)
                (81, 0.190)
                (89, 0.188)
                (97, 0.189)
                (105, 0.189)
            };

            \end{axis}

        \end{tikzpicture}
        \caption{Inference Length -- Tracking.}
        \label{fig:flexibility_length_generalization}
    \end{subfigure}

    \label{fig:spatiotemporal_flexibility}
\vspace{-20pt}
\end{figure}
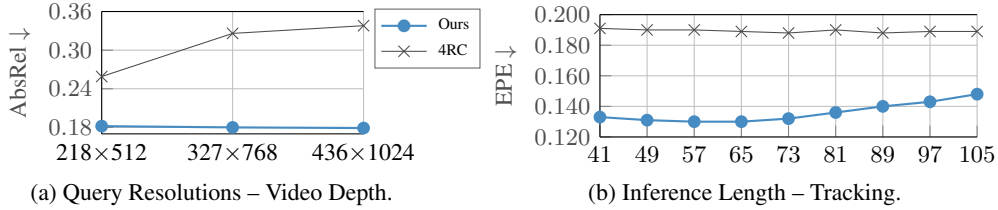

\subsection{Analysis}
\label{sec:flexibility}

We run a number of ablations to analyze \methodname.
All variants are trained for 45K steps at an effective batch size of 16,
on a mixture of DR, PO, Kubric CoTracker and Syn4D
with an encoder frame resolution of $256\times512$ and a 6-layer decoder. 
We evaluate on both world-coordinate tracking and video depth,
aligning to GT using a global scale and shift factor.

\heading{Flexibility}
Rows (a)--(b) of \Cref{tab:ablation} ablate our time-embedding design,
showing large improvements in ADT,
while maintaining strong performance on other metrics.
As shown in \Cref{fig:flexibility_length_generalization},
\methodname is naturally flexible in handling variable lengths,
from sequences within the 49-frame training length to substantially longer sequences of up to 105 frames, without significant degradation in performance.
We also evaluate the spatial flexibility of our method by testing it on different query resolutions.
Results in \Cref{fig:flexibility_resolution_generalization} show that AbsRel remains stable across a wide range of query resolutions.


\begin{figure*}[tb!]
    \centering
    \setlength{\tabcolsep}{2pt}
    \renewcommand{\arraystretch}{0}

    \newcommand{\inputpair}[2]{
        \begin{minipage}[t]{\linewidth}
            \vspace{0pt}
            \centering
            #1\\[-1pt]
            #2
        \end{minipage}
    }

    \newcommand{\resultcell}[1]{
        \begin{minipage}[t]{\linewidth}
            \vspace{0pt}
            \centering
            \makebox[\linewidth][c]{#1}
        \end{minipage}
    }

    \begin{tabular}{
        @{}
        p{0.106\textwidth}
        p{0.210\textwidth}
        p{0.210\textwidth}
        p{0.210\textwidth}
        p{0.210\textwidth}
        @{}
    }

    &
    \multicolumn{2}{c}{\bfseries Ours}
    &
    \multicolumn{2}{c}{\bfseries 4RC}
    \\[2pt]

    \inputpair
    {
        \includegraphics[width=\linewidth]{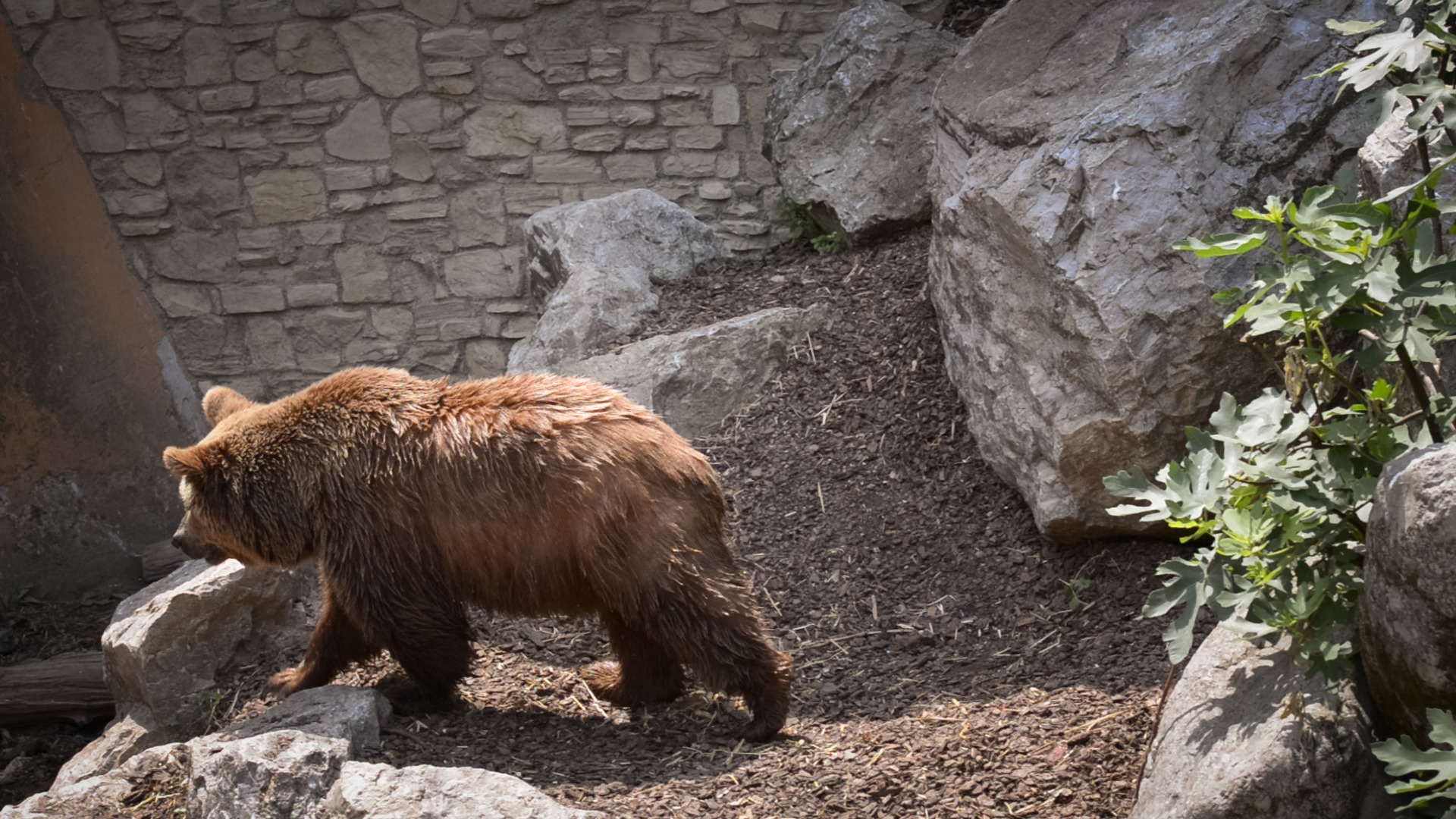}
    }
    {
        \includegraphics[width=\linewidth]{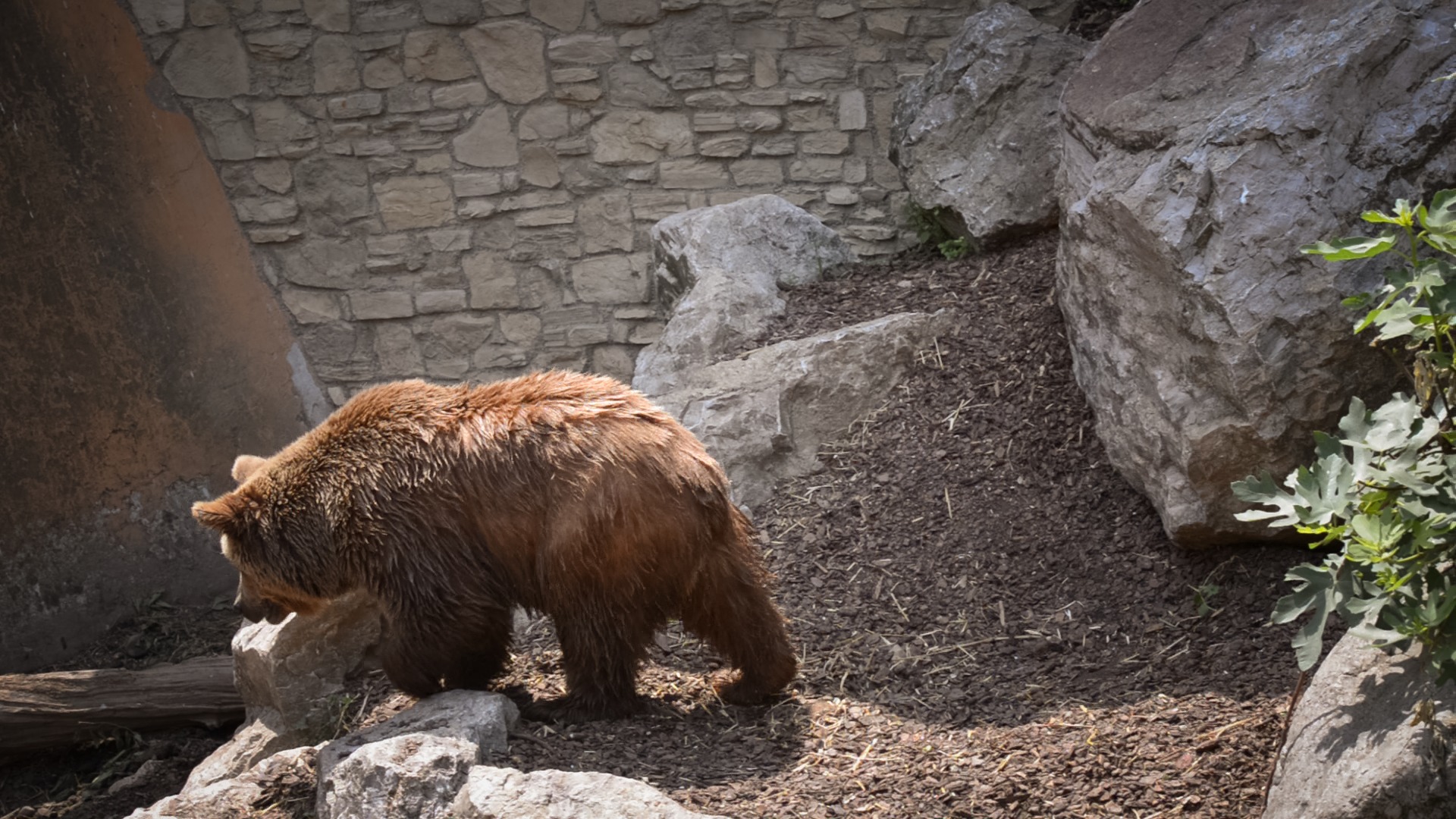}
    }
    &
    \resultcell{
        \includegraphics[width=\linewidth]{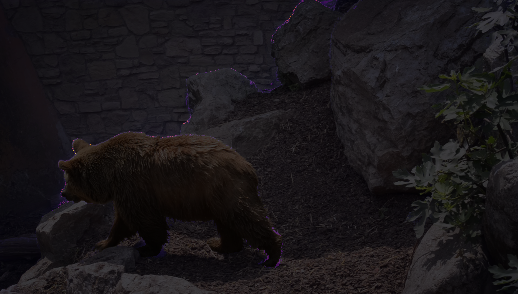}
    }
    &
    \resultcell{
        \includegraphics[width=\linewidth]{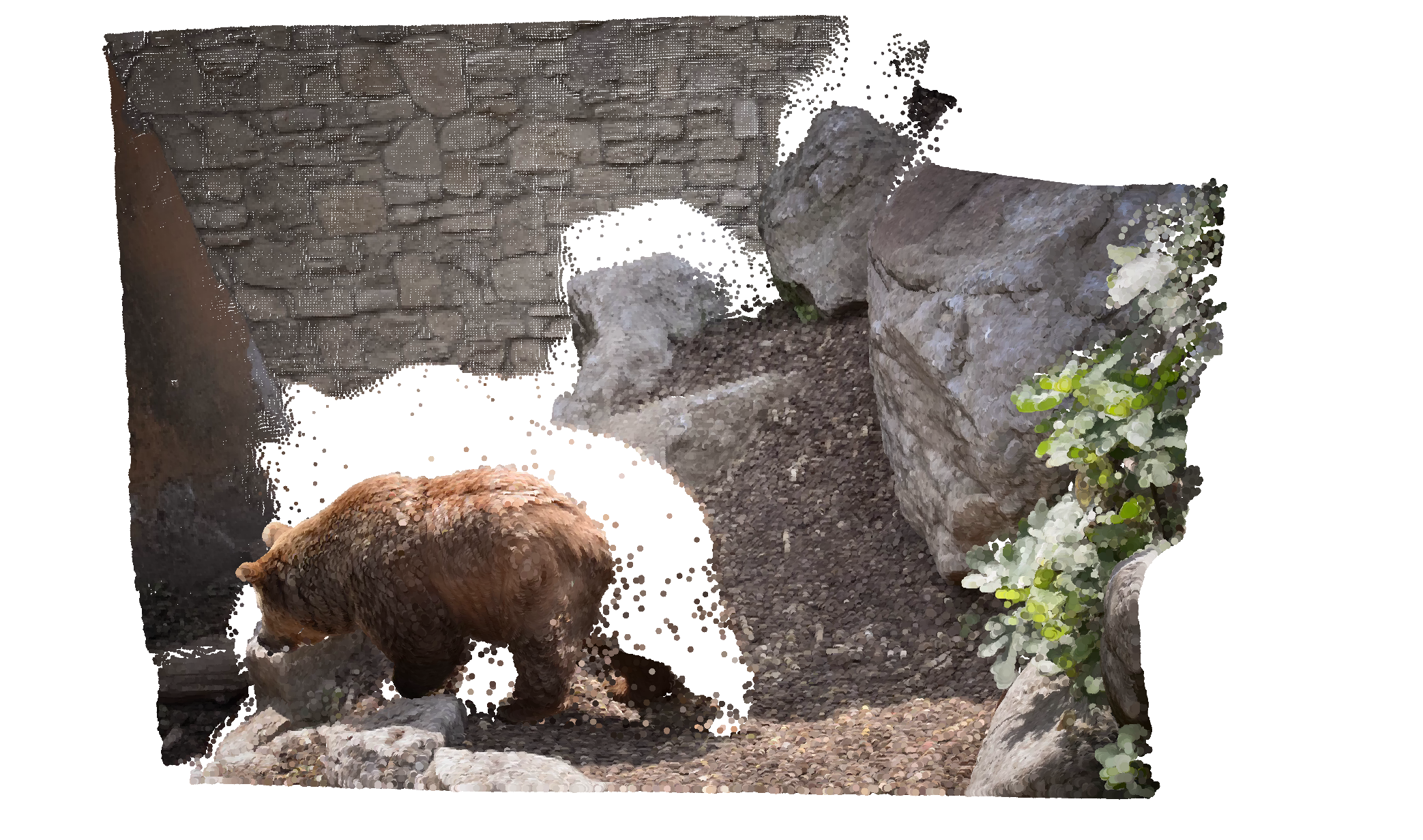}
    }
    &
    \resultcell{
        \includegraphics[width=\linewidth]{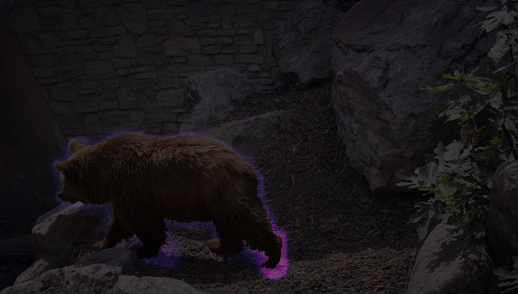}

    }
    &
    \resultcell{
        \includegraphics[width=\linewidth]{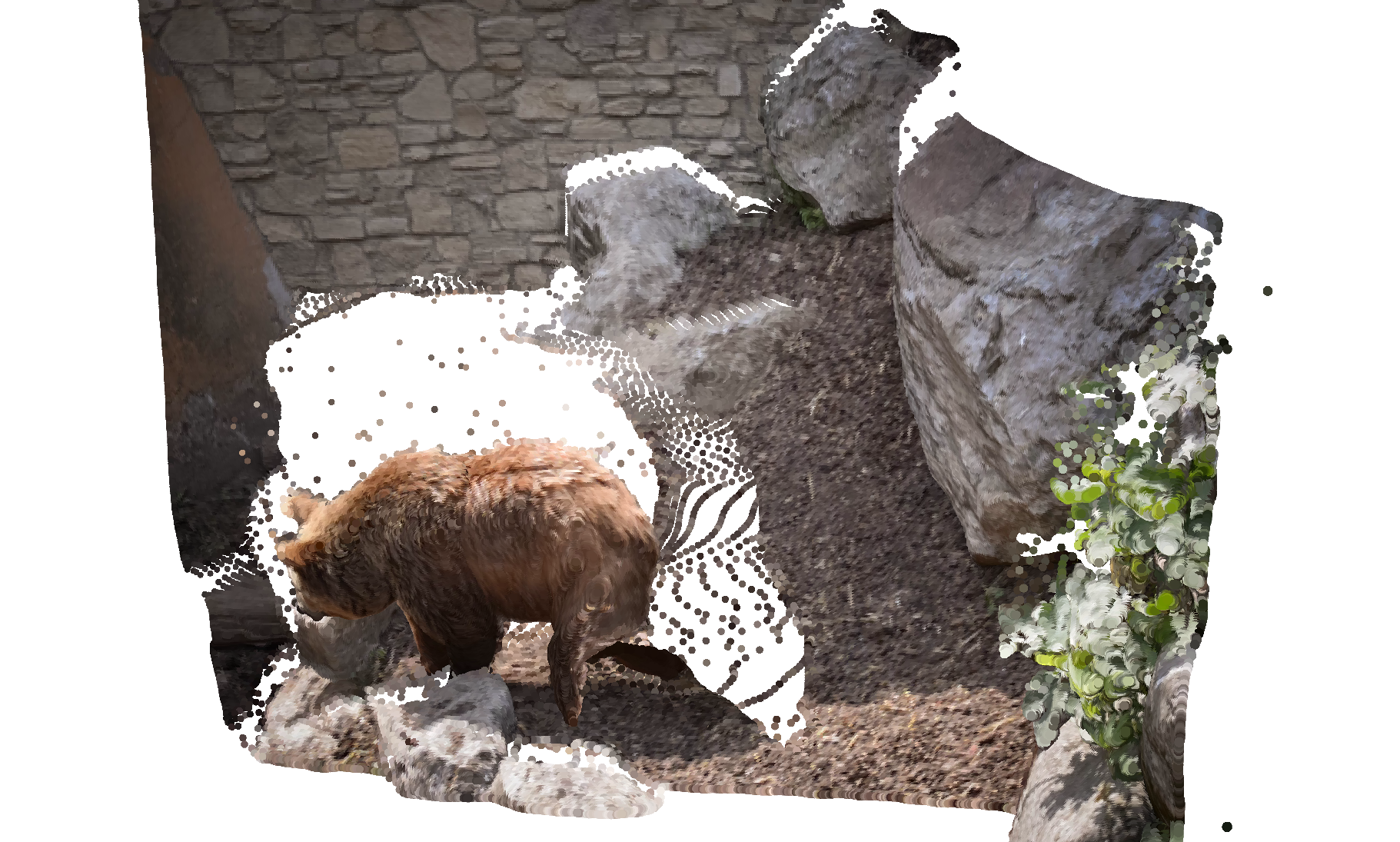}
    }
    \\[-10pt]

    \inputpair
    {
        \includegraphics[width=\linewidth]{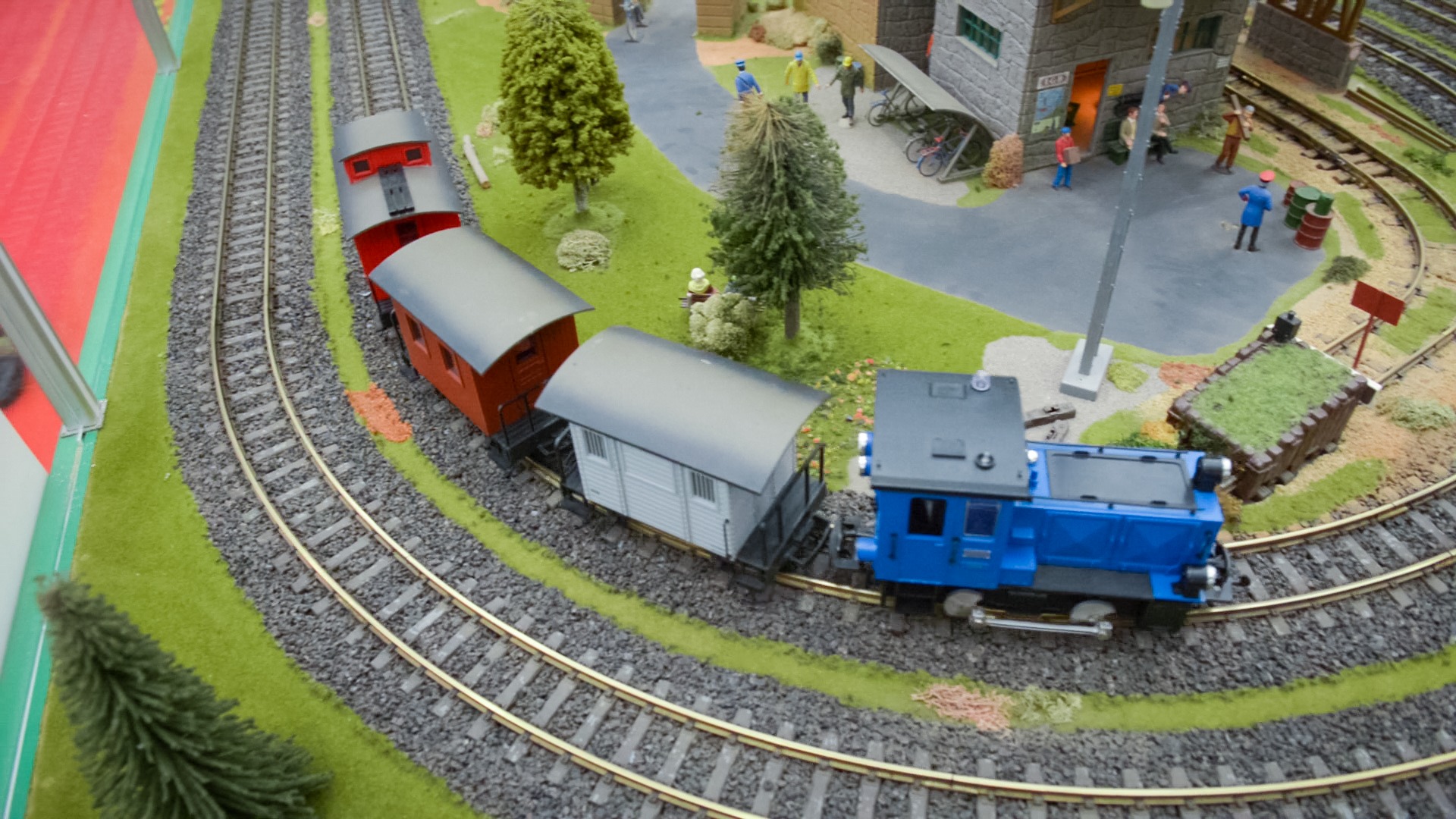}
    }
    {
        \includegraphics[width=\linewidth]{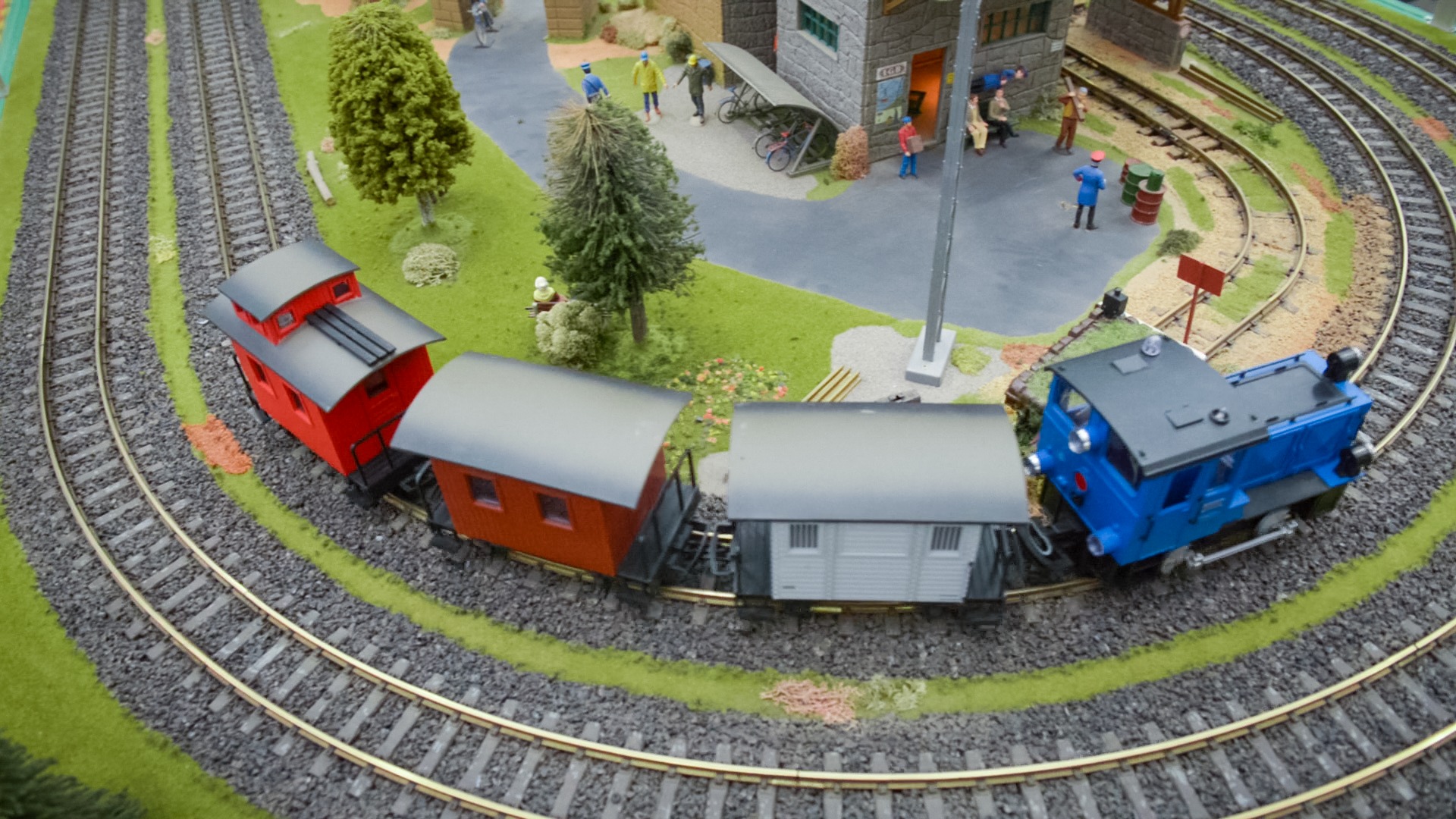}
    }
    &
    \resultcell{
        \includegraphics[width=\linewidth]{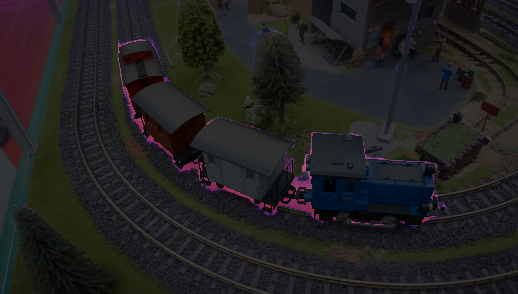}
    }
    &
    \resultcell{
        \includegraphics[width=\linewidth]{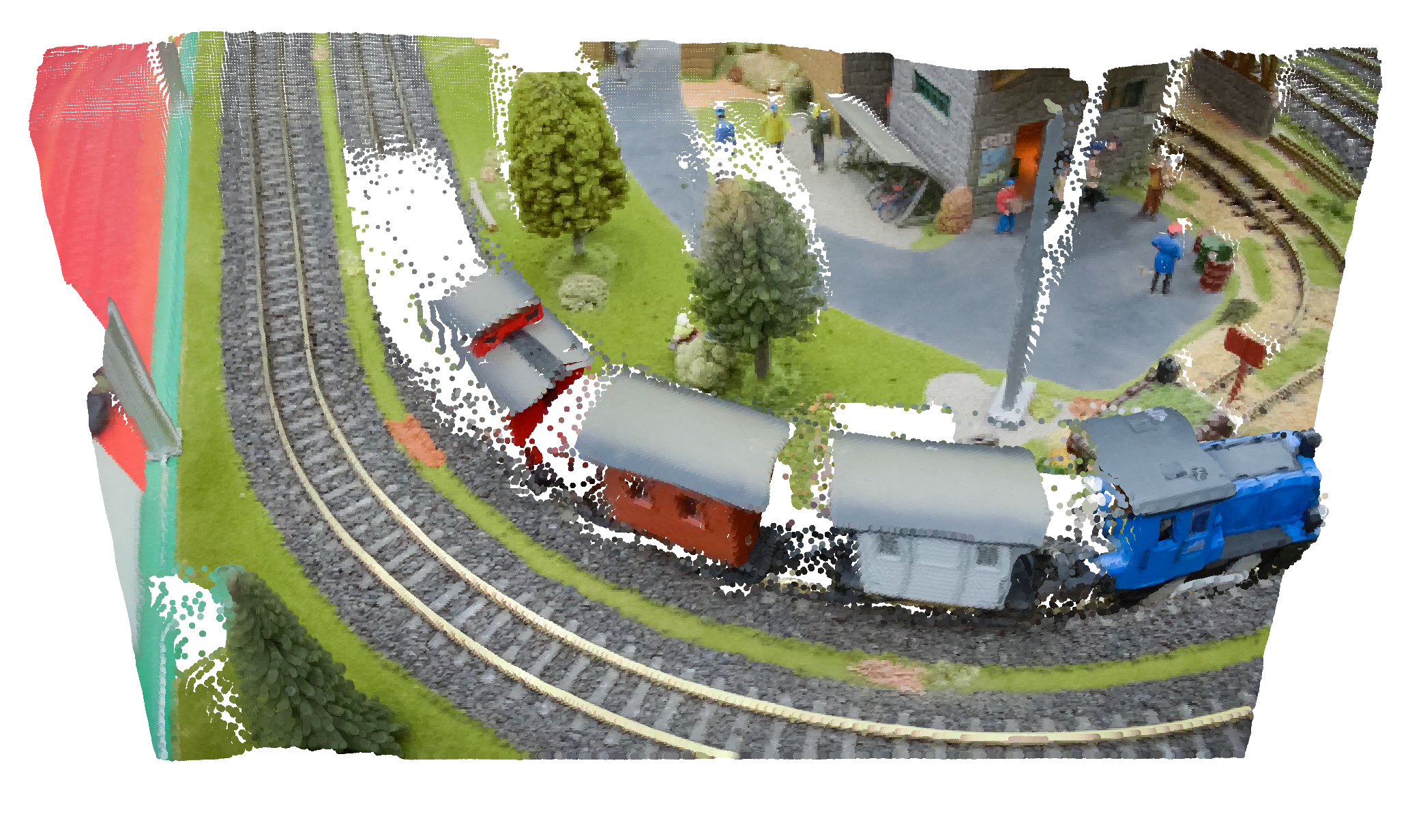}
    }
    &
    \resultcell{
        \includegraphics[width=\linewidth]{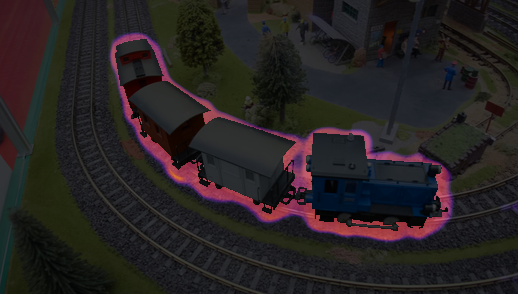}
    }
    &
    \resultcell{
        \includegraphics[width=\linewidth]{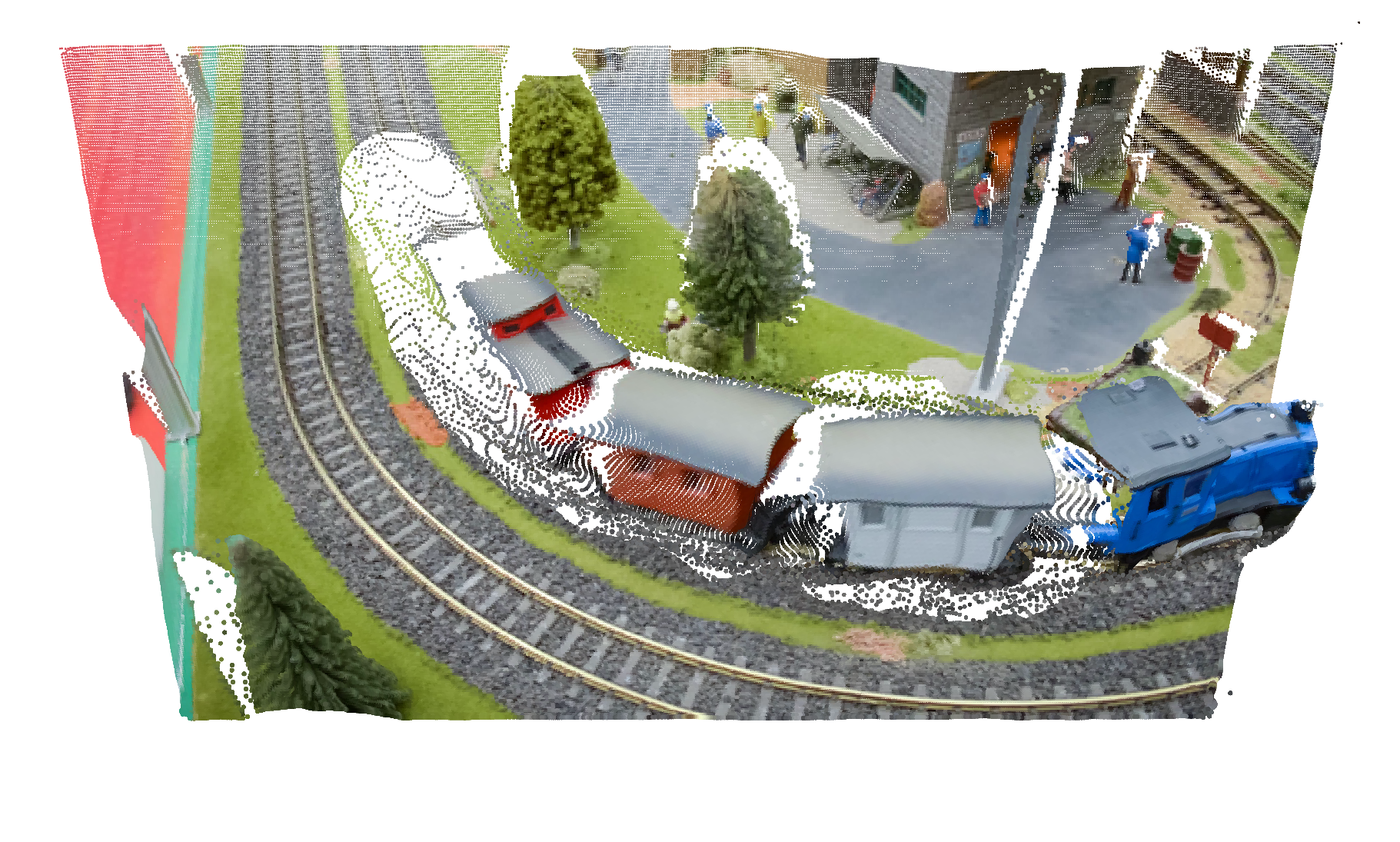}
    }
    \\ [-5pt]

    \end{tabular}
    \captionsetup{skip=-4pt}
    \caption{
        \textbf{Displacement Map and Point Cloud.} 
        Points are mean-normalized to unit scale before computing accumulated motions.
    }
    \label{fig:dispmap_visualization}
\vspace{-5pt}
\end{figure*}
\begin{table*}[tb!]
\centering
\captionsetup{skip=2pt}
\caption{\textbf{Ablation Study on Different Modules.} We report 3D tracking on 64 evaluated frames on DR and ADT as All / Dynamic groups, and video depth estimation on 110 evaluated frames with 49-frame inference chunks on Bonn.
$\dagger$ denotes discrete time embedding.
}
\label{tab:ablation}

\resizebox{\textwidth}{!}{%

\setlength{\tabcolsep}{3pt}
\renewcommand{\arraystretch}{1.1}

\begin{tabular}{@{}l c c cccc cc@{}}
\toprule

\multirow{3}{*}{\textbf{Item}}
& \multirow{3}{*}{\textbf{Adaptation}}
& \multirow{3}{*}{\shortstack{\textbf{Params.}\\\textbf{(B)}}}
& \multicolumn{4}{c}{\textbf{Tracking}}
& \multicolumn{2}{c}{\textbf{Video Depth}} \\
\cmidrule(lr){4-7}
\cmidrule(lr){8-9}
& 
&
& \multicolumn{2}{c}{\textbf{DR}}
& \multicolumn{2}{c}{\textbf{ADT}}
& \multicolumn{2}{c}{\textbf{Bonn}} \\
\cmidrule(lr){4-5}
\cmidrule(lr){6-7}
\cmidrule(lr){8-9}
&
&
& \shortstack{APD $\uparrow$\\{\footnotesize All / Dyn.}}
& \shortstack{EPE $\downarrow$\\{\footnotesize All / Dyn.}}
& \shortstack{APD $\uparrow$\\{\footnotesize All / Dyn.}}
& \shortstack{EPE $\downarrow$\\{\footnotesize All / Dyn.}}
& \shortstack{AbsRel $\downarrow$\\{\footnotesize \phantom{All / Dyn.}}}
& \shortstack{$\delta_{<1.25}$ $\uparrow$\\{\footnotesize \phantom{All / Dyn.}}} \\

\midrule

(a) & $\Delta W_\mathrm{4D}^{512\dagger}$ & 0.84
& 83.51 / 91.26
& 0.186 / 0.101
& 74.16 / 91.37
& 0.358 / 0.101
& 0.074
& 97.13 \\

\midrule

(b) & $\Delta W_\mathrm{4D}^{512}$ & 0.84
& 83.24 / 91.31
& 0.186 / 0.098
& 77.54 / 92.19
& 0.312 / 0.077
& 0.074
& 97.14 \\

(c) & $\Delta W_\mathrm{4D}^{1024}$ & 1.54
& 82.80 / \textbf{91.54}
& 0.195 / \textbf{0.096}
& 76.83 / 92.83
& 0.325 / 0.077
& 0.078
& 96.96 \\

(d) & $W_\mathrm{DiT}$ & 1.53
& 81.85 / 90.63
& 0.207 / 0.105
& 76.34 / 92.37
& 0.328 / 0.079
& 0.081
& 96.62 \\

(e) & $\Delta W_\mathrm{Dep}^{512}$ & 0.84
& 84.23 / 91.24
& 0.172 / 0.098
& 77.92 / 91.79
& 0.295 / 0.082
& 0.071
& 97.22 \\

(f) & $\Delta W_\mathrm{Dep}^{512\;\frozen} + \Delta W_\mathrm{4D}^{512}$ & 0.84
& \textbf{84.28} / 91.33
& \textbf{0.168} / 0.098
& \textbf{79.99} / \textbf{92.90}
& \textbf{0.273} / \textbf{0.075}
& \textbf{0.069}
& \textbf{97.28} \\

\bottomrule
\end{tabular}
}
{
    \footnotesize
    *
    $\Delta W_\mathrm{\{{4D, Dep}\}}^r$ = Rank-$r$ \{4D, Depth\}-LoRA fine-tuning.
    $W_\mathrm{DiT}$ = Full DiT fine-tuning. 
    \frozen\ = Frozen.
}
\vspace{-13pt}
\end{table*}

\heading{Displacement Map}
To further analyze motion leakage around foreground boundaries, we compare the background motion of our method and 4RC. 
For both methods, we mean-normalize the query-frame points, compute their accumulated 3D displacement, and colorize the displacement magnitudes over a shared range of $[0, 0.5]$.
We retain only the background region, where non-zero displacement indicates erroneous motion.
As shown in \Cref{fig:dispmap_visualization}, our method exhibits less background motion leakage and a clearer motion boundary, consistent with the corresponding point-cloud visualization.

\heading{Ablation on Adaptation}
Rows (b)--(d) of \Cref{tab:ablation} ablate different adaptation strategies.
Both (c) \& (d) nearly double the trainable parameters.
However, using a wider LoRA does not guarantee overall better performance, 
while full DiT fine-tuning results in degraded performance on several metrics despite using more trainable parameters

\heading{Importance of geometry prior}
Another important aspect in our method is the incorporation of the geometry prior.
To measure its impact,
we compare different variants of our method, with and without the geometry prior.
Comparing (b) and (e) in \Cref{tab:ablation},
the geometry prior clearly contributes to the performance along all metrics.
Combining $\Delta W_\mathrm{4D}$ and $\Delta W_\mathrm{dep}$ in (f) further improves the overall performance,
demonstrating the complementary benefits of both components.




\section{Conclusion}
\label{sec:conclusion}

We present \methodname,
a feed-forward 4D perceiver that processes monocular video input for full 4D reconstruction.
The key to its success is to leverage prior knowledge from pretrained video generators in a single-step manner,
which differs from traditional multi-step denoising-based 4D perceivers.
Our on-demand decoder enables flexible spatial and temporal querying, while a frozen video-depth LoRA provides a plug-in geometry prior without additional trainable parameters.
Crucially,
we demonstrate that \methodname achieves state-of-the-art and flexible 4D reconstruction,
with future improvement expected as video generative pretraining continues to advance.
We also demonstrate its capability to accurately capture dynamic motion and intricate details in real-world videos.
Together, these findings support video generators as a promising foundation for unified 4D reconstruction.

\section*{Acknowledgement}
This research is supported by the NTU SUG-NAP and National Research Foundation, Singapore, under its NRF Fellowship Award NRF-NRFF17-2025-0009.
We would like to acknowledge the High Performance Computing Centre of Nanyang Technological University Singapore, for providing the computing resources, facilities, and services that have contributed significantly to this work.
\newpage
\section*{AI Use Statement}
We used LLM-based AI tools to assist with grammar correction, language polishing, and adjustments to figure and table layouts in the manuscript. 
The authors reviewed and revised the AI-assisted text and verified its consistency with the methods and experimental results.
The authors assume full responsibility for the final manuscript, including its scientific claims and reported results.

\section*{Reproducibility Statement}
\Cref{sec:methods} describes the model architecture, the continuous-time query, the plug-in geometry prior, and the training objective; \Cref{sec:appendix_losses} gives the definition and weight of each loss term.
\Cref{sec:exp} reports the training data, model variants, and optimization settings, and \Cref{sec:appendix_dataset,sec:appendix_preprocess_and_sampling} detail the frame resolutions, dataset sampling probabilities, preprocessing, and query sampling, with the ADT sequences excluded from training listed in \Cref{tab:appendix_adt_test_sequence_exclusion}.
\Cref{sec:appendix_world_coord_tracking,sec:appendix_video_depth,sec:appendix_camera_pose} specify the evaluation protocols, metrics, alignment procedures, inference setups, and baseline configurations.
All training datasets are publicly available, and the Wan2.1 backbone~\citep{wan2025} and the depth LoRA~\citep{zhang2026dvd} have been publicly released.


\bibliographystyle{iclr2027_conference}
\bibliography{chuanxia_general,chuanxia_specific,iclr2027_conference}

@String(CVPR= {IEEE Conf. Comput. Vis. Pattern Recog.})

@String(ICCV= {Int. Conf. Comput. Vis.})

@String(ECCV= {Eur. Conf. Comput. Vis.})

@String(NIPS= {Adv. Neural Inform. Process. Syst.})

@String(ICLR = {Int. Conf. Learn. Represent.})

@String(IROS = {IEEE/RSJ Int. Conf. Intell. Robots Syst.})

@STRING{icml = "Proceedings of the International Conference on Machine Learning"}

@String(CVPR  = {CVPR})

@String(ICCV  = {ICCV})

@String(ECCV  = {ECCV})

@String(NIPS  = {NeurIPS})

@String(ICLR  = {ICLR})

@String(TMLR = {TMLR})

@article{doersch2022tap,
  title={{TAP}-Vid: A Benchmark for Tracking Any Point in a Video},
  author={Doersch, Carl and Gupta, Ankush and Markeeva, Larisa and Recasens, Adria and Smaira, Lucas and Aytar, Yusuf and Carreira, Joao and Zisserman, Andrew and Yang, Yi},
  journal=NIPS,
  volume={35},
  pages={13610--13626},
  year={2022}
}

@article{oquabdinov2,
  title={DINOv2: Learning Robust Visual Features without Supervision},
  author={Oquab, Maxime and Darcet, Timoth{\'e}e and Moutakanni, Th{\'e}o and Vo, Huy V and Szafraniec, Marc and Khalidov, Vasil and Fernandez, Pierre and Haziza, Daniel and Massa, Francisco and El-Nouby, Alaaeldin and others},
  journal=TMLR,
  year={2023}
}

@InProceedings{karaev23dynamicstereo,
  author     = {Nikita Karaev and Ignacio Rocco and Benjamin Graham and
               Natalia Neverova and Andrea Vedaldi and Christian Rupprecht},
  booktitle  = CVPR,
  title      = {{DynamicStereo}: Consistent Dynamic Depth from Stereo
               Videos},
  year       = {2023}
}

@inproceedings{zheng2023pointodyssey,
  title={Pointodyssey: A large-scale synthetic dataset for long-term point tracking},
  author={Zheng, Yang and Harley, Adam W and Shen, Bokui and Wetzstein, Gordon and Guibas, Leonidas J},
  booktitle=ICCV,
  pages={19855--19865},
  year={2023}
}

@article{hedlin2023unsupervised,
title={Unsupervised Semantic Correspondence Using Stable Diffusion},
author={Eric Hedlin and Gopal Sharma and Shweta Mahajan and Hossam Isack and Abhishek Kar and Andrea Tagliasacchi and Kwang Moo Yi},
journal=NIPS,
year={2023},
url={https://openreview.net/forum?id=sovxUzPzLN}
}

@inproceedings{ke2024repurposing,
  title={Repurposing diffusion-based image generators for monocular depth estimation},
  author={Ke, Bingxin and Obukhov, Anton and Huang, Shengyu and Metzger, Nando and Daudt, Rodrigo Caye and Schindler, Konrad},
  booktitle=CVPR,
  pages={9492--9502},
  year={2024}
}

@inproceedings{wang2024dust3r,
  title={Dust3r: Geometric 3d vision made easy},
  author={Wang, Shuzhe and Leroy, Vincent and Cabon, Yohann and Chidlovskii, Boris and Revaud, Jerome},
  booktitle=CVPR,
  pages={20697--20709},
  year={2024}
}

@inproceedings{karaev2024cotracker,
  title={Cotracker: It is better to track together},
  author={Karaev, Nikita and Rocco, Ignacio and Graham, Benjamin and Neverova, Natalia and Vedaldi, Andrea and Rupprecht, Christian},
  booktitle=ECCV,
  pages={18--35},
  year={2024},
  organization={Springer}
}

@inproceedings{karaev2025cotracker3,
  title={Cotracker3: Simpler and better point tracking by pseudo-labelling real videos},
  author={Karaev, Nikita and Makarov, Yuri and Wang, Jianyuan and Neverova, Natalia and Vedaldi, Andrea and Rupprecht, Christian},
  booktitle=ICCV,
  pages={6013--6022},
  year={2025}
}

@inproceedings{lu2025matrix3d,
  title={Matrix3D: Large Photogrammetry Model All-in-One},
  author={Lu, Yuanxun and Zhang, Jingyang and Fang, Tian and Nahmias, Jean-Daniel and Tsin, Yanghai and Quan, Long and Cao, Xun and Yao, Yao and Li, Shiwei},
  booktitle=CVPR,
  year={2025}
}

@inproceedings{zhang2024world,
  title={World-consistent Video Diffusion with Explicit 3D Modeling},
  author={Zhang, Qihang and Zhai, Shuangfei and Bautista, Miguel Angel and Miao, Kevin and Toshev, Alexander and Susskind, Joshua and Gu, Jiatao},
  booktitle=CVPR,
  year={2025}
}

@InProceedings{wang2025continuous,
  title={Continuous 3D Perception Model with Persistent State},
  author={Wang, Qianqian and Zhang, Yifei and Holynski, Aleksander and Efros, Alexei A and Kanazawa, Angjoo},
  booktitle=CVPR,
  month={June},
  year={2025},
}

@inproceedings{wang2025vggt,
  title={Vggt: Visual geometry grounded transformer},
  author={Wang, Jianyuan and Chen, Minghao and Karaev, Nikita and Vedaldi, Andrea and Rupprecht, Christian and Novotny, David},
  booktitle=CVPR,
  pages={5294--5306},
  year={2025}
}

@inproceedings{xu2025geometrycrafter,
      title={GeometryCrafter: Consistent Geometry Estimation for Open-world Videos with Diffusion Priors}, 
      author={Tian-Xing Xu and Xiangjun Gao and Wenbo Hu and Xiaoyu Li and Song-Hai Zhang and Ying Shan},
      booktitle   = ICCV,
      month     = {October},
      year      = {2025}
}

@inproceedings{wang2025pi,
  title={{$\pi^3$}: Permutation-Equivariant Visual Geometry Learning},
  author={Wang, Yifan and Zhou, Jianjun and Zhu, Haoyi and Chang, Wenzheng and Zhou, Yang and Li, Zizun and Chen, Junyi and Pang, Jiangmiao and Shen, Chunhua and He, Tong},
  booktitle=ICLR,
  year={2026}
}

@inproceedings{zhang2026d4rt,
  title={Efficiently Reconstructing Dynamic Scenes One D4RT at a Time},
  author={Zhang, Chuhan and Le Moing, Guillaume and Koppula, Skanda and Rocco, Ignacio and Momeni, Liliane and Xie, Junyu and Sun, Shuyang and Sukthankar, Rahul and Barral, Jo{\"e}lle K. and Hadsell, Raia and Ghahramani, Zoubin and Zisserman, Andrew and Zhang, Junlin and Sajjadi, Mehdi S. M.},
  booktitle=CVPR,
  year={2026}
}

@inproceedings{wang2026vggt,
  title={VGGT-Omega},
  author={Wang, Jianyuan and Chen, Minghao and Zhang, Shangzhan and Karaev, Nikita and Sch{\"o}nberger, Johannes and Labatut, Patrick and Bojanowski, Piotr and Novotny, David and Vedaldi, Andrea and Rupprecht, Christian},
  booktitle=CVPR,
  year={2026}
}

@article{wan2025,
      title={Wan: Open and Advanced Large-Scale Video Generative Models}, 
      author={Team Wan and Ang Wang and Baole Ai and Bin Wen and Chaojie Mao and Chen-Wei Xie and Di Chen and Feiwu Yu and Haiming Zhao and Jianxiao Yang and Jianyuan Zeng and Jiayu Wang and Jingfeng Zhang and Jingren Zhou and Jinkai Wang and Jixuan Chen and Kai Zhu and Kang Zhao and Keyu Yan and Lianghua Huang and Mengyang Feng and Ningyi Zhang and Pandeng Li and Pingyu Wu and Ruihang Chu and Ruili Feng and Shiwei Zhang and Siyang Sun and Tao Fang and Tianxing Wang and Tianyi Gui and Tingyu Weng and Tong Shen and Wei Lin and Wei Wang and Wei Wang and Wenmeng Zhou and Wente Wang and Wenting Shen and Wenyuan Yu and Xianzhong Shi and Xiaoming Huang and Xin Xu and Yan Kou and Yangyu Lv and Yifei Li and Yijing Liu and Yiming Wang and Yingya Zhang and Yitong Huang and Yong Li and You Wu and Yu Liu and Yulin Pan and Yun Zheng and Yuntao Hong and Yupeng Shi and Yutong Feng and Zeyinzi Jiang and Zhen Han and Zhi-Fan Wu and Ziyu Liu},
      journal = {arXiv preprint arXiv:2503.20314},
      year={2025}
}

@misc{siméoni2025dinov3,
      title={DINOv3}, 
      author={Oriane Siméoni and Huy V. Vo and Maximilian Seitzer and Federico Baldassarre and Maxime Oquab and Cijo Jose and Vasil Khalidov and Marc Szafraniec and Seungeun Yi and Michaël Ramamonjisoa and Francisco Massa and Daniel Haziza and Luca Wehrstedt and Jianyuan Wang and Timothée Darcet and Théo Moutakanni and Leonel Sentana and Claire Roberts and Andrea Vedaldi and Jamie Tolan and John Brandt and Camille Couprie and Julien Mairal and Hervé Jégou and Patrick Labatut and Piotr Bojanowski},
      year={2025},
      eprint={2508.10104},
      archivePrefix={arXiv},
      primaryClass={cs.CV},
      url={https://arxiv.org/abs/2508.10104}, 
}

@misc{bytedance2026seedance,
  author       = {{ByteDance Seed Team}},
  title        = {One-take Creation, Flexible Referencing: Introducing {Seedance} 2.5},
  year         = {2026},
  month        = {July},
  howpublished = {\url{https://seed.bytedance.com/en/blog/one-take-creation-flexible-referencing-introducing-seedance-2-5}},
  note         = {Accessed: 2026-09-26}
}

@inproceedings{jiang2026syn4d,
  title={Syn4D: A Multiview Synthetic 4D Dataset},
  author={Jiang, Zeren and Lan, Yushi and Luo, Yihang and Deng, Yufan and Lai, Zihang and Sucar, Edgar and Rupprecht, Christian and Laina, Iro and Larlus, Diane and Zheng, Chuanxia and others},
  booktitle=ECCV,
  year={2026}
}

@inproceedings{zhu2026motioncrafter,
  title={MotionCrafter: Dense Geometry and Motion Reconstruction with a 4D VAE},
  author={Zhu, Ruijie and Lu, Jiahao and Hu, Wenbo and Han, Xiaoguang and Cai, Jianfei and Shan, Ying and Zheng, Chuanxia},
  booktitle=CVPR,
  year={2026}
}

@inproceedings{jiang2025geo4d,
      title={Geo4D: Leveraging Video Generators for Geometric 4D Scene Reconstruction}, 
      author={Zeren Jiang and Chuanxia Zheng and Iro Laina and Diane Larlus and Andrea Vedaldi},
      booktitle=ICCV,
      year={2025}
}

@inproceedings{zhan2024does,
  title={What Does Stable Diffusion Know about the 3D Scene?},
  author={Zhan, Guanqi and Zheng, Chuanxia and Xie, Weidi and Zisserman, Andrew},
  booktitle=NIPS,
  year={2024}
}

@InProceedings{wang2023videomaev2,
    author    = {Wang, Limin and Huang, Bingkun and Zhao, Zhiyu and Tong, Zhan and He, Yinan and Wang, Yi and Wang, Yali and Qiao, Yu},
    title     = {VideoMAE V2: Scaling Video Masked Autoencoders With Dual Masking},
    booktitle = {Proceedings of the IEEE/CVF Conference on Computer Vision and Pattern Recognition (CVPR)},
    month     = {June},
    year      = {2023},
    pages     = {14549-14560}
}

@article{hu2021lora,
  title={Lora: Low-rank adaptation of large language models},
  author={Hu, Edward J and Shen, Yelong and Wallis, Phillip and Allen-Zhu, Zeyuan and Li, Yuanzhi and Wang, Shean and Wang, Lu and Chen, Weizhu},
  journal={arXiv preprint arXiv:2106.09685},
  year={2021}
}

@inproceedings{yu2025progressivelora,
  title={Progressive lora for multimodal continual instruction tuning},
  author={Yu, Yahan and Zhang, Duzhen and Ren, Yong and Zhao, Xuanle and Chen, Xiuyi and Chu, Chenhui},
  booktitle={Findings of the Association for Computational Linguistics: ACL 2025},
  pages={2779--2796},
  year={2025}
}

@inproceedings{he2025lotus,
  title={Lotus: Diffusion-based visual foundation model for high-quality dense prediction},
  author={He, Jing and Li, Haodong and Yin, Wei and Liang, Yixun and Li, Leheng and Zhou, Kaiqiang and Zhang, Hongbo and Liu, Bingbing and Chen, YingCong},
  booktitle={International Conference on Learning Representations},
  volume={2025},
  pages={89454--89467},
  year={2025}
}

@inproceedings{zhang2026dvd,
  title={Dvd: Deterministic video depth estimation with generative priors},
  author={Zhang, Hongfei and Chen, Harold Haodong and Liao, Chenfei and He, Jing and Zhang, Zixin and Li, Haodong and Liang, Yihao and Chen, Kanghao and Ren, Bin and Zheng, Xu and others},
  booktitle={ICML 2026 Workshop on Foundations of Deep Generative Models: Understanding Memorization, Generalization, and Reasoning},
  year={2026}
}

@inproceedings{huang2026vidfm,
  title={How Much 3D Do Video Foundation Models Encode?},
  author={Huang, Zixuan and Li, Xiang and Lv, Zhaoyang and Rehg, James M},
  booktitle={Proceedings of the IEEE/CVF Conference on Computer Vision and Pattern Recognition},
  pages={384--394},
  year={2026}
}

@article{wang2026genception,
  title={Video Generation Models are General-Purpose Vision Learners},
  author={Wang, Letian and Zhang, Chuhan and Kabra, Rishabh and Uijlings, Jasper and Waslander, Steven and Zisserman, Andrew and Carreira, Joao and He, Kaiming and Andriluka, Misha and Bazavan, Eduard Gabriel and others},
  journal={arXiv preprint arXiv:2607.09024},
  year={2026}
}

@inproceedings{hu2025depthcrafter,
  title={{DepthCrafter}: Generating Consistent Long Depth Sequences for Open-world Videos},
  author={Hu, Wenbo and Gao, Xiangjun and Li, Xiaoyu and Zhao, Sijie and Cun, Xiaodong and Zhang, Yong and Quan, Long and Shan, Ying},
  booktitle={Proceedings of the IEEE/CVF Conference on Computer Vision and Pattern Recognition (CVPR)},
  pages={2005--2015},
  year={2025}
}

@article{son2025ditracker,
  title={Probing and Leveraging Video Diffusion Transformer Features for Robust Point Tracking},
  author={Son, Soowon and An, Honggyu and Nam, Jisu and Ko, Hyunah and Kim, Chaehyun and Chung, Dahyun and Jin, Siyoon and Yi, Jung and Hur, Junhwa and Kim, Seungryong},
  journal={arXiv preprint arXiv:2512.20606},
  year={2025}
}

@article{nam2026trackcraft3r,
  title={{TrackCraft3R}: Repurposing Video Diffusion Transformers for Dense {3D} Tracking},
  author={Nam, Jisu and Koo, Jahyeok and Son, Soowon and Jung, Jaewoo and An, Honggyu and Hur, Junhwa and Kim, Seungryong},
  journal={arXiv preprint arXiv:2605.12587},
  year={2026}
}

@inproceedings{zhang2025monst3r,
  title={Monst3r: A simple approach for estimating geometry in the presence of motion},
  author={Zhang, Junyi and Herrmann, Charles and Hur, Junhwa and Jampani, Varun and Darrell, Trevor and Cole, Forrester and Sun, Deqing and Yang, Ming-Hsuan},
  booktitle={International Conference on Learning Representations},
  volume={2025},
  pages={82863--82886},
  year={2025}
}

@inproceedings{wang2025spann3r,
title={3d reconstruction with spatial memory},
author={Wang, Hengyi and Agapito, Lourdes},
booktitle={2025 International Conference on 3D Vision (3DV)},
pages={78--89},
year={2025},
organization={IEEE}
}

@inproceedings{xiao2024spatialtracker,
  title={Spatialtracker: Tracking any 2d pixels in 3d space},
  author={Xiao, Yuxi and Wang, Qianqian and Zhang, Shangzhan and Xue, Nan and Peng, Sida and Shen, Yujun and Zhou, Xiaowei},
  booktitle={2024 IEEE/CVF Conference on Computer Vision and Pattern Recognition (CVPR)},
  pages={20406--20417},
  year={2024},
  organization={IEEE}
}

@inproceedings{xiao2025spatialtrackerv2,
  title={Spatialtrackerv2: Advancing 3d point tracking with explicit camera motion},
  author={Xiao, Yuxi and Wang, Jianyuan and Xue, Nan and Karaev, Nikita and Makarov, Yuri and Kang, Bingyi and Zhu, Xing and Bao, Hujun and Shen, Yujun and Zhou, Xiaowei},
  booktitle={2025 IEEE/CVF International Conference on Computer Vision (ICCV)},
  pages={6726--6737},
  year={2025},
  organization={IEEE}
}

@inproceedings{feng2025st4rtrack,
  title={St4rtrack: Simultaneous 4d reconstruction and tracking in the world},
  author={Feng, Haiwen and Zhang, Junyi and Wang, Qianqian and Ye, Yufei and Yu, Pengcheng and Black, Michael J and Darrell, Trevor and Kanazawa, Angjoo},
  booktitle={2025 IEEE/CVF International Conference on Computer Vision (ICCV)},
  pages={8503--8513},
  year={2025},
  organization={IEEE}
}

@inproceedings{liu2026trace,
  title={Trace anything: Representing any video in 4d via trajectory fields},
  author={Liu, Xinhang and Xiao, Yuxi and Chen, Donny Y and Feng, Jiashi and Tai, Yu-Wing and Tang, Chi-Keung and Kang, Bingyi},
  booktitle={International Conference on Learning Representations},
  volume={2026},
  pages={37738--37767},
  year={2026}
}

@inproceedings{karhade2026any4d,
  title={Any4d: Unified feed-forward metric 4d reconstruction},
  author={Karhade, Jay and Keetha, Nikhil and Zhang, Yuchen and Gupta, Tanisha and Sharma, Akash and Scherer, Sebastian and Ramanan, Deva},
  booktitle={Proceedings of the IEEE/CVF Conference on Computer Vision and Pattern Recognition},
  pages={14578--14589},
  year={2026}
}

@inproceedings{sucar2026vdpm,
  title={{V-DPM}: 4D Video Reconstruction with Dynamic Point Maps},
  author={Sucar, Edgar and Insafutdinov, Eldar and Lai, Zihang and Vedaldi, Andrea},
  booktitle={Proceedings of the IEEE/CVF Conference on Computer Vision and Pattern Recognition (CVPR)},
  year={2026}
}

@inproceedings{luo20264rc,
  title={4RC: 4D Reconstruction via Conditional Querying Anytime and Anywhere},
  author={Yihang Luo and Shangchen Zhou and Yushi Lan and Xingang Pan and Chen Change Loy},
  booktitle={ICML},
  year={2026}
}

@misc{li2026opend4rt,
  author       = {Li, Jiaxin and RHOS Team},
  title        = {{OpenD4RT}: An Unofficial PyTorch/GPU Implementation of {D4RT}},
  year         = {2026},
  howpublished = {\url{https://github.com/Lijiaxin0111/Open-d4rt}},
  note         = {GitHub repository}
}

@inproceedings{keetha2026mapanything,
  title={Mapanything: Universal feed-forward metric 3d reconstruction},
  author={Keetha, Nikhil and M{\"u}ller, Norman and Sch{\"o}nberger, Johannes and Porzi, Lorenzo and Zhang, Yuchen and Fischer, Tobias and Knapitsch, Arno and Zauss, Duncan and Weber, Ethan and Antunes, Nelson and others},
  booktitle={2026 International Conference on 3D Vision (3DV)},
  pages={499--509},
  year={2026},
  organization={IEEE}
}

@article{lin2025da3,
  title={Depth anything 3: Recovering the visual space from any views},
  author={Lin, Haotong and Chen, Sili and Liew, Junhao and Chen, Donny Y and Li, Zhenyu and Shi, Guang and Feng, Jiashi and Kang, Bingyi},
  journal={arXiv preprint arXiv:2511.10647},
  year={2025}
}

@article{lin2026sm4rt,
   title     = {SM4RT: Learning Structured Motion Geometry for 4D Reconstruction},
   author    = {Lin, Shing Ho J. and Zheng, Wenzhao and Zhuo, Dong and Wu, Yuqi and Zhou, Jie and Lu, Jiwen},
   journal   = {arXiv preprint arXiv:2607.22534},
   year      = {2026}
}

@article{chen2026uniquery4r,
  title={{UniQuery4R}: Unified {4D} Scene Reconstruction from a Single Query},
  author={Chen, Tiancheng and Tang, Sheng and Jin, Wenhua and Zhang, Weiqi and Fang, Juntong and Zhou, Junsheng and Li, Zesong},
  journal={arXiv preprint arXiv:2608.17283},
  year={2026}
}

@inproceedings{lai2026cowtracker,
  title={{CoWTracker}: Tracking by Warping instead of Correlation},
  author={Lai, Zihang and Insafutdinov, Eldar and Sucar, Edgar and Vedaldi, Andrea},
  booktitle={Proceedings of the IEEE/CVF Conference on Computer Vision and Pattern Recognition (CVPR)},
  year={2026}
}

@inproceedings{ngo2025delta,
  title={{DELTA}: Dense Efficient Long-range {3D} Tracking for Any Video},
  author={Ngo, Tuan Duc and Zhuang, Peiye and Gan, Chuang and Kalogerakis, Evangelos and Tulyakov, Sergey and Lee, Hsin-Ying and Wang, Chaoyang},
  booktitle={International Conference on Learning Representations (ICLR)},
  year={2025}
}

@article{ngo2025deltav2,
  title={DELTAv2: Accelerating dense 3D tracking},
  author={Ngo, Tuan Duc and Mirzaei, Ashkan and Qian, Guocheng and Liang, Hanwen and Gan, Chuang and Kalogerakis, Evangelos and Wonka, Peter and Wang, Chaoyang},
  journal={arXiv preprint arXiv:2508.01170},
  year={2025}
}

@article{qian2026flow4r,
  title={Flow4r: Unifying 4d reconstruction and tracking with scene flow},
  author={Qian, Shenhan and Zhang, Ganlin and Wu, Shangzhe and Cremers, Daniel},
  journal={arXiv preprint arXiv:2602.14021},
  year={2026}
}

@article{nam2025emergenttemporalcorrespondencesvideo,
  title={Emergent temporal correspondences from video diffusion transformers},
  author={Nam, Jisu and Son, Soowon and Chung, Dahyun and Kim, Jiyoung and Jin, Siyoon and Hur, Junhwa and Kim, Seungryong},
  journal={Advances in Neural Information Processing Systems},
  volume={38},
  pages={123683--123727},
  year={2026}
}

@inproceedings{ling2024dl3dv,
  title={Dl3dv-10k: A large-scale scene dataset for deep learning-based 3d vision},
  author={Ling, Lu and Sheng, Yichen and Tu, Zhi and Zhao, Wentian and Xin, Cheng and Wan, Kun and Yu, Lantao and Guo, Qianyu and Yu, Zixun and Lu, Yawen and others},
  booktitle={2024 IEEE/CVF Conference on Computer Vision and Pattern Recognition (CVPR)},
  pages={22160--22169},
  year={2024},
  organization={IEEE}
}

@inproceedings{yeshwanthliu2023scannetpp,
  title={ScanNet++: A High-Fidelity Dataset of 3D Indoor Scenes},
  author={Yeshwanth, Chandan and Liu, Yueh-Cheng and Nie{\ss}ner, Matthias and Dai, Angela},
  booktitle = {Proceedings of the International Conference on Computer Vision ({ICCV})},
  year={2023}
}

@article{cabon2020virtualkitti2,
  title={Virtual kitti 2},
  author={Cabon, Yohann and Murray, Naila and Humenberger, Martin},
  journal={arXiv preprint arXiv:2001.10773},
  year={2020}
}

@inproceedings{huang2018deepmvs,
  title={Deepmvs: Learning multi-view stereopsis},
  author={Huang, Po-Han and Matzen, Kevin and Kopf, Johannes and Ahuja, Narendra and Huang, Jia-Bin},
  booktitle={2018 IEEE/CVF Conference on Computer Vision and Pattern Recognition},
  pages={2821--2830},
  year={2018},
  organization={IEEE}
}

@inproceedings{mehl2023spring,
    title={Spring: A High-Resolution High-Detail Dataset and Benchmark for Scene Flow, Optical Flow and Stereo},
    author={Mehl, Lukas and Schmalfuss, Jenny and Jahedi, Azin and Nalivayko, Yaroslava and Bruhn, Andr{\'e}s},
    booktitle={Proc. IEEE/CVF Conference on Computer Vision and Pattern Recognition (CVPR)},
    pages={4981--4991},
    year={2023}
}

@inproceedings{sun2020waymo,
  title={Scalability in perception for autonomous driving: Waymo open dataset},
  author={Sun, Pei and Kretzschmar, Henrik and Dotiwalla, Xerxes and Chouard, Aurelien and Patnaik, Vijaysai and Tsui, Paul and Guo, James and Zhou, Yin and Chai, Yuning and Caine, Benjamin and others},
  booktitle={2020 IEEE/CVF Conference on Computer Vision and Pattern Recognition (CVPR)},
  pages={2443--2451},
  year={2020},
  organization={IEEE}
}

@article{loshchilov2017adamw,
  title={Decoupled weight decay regularization},
  author={Loshchilov, Ilya and Hutter, Frank},
  journal={arXiv preprint arXiv:1711.05101},
  year={2017}
}

@inproceedings{palazzolo2019bonn,
  title={ReFusion: 3D reconstruction in dynamic environments for RGB-D cameras exploiting residuals},
  author={Palazzolo, Emanuele and Behley, Jens and Lottes, Philipp and Giguere, Philippe and Stachniss, Cyrill},
  booktitle={2019 IEEE/RSJ International Conference on Intelligent Robots and Systems (IROS)},
  pages={7855--7862},
  year={2019},
  organization={IEEE}
}

@inproceedings{butler2012sintel,
  title={A naturalistic open source movie for optical flow evaluation},
  author={Butler, Daniel J and Wulff, Jonas and Stanley, Garrett B and Black, Michael J},
  booktitle={European conference on computer vision},
  pages={611--625},
  year={2012},
  organization={Springer}
}

@article{geiger2013kitti,
  title={Vision meets robotics: The kitti dataset},
  author={Geiger, Andreas and Lenz, Philip and Stiller, Christoph and Urtasun, Raquel},
  journal={The international journal of robotics research},
  volume={32},
  number={11},
  pages={1231--1237},
  year={2013},
  publisher={Sage Publications Sage UK: London, England}
}

@inproceedings{pan2023aria,
  title={Aria digital twin: A new benchmark dataset for egocentric 3d machine perception},
  author={Pan, Xiaqing and Charron, Nicholas and Yang, Yongqian and Peters, Scott and Whelan, Thomas and Kong, Chen and Parkhi, Omkar and Newcombe, Richard and Ren, Yuheng Carl},
  booktitle={2023 IEEE/CVF International Conference on Computer Vision (ICCV)},
  pages={20076--20086},
  year={2023},
  organization={IEEE}
}

@inproceedings{perazzi2016davis,
  title={A benchmark dataset and evaluation methodology for video object segmentation},
  author={Perazzi, Federico and Pont-Tuset, Jordi and McWilliams, Brian and Van Gool, Luc and Gross, Markus and Sorkine-Hornung, Alexander},
  booktitle={Proceedings of the IEEE conference on computer vision and pattern recognition},
  pages={724--732},
  year={2016}
}

@inproceedings{joo2015pstudio,
  title={Panoptic studio: A massively multiview system for social motion capture},
  author={Joo, Hanbyul and Liu, Hao and Tan, Lei and Gui, Lin and Nabbe, Bart and Matthews, Iain and Kanade, Takeo and Nobuhara, Shohei and Sheikh, Yaser},
  booktitle={Proceedings of the IEEE international conference on computer vision},
  pages={3334--3342},
  year={2015}
}

@inproceedings{sturm2012tumdynamic,
  title={A benchmark for the evaluation of RGB-D SLAM systems},
  author={Sturm, J{\"u}rgen and Engelhard, Nikolas and Endres, Felix and Burgard, Wolfram and Cremers, Daniel},
  booktitle={2012 IEEE/RSJ international conference on intelligent robots and systems},
  pages={573--580},
  year={2012},
  organization={IEEE}
}

@inproceedings{dai2017scannet,
  title={ScanNet: Richly-annotated 3D Reconstructions of Indoor Scenes},
  author={Dai, Angela and Chang, Angel X. and Savva, Manolis and Halber, Maciej and Funkhouser, Thomas and Nie{\ss}ner, Matthias},
  booktitle={Proc. Computer Vision and Pattern Recognition (CVPR), IEEE},
  year={2017}
}

@article{umeyama1991umeyama,
  title={Least-squares estimation of transformation parameters between two point patterns},
  author={Umeyama, Shinji},
  journal={IEEE Transactions on pattern analysis and machine intelligence},
  volume={13},
  number={4},
  pages={376--380},
  year={1991},
  publisher={IEEE}
}

@misc{jiang2026d4rtpytorch,
  author = {Jiang, Yurong},
  title = {{d4rt-pytorch}: PyTorch Implementation of {D4RT}},
  year = {2026},
  howpublished = {\url{https://github.com/jiangyurong609/d4rt-pytorch}},
  note = {GitHub repository}
}

\newpage
\appendix

\setcounter{table}{0}
\renewcommand{\thetable}{A\arabic{table}}
\counterwithin*{equation}{section}
\renewcommand{\theequation}{\thesection\arabic{equation}}
\newcommand{\doubletilde}[1]{\tilde{\tilde{#1}}}

\etocdepthtag.toc{appendix}
\etocsettagdepth{mainpaper}{none}
\etocsettagdepth{appendix}{subsection}
\etocsettocstyle{\section*{Appendix Contents}}{}
\tableofcontents
\newpage

\section{Additional Results}
\label{sec:append_qualitative_results}
\subsection{Additional Qualitative Results}

\begin{figure*}[htbp]
    \centering
    \setlength{\tabcolsep}{1pt}
    \renewcommand{\arraystretch}{0}
    \newcommand{\exampleblock}[4]{
        \begin{minipage}[t]{0.48\textwidth}
            \vspace{0pt}
            \centering

            \begin{minipage}[t]{0.333333\linewidth}
                \vspace{0pt}
                #1
            \end{minipage}%
            \begin{minipage}[t]{0.333333\linewidth}
                \vspace{0pt}
                #2
            \end{minipage}%
            \begin{minipage}[t]{0.333333\linewidth}
                \vspace{0pt}
                #3
            \end{minipage}

            \vspace{3pt}

            #4
        \end{minipage}
    }

    \newlength{\sideInputHeight}
    \setlength{\sideInputHeight}{0.075\textwidth}

    \newcommand{\sideexampleblock}[4]{%
        \begin{minipage}[t]{0.48\textwidth}
            \vspace{0pt}

            \hbox to \linewidth{%
                \begin{minipage}[t]{0.31\linewidth}
                    \vspace{0pt}
                    \centering
                    \vbox{%
                        \offinterlineskip
                        \hbox to \linewidth{\hfil#1\hfil}%
                        \hbox to \linewidth{\hfil#2\hfil}%
                        \hbox to \linewidth{\hfil#3\hfil}%
                    }
                \end{minipage}%
                \hspace{0.02\linewidth}%
                \begin{minipage}[t]{0.67\linewidth}
                    \vspace{0pt}
                    #4
                \end{minipage}%
            }%
        \end{minipage}%
    }

    \begin{tabular}{@{}cc@{}}

        \exampleblock
        {
            \includegraphics[width=\linewidth]{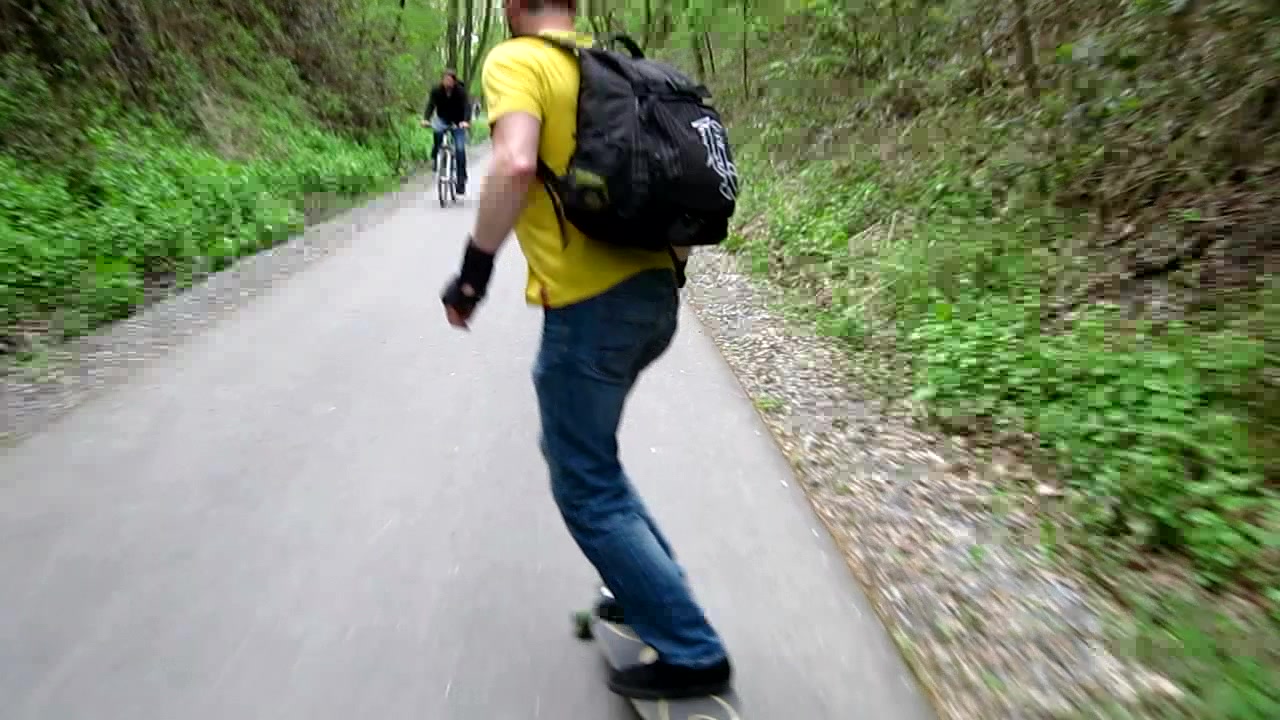}
        }
        {
            \includegraphics[width=\linewidth]{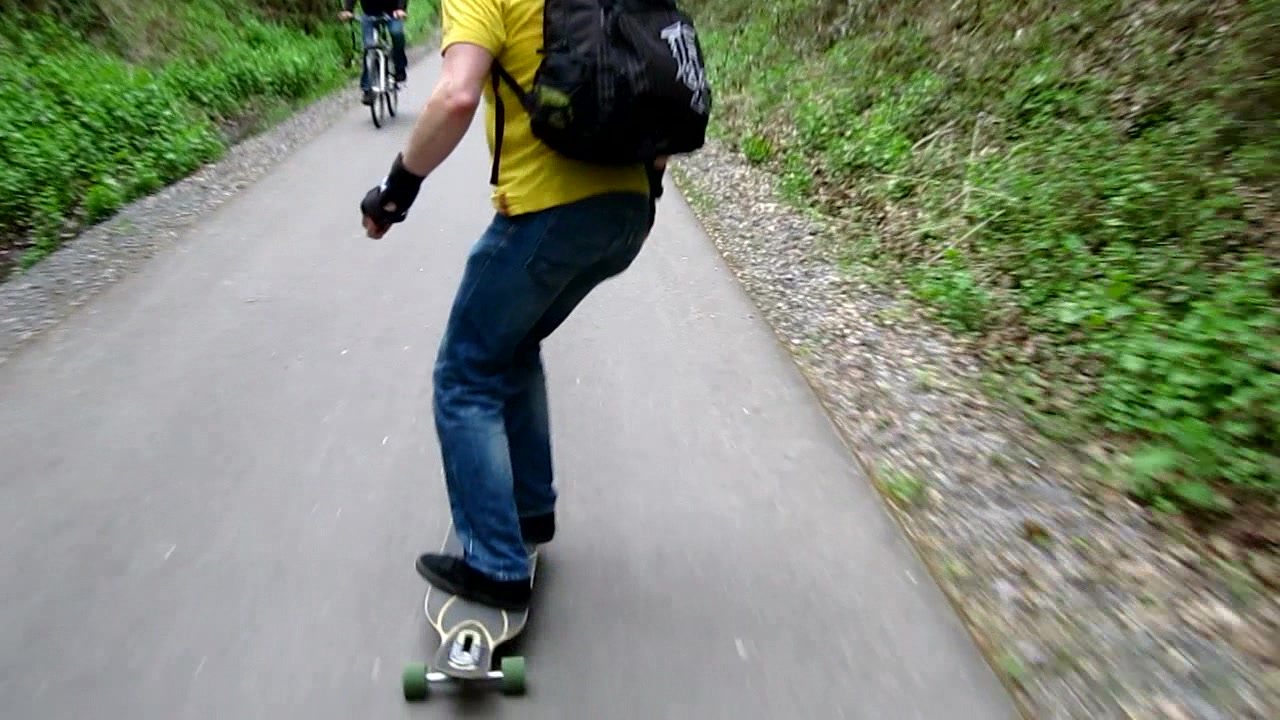}
        }
        {
            \includegraphics[width=\linewidth]{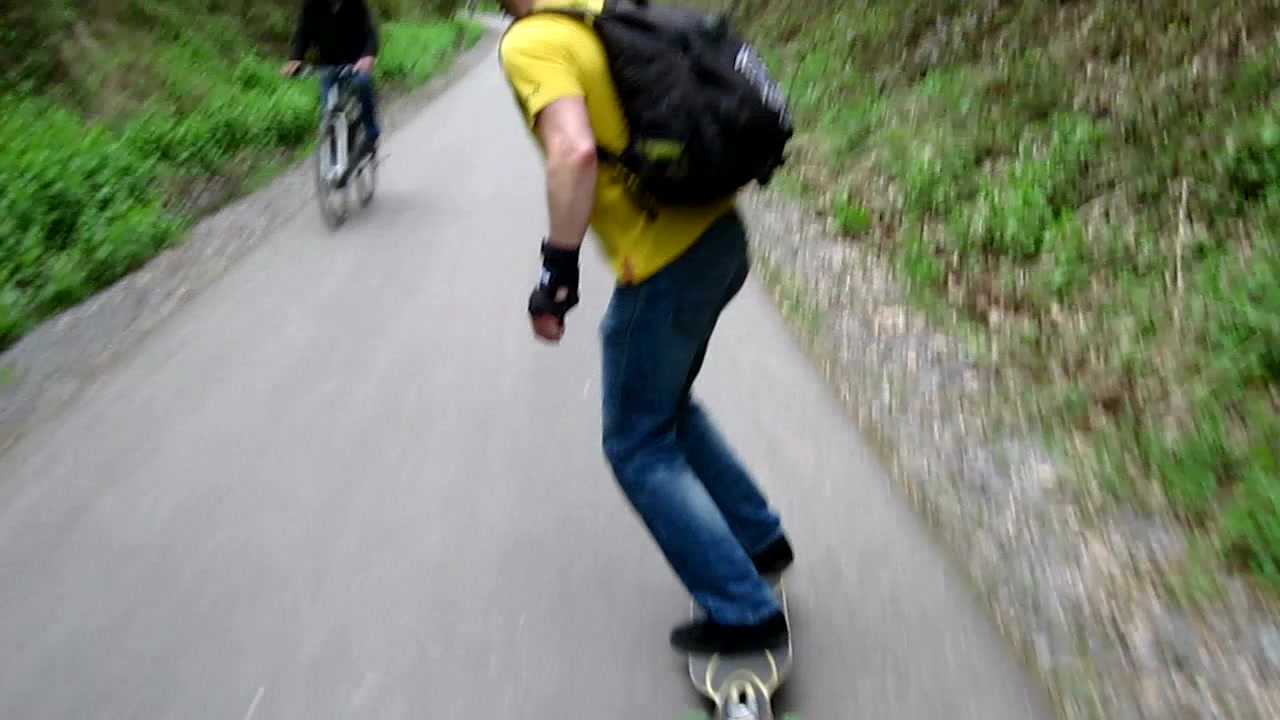}
        }
        {
            \includegraphics[width=\linewidth,trim=90mm 20mm 100mm 50mm,clip]{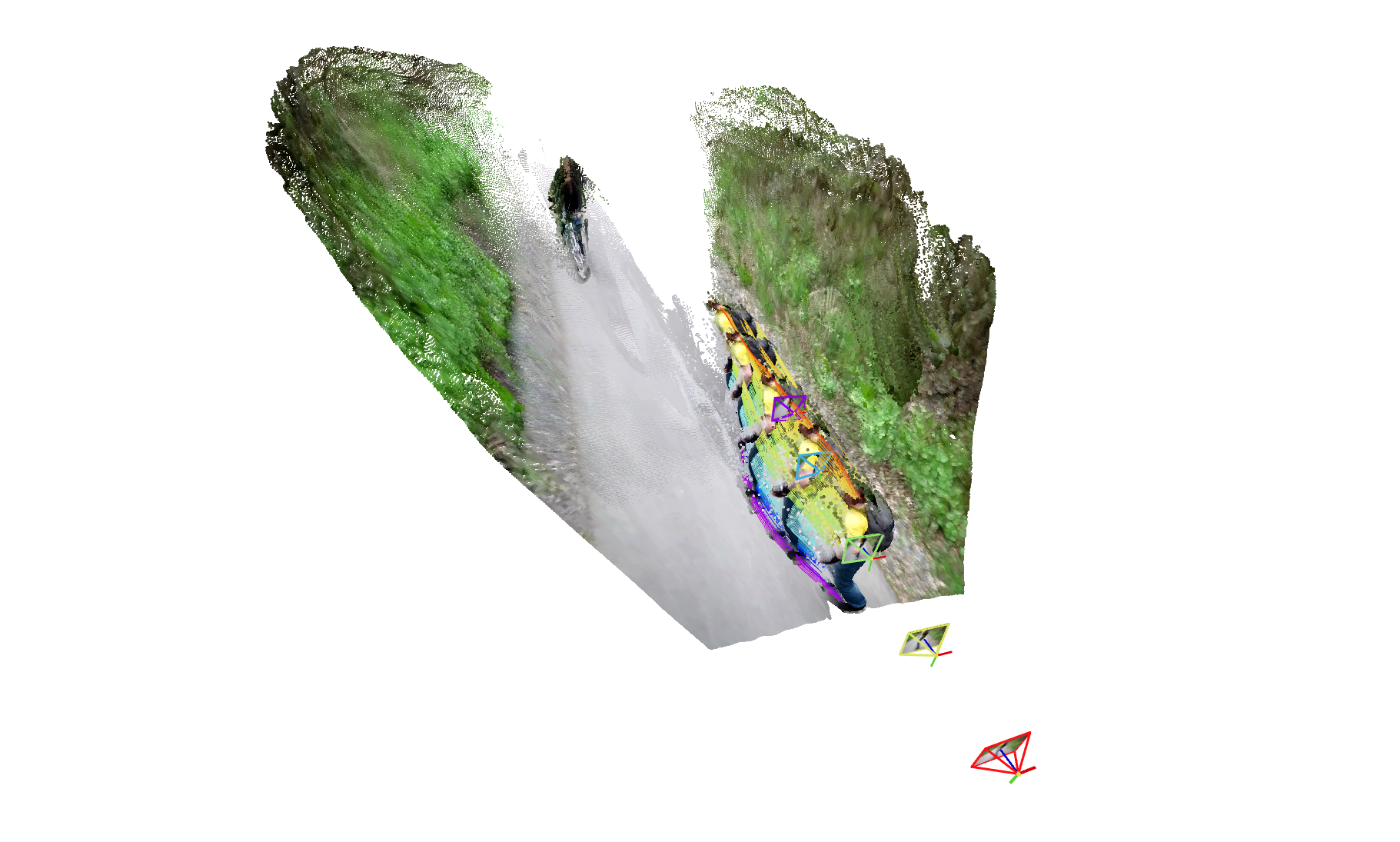}
        }

        &

        \exampleblock
        {
            \includegraphics[width=\linewidth]{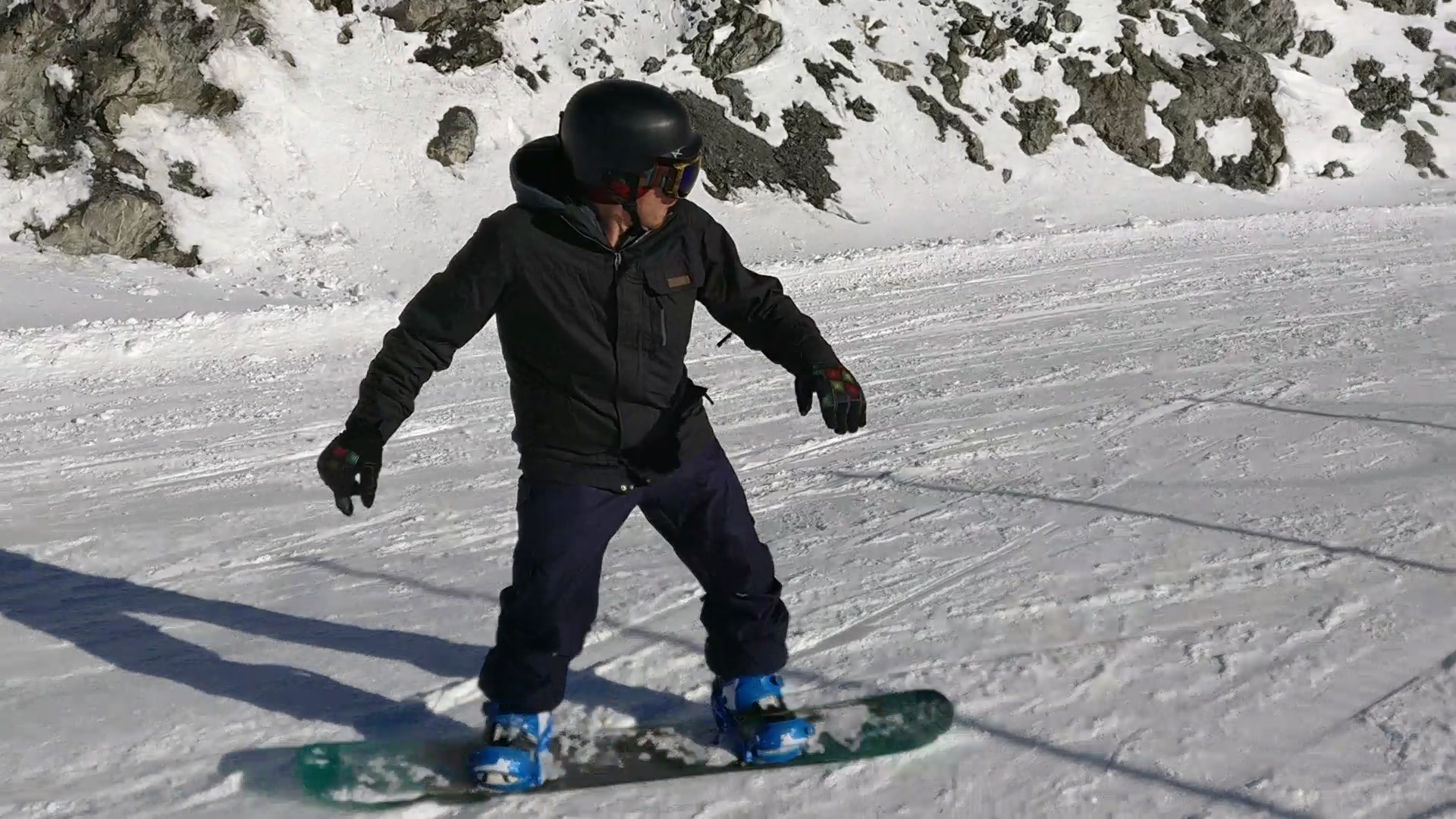}
        }
        {
            \includegraphics[width=\linewidth]{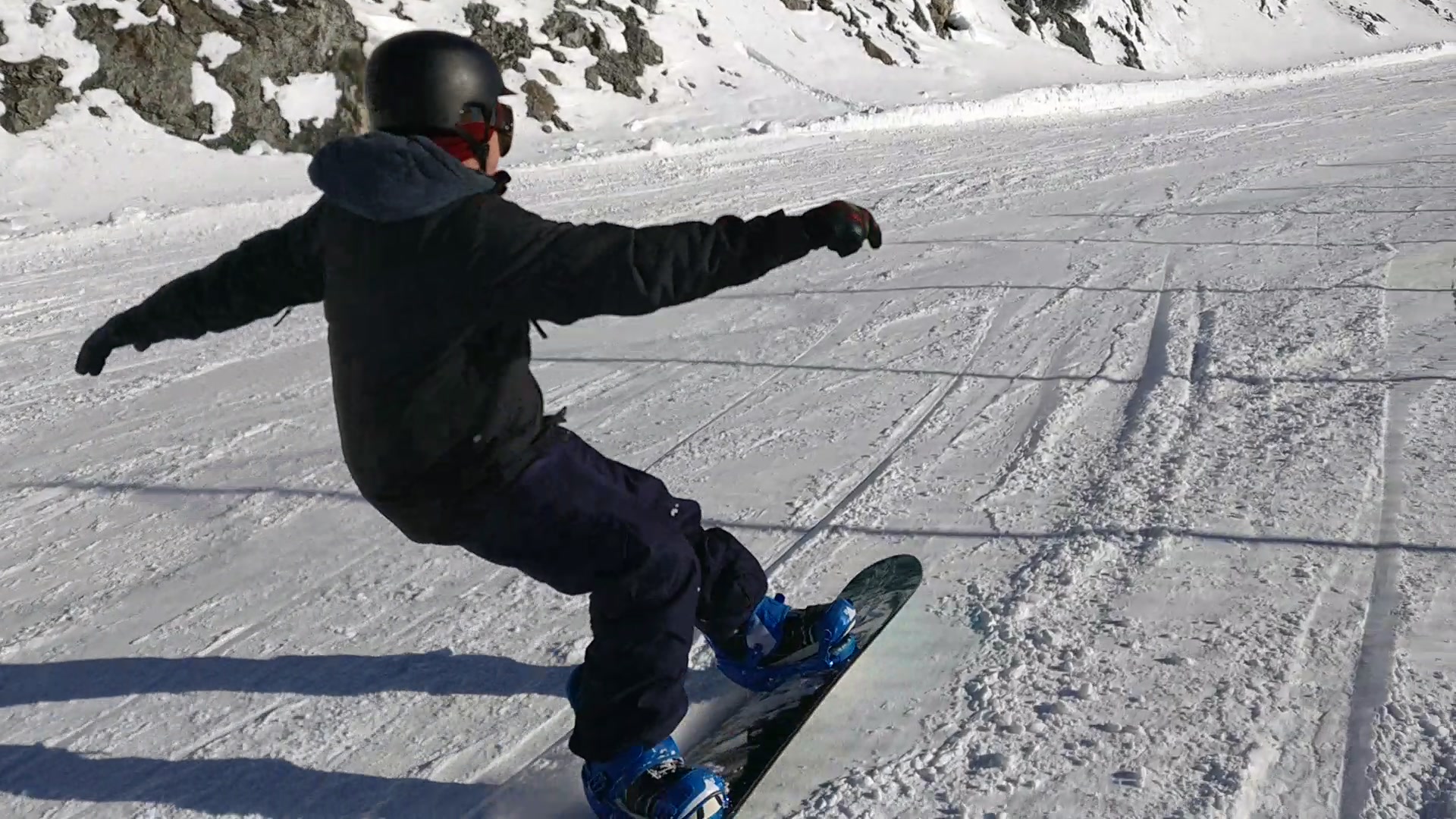}
        }
        {
            \includegraphics[width=\linewidth]{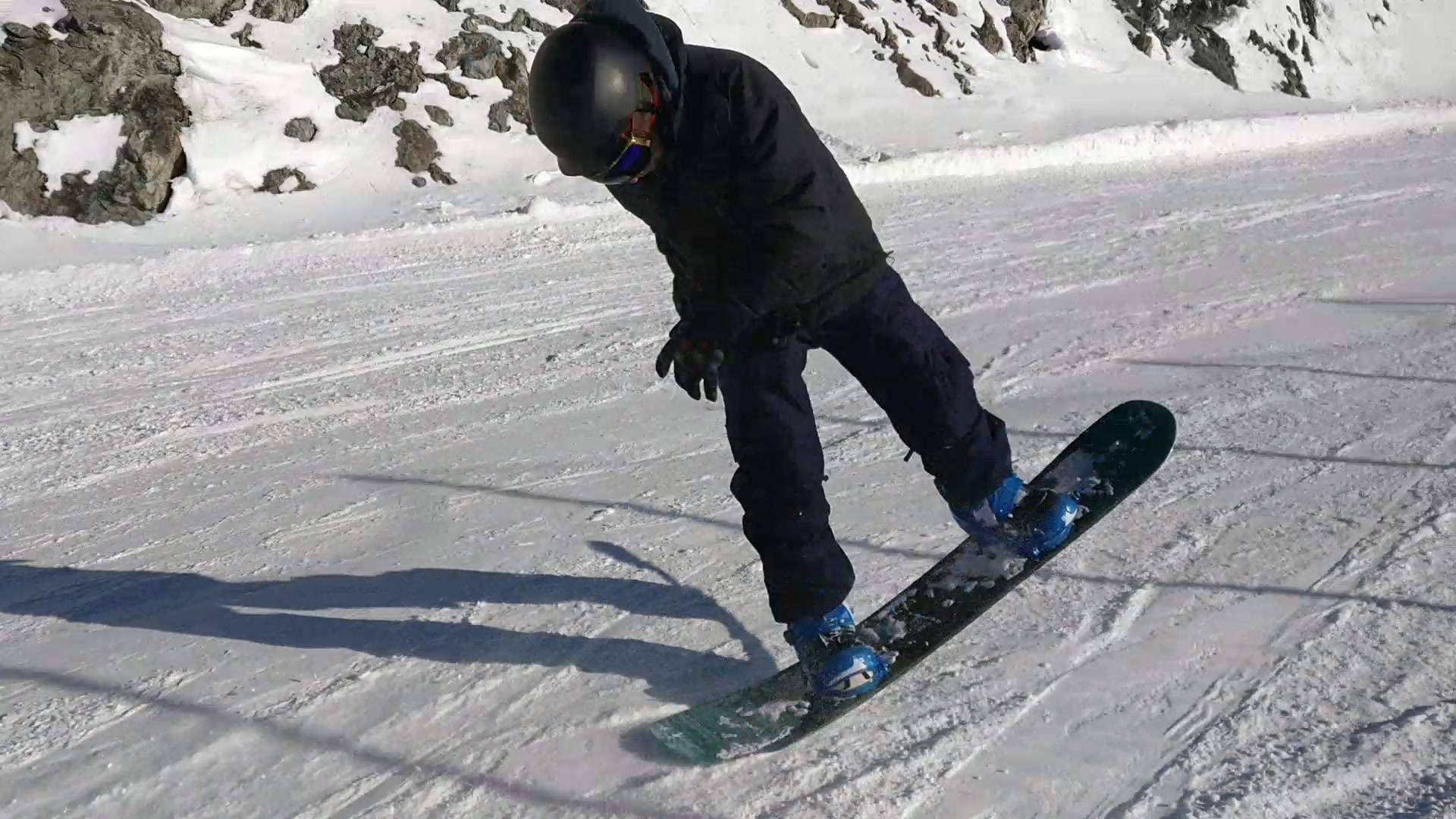}
        }
        {
            \includegraphics[width=\linewidth,trim=90mm 10mm 90mm 50mm,clip]{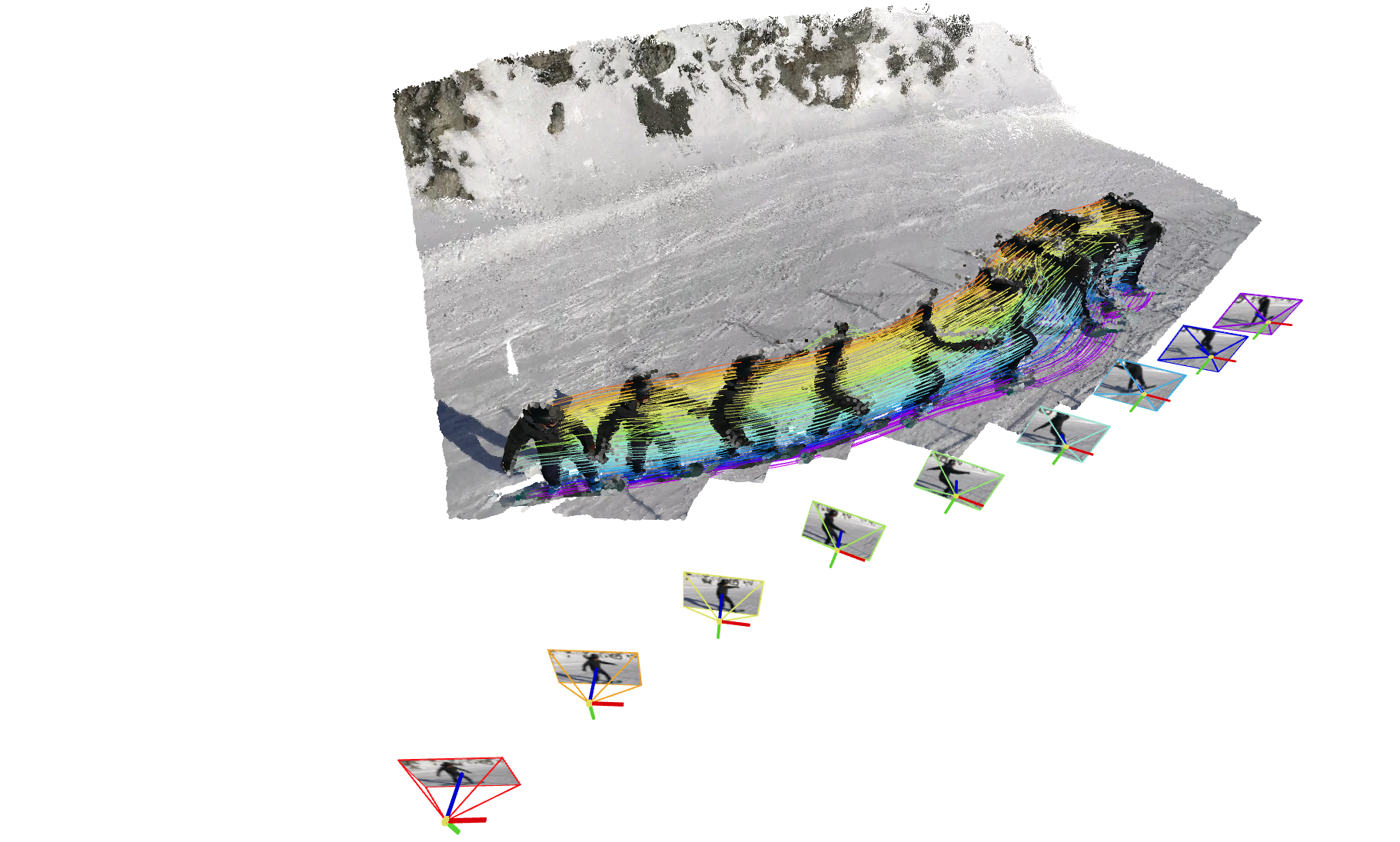}
        }

        \\

        \exampleblock
        {
            \includegraphics[width=\linewidth]{results/appendix_qualitative/input/kite-surf/00000.jpg}
        }
        {
            \includegraphics[width=\linewidth]{results/appendix_qualitative/input/kite-surf/00023.jpg}
        }
        {
            \includegraphics[width=\linewidth]{results/appendix_qualitative/input/kite-surf/00048.jpg}
        }
        {
            \includegraphics[width=\linewidth,trim=90mm 0mm 100mm 0mm,clip]{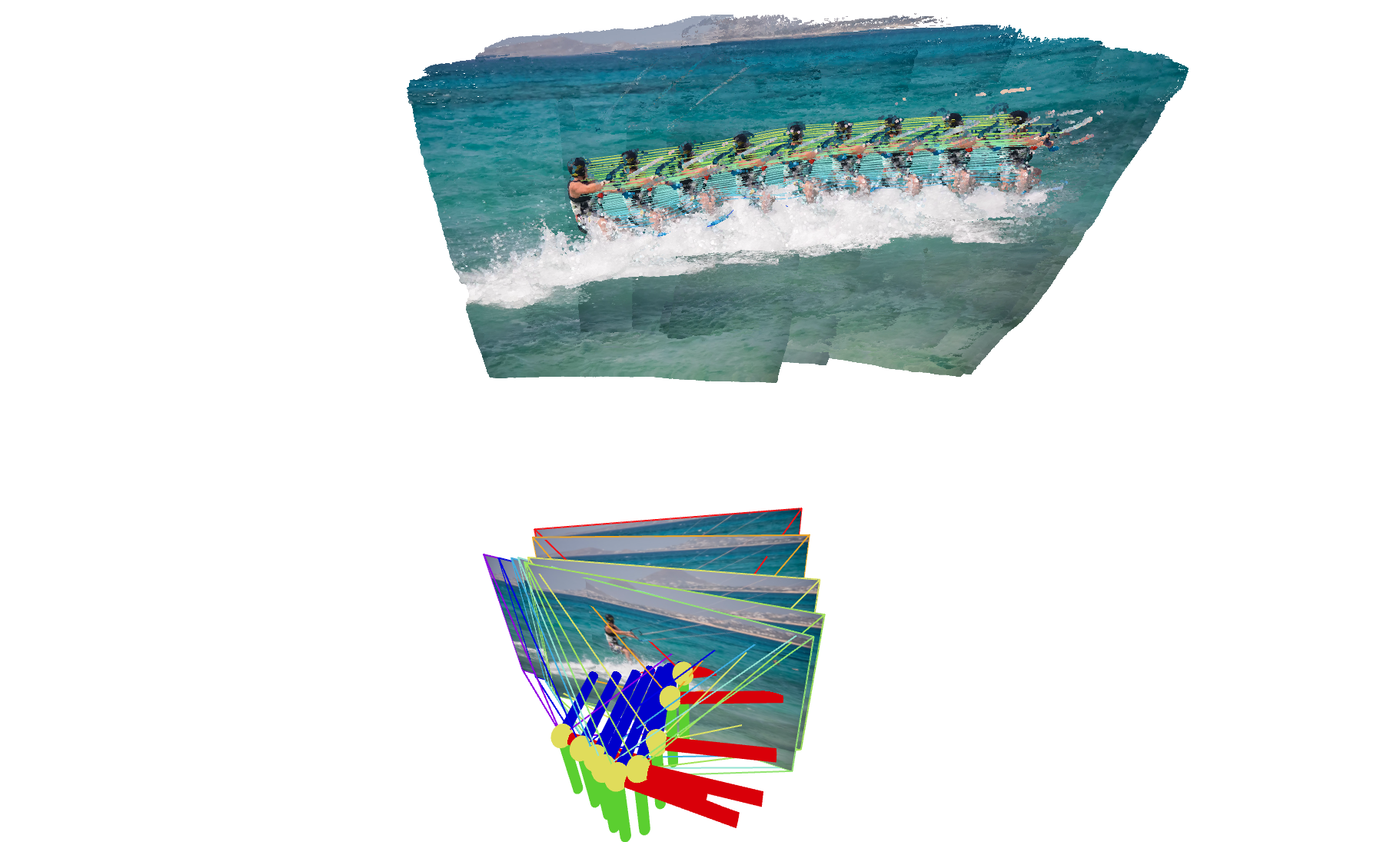}
        }

        &

        \exampleblock
        {
            \includegraphics[width=\linewidth]{results/appendix_qualitative/input/soapbox/00000.jpg}
        }
        {
            \includegraphics[width=\linewidth]{results/appendix_qualitative/input/soapbox/00023.jpg}
        }
        {
            \includegraphics[width=\linewidth]{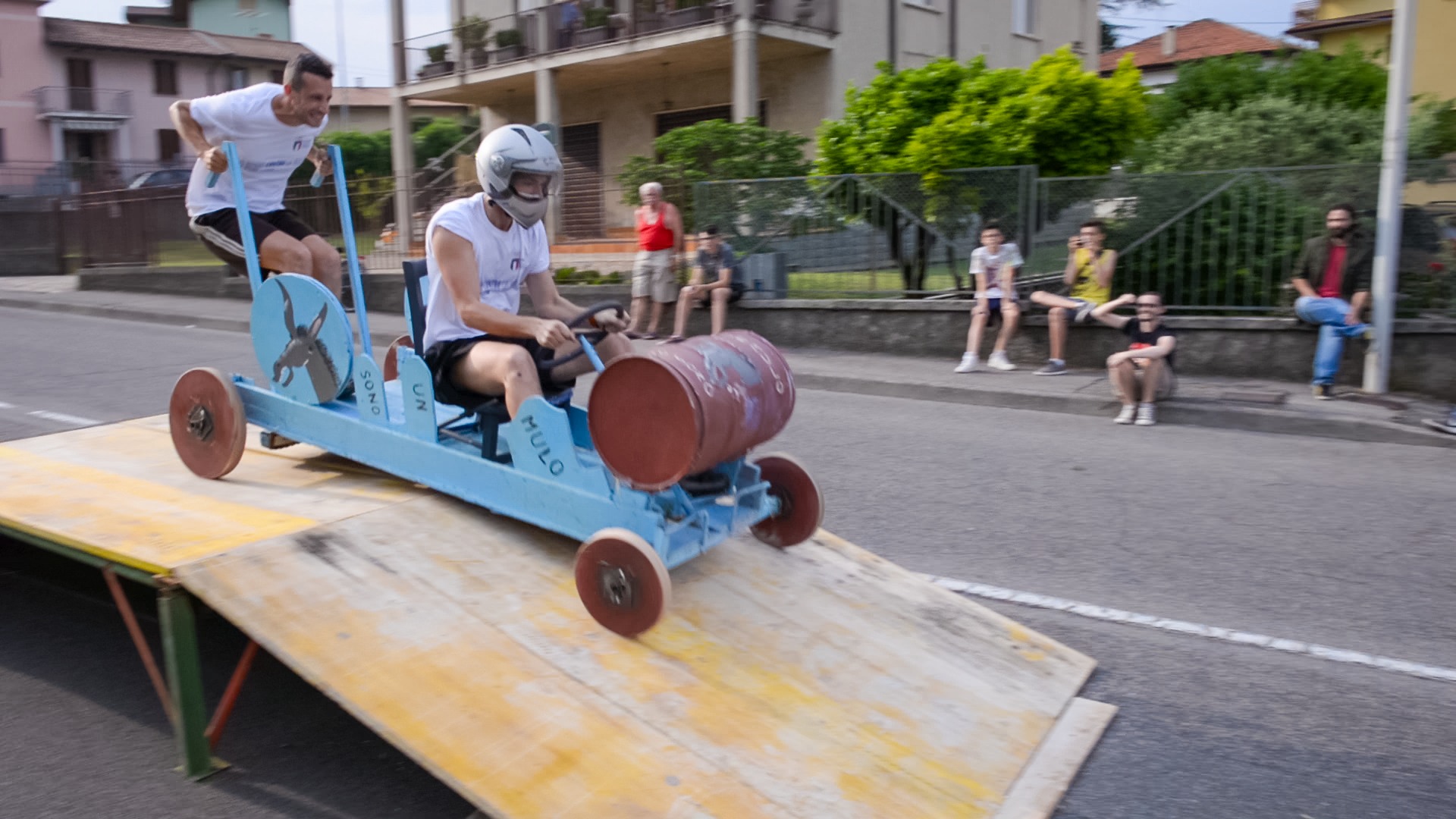}
        }
        {
            \includegraphics[width=\linewidth,trim=90mm 0mm 100mm 30mm,clip]{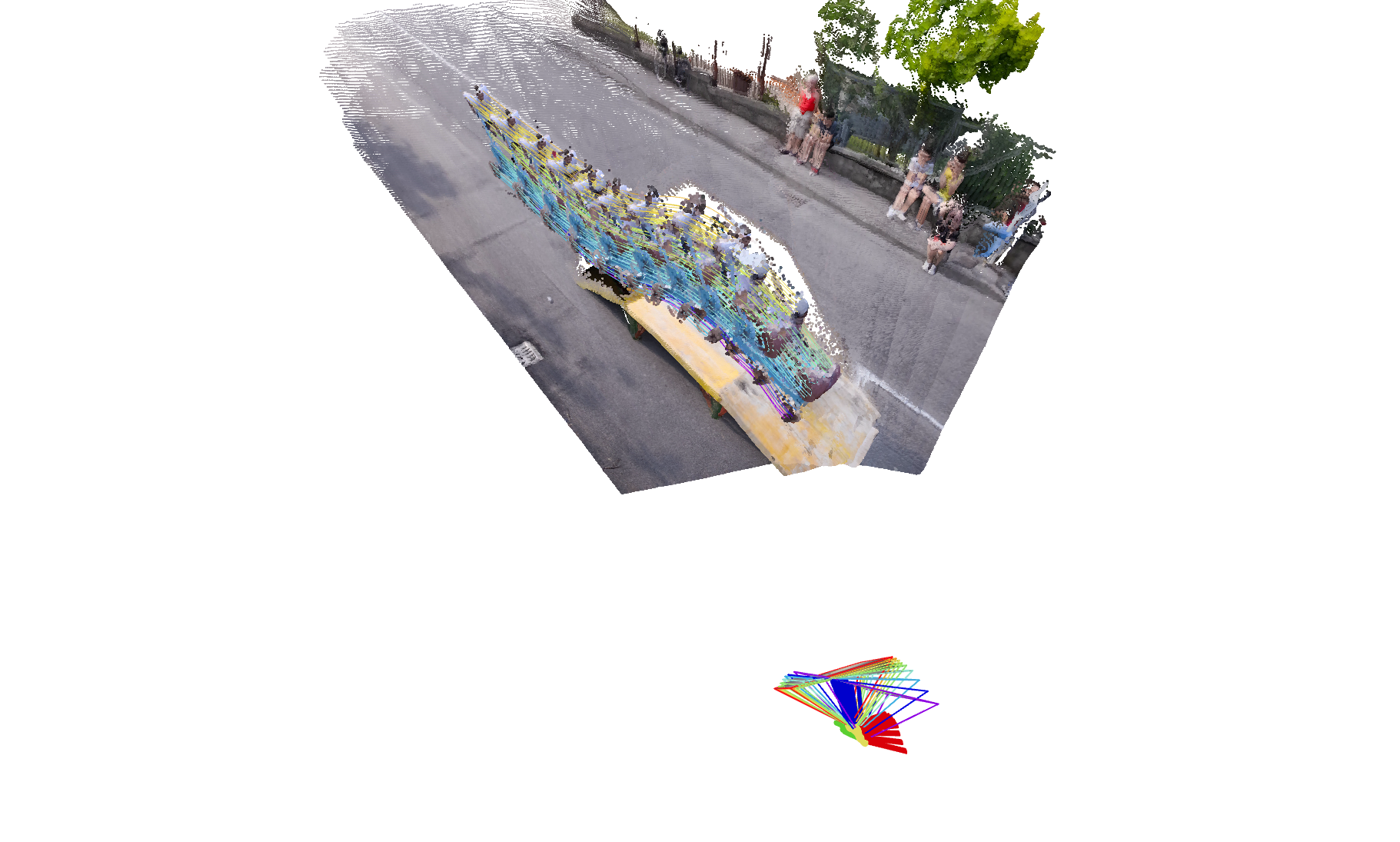}
        }

        \\

        \sideexampleblock
        {
            \includegraphics[width=\linewidth,height=\sideInputHeight]{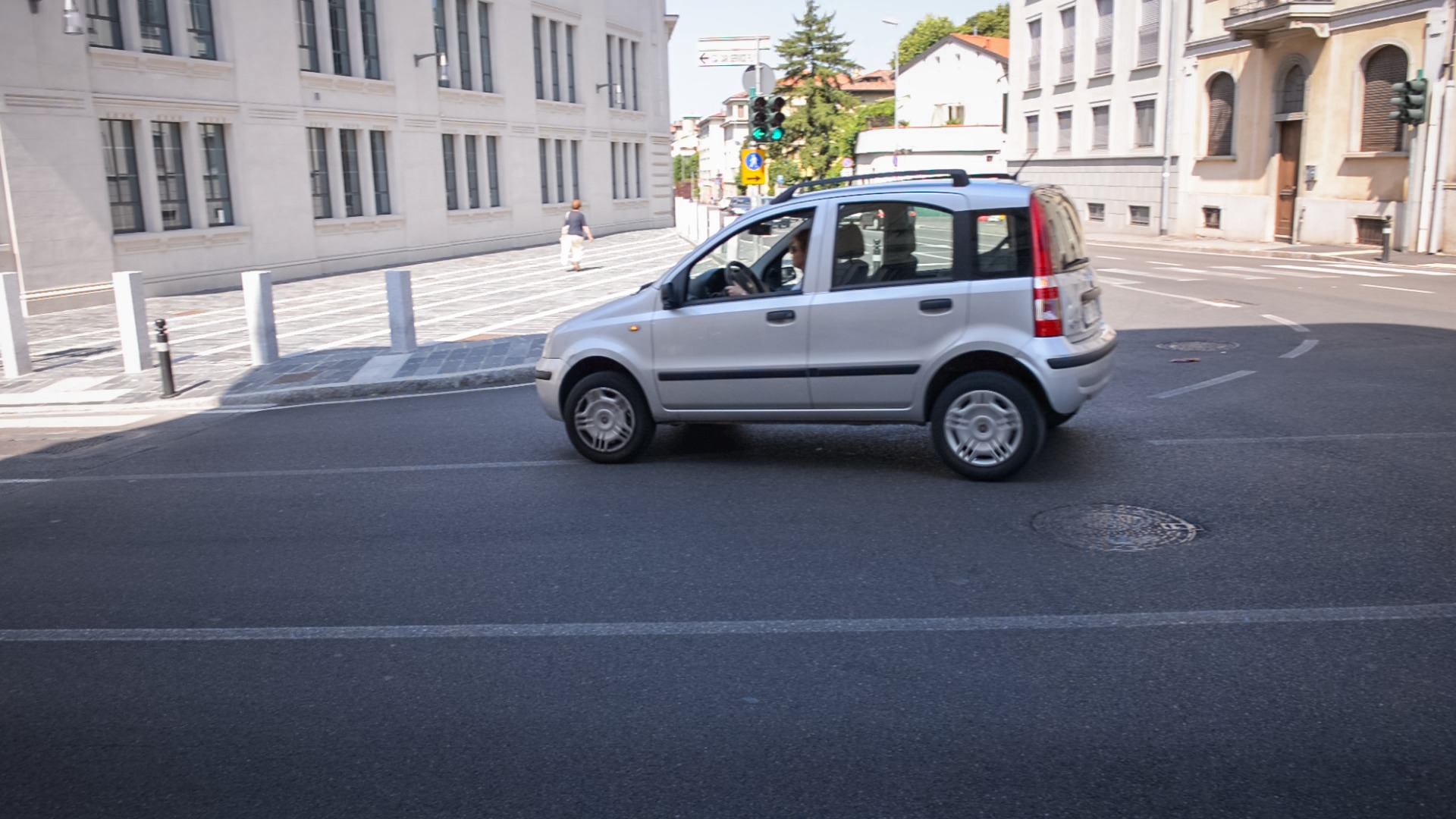}
        }
        {
            \includegraphics[width=\linewidth,height=\sideInputHeight]{results/appendix_qualitative/input/car-shadow/00012.jpg}
        }
        {
            \includegraphics[width=\linewidth,height=\sideInputHeight]{results/appendix_qualitative/input/car-shadow/00023.jpg}
        }
        {
            \includegraphics[width=\linewidth,trim=10mm 10mm 10mm 10mm,clip]{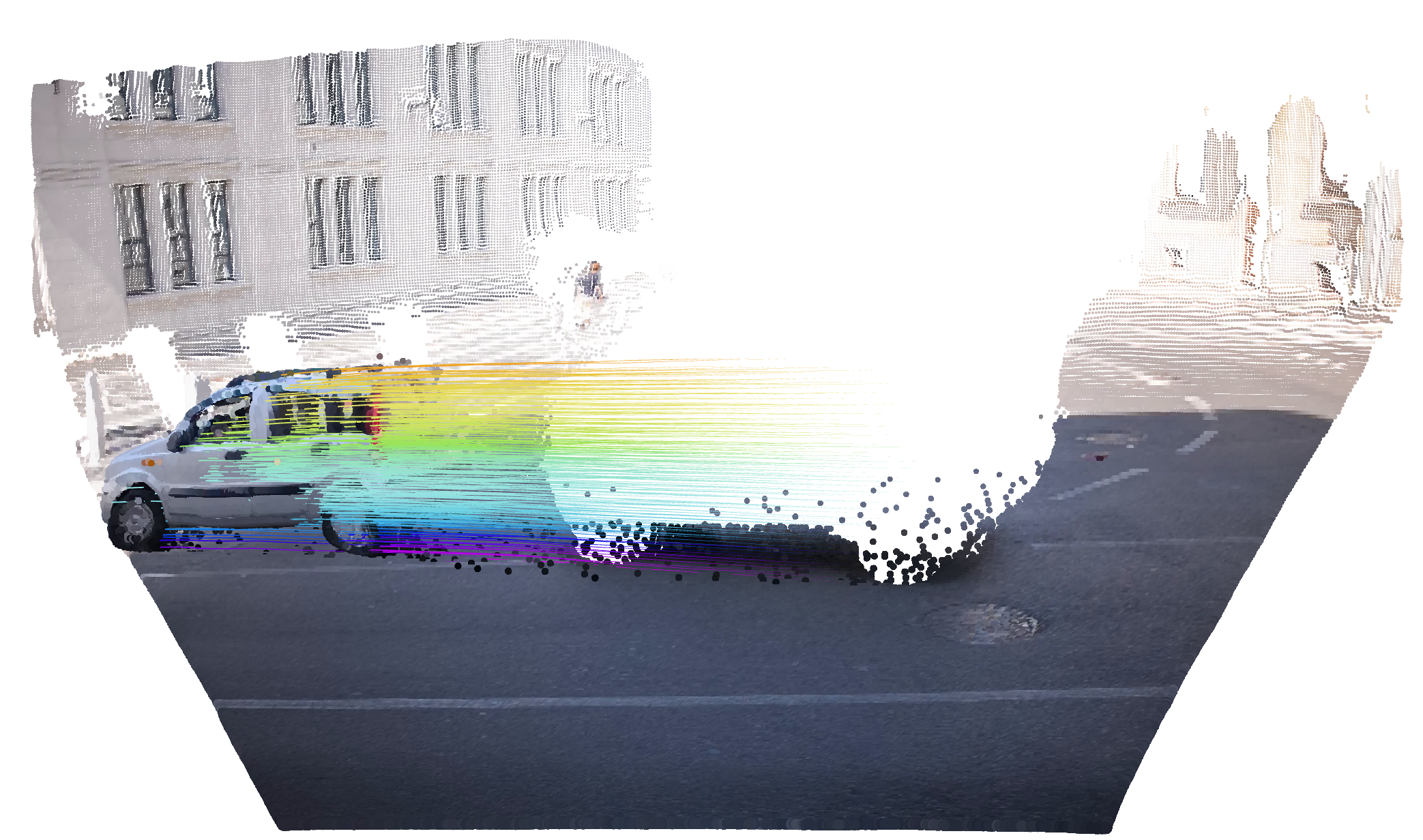}
        }

        &

        \sideexampleblock
        {
            \includegraphics[width=\linewidth,height=\sideInputHeight]{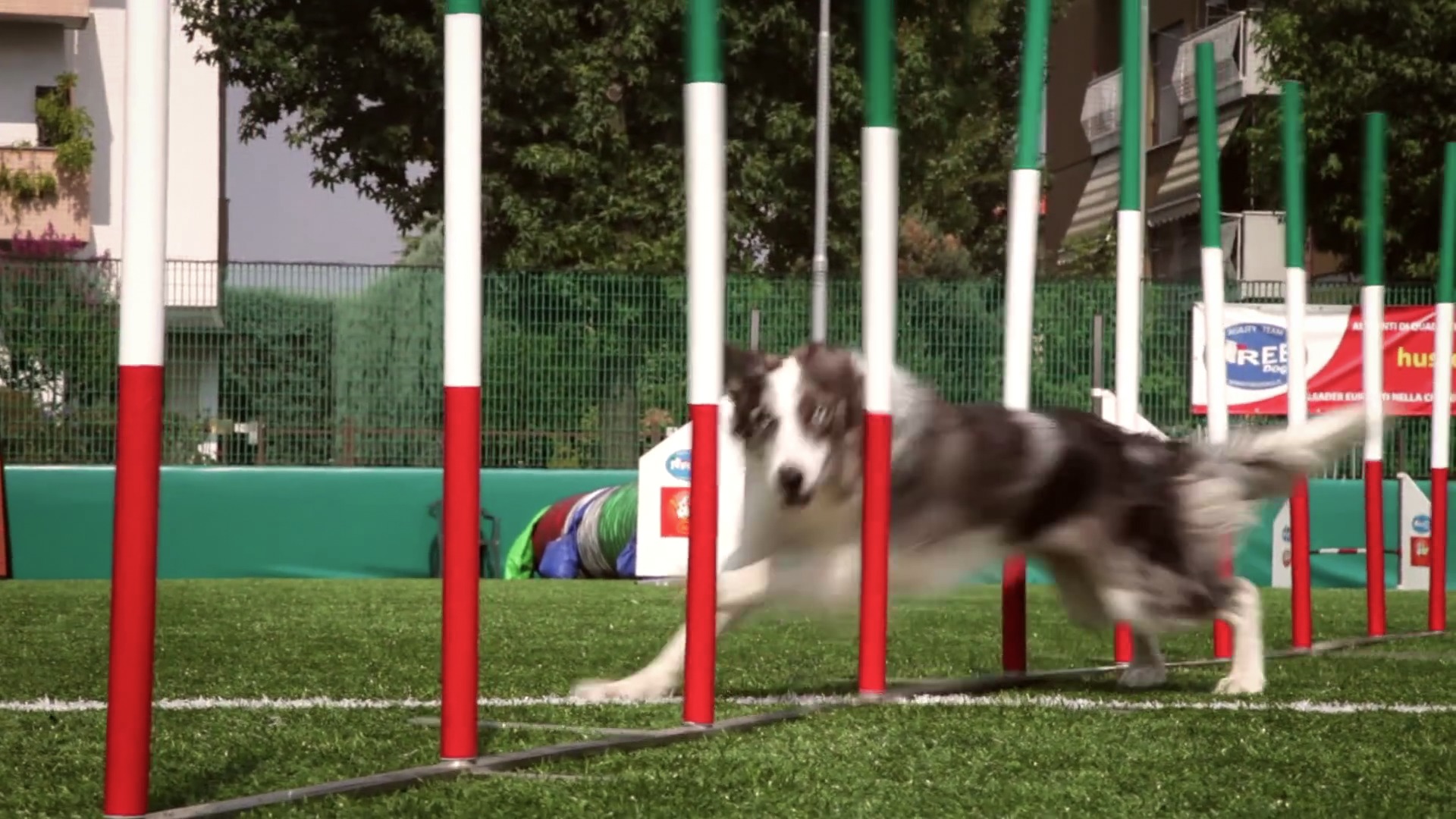}
        }
        {
            \includegraphics[width=\linewidth,height=\sideInputHeight]{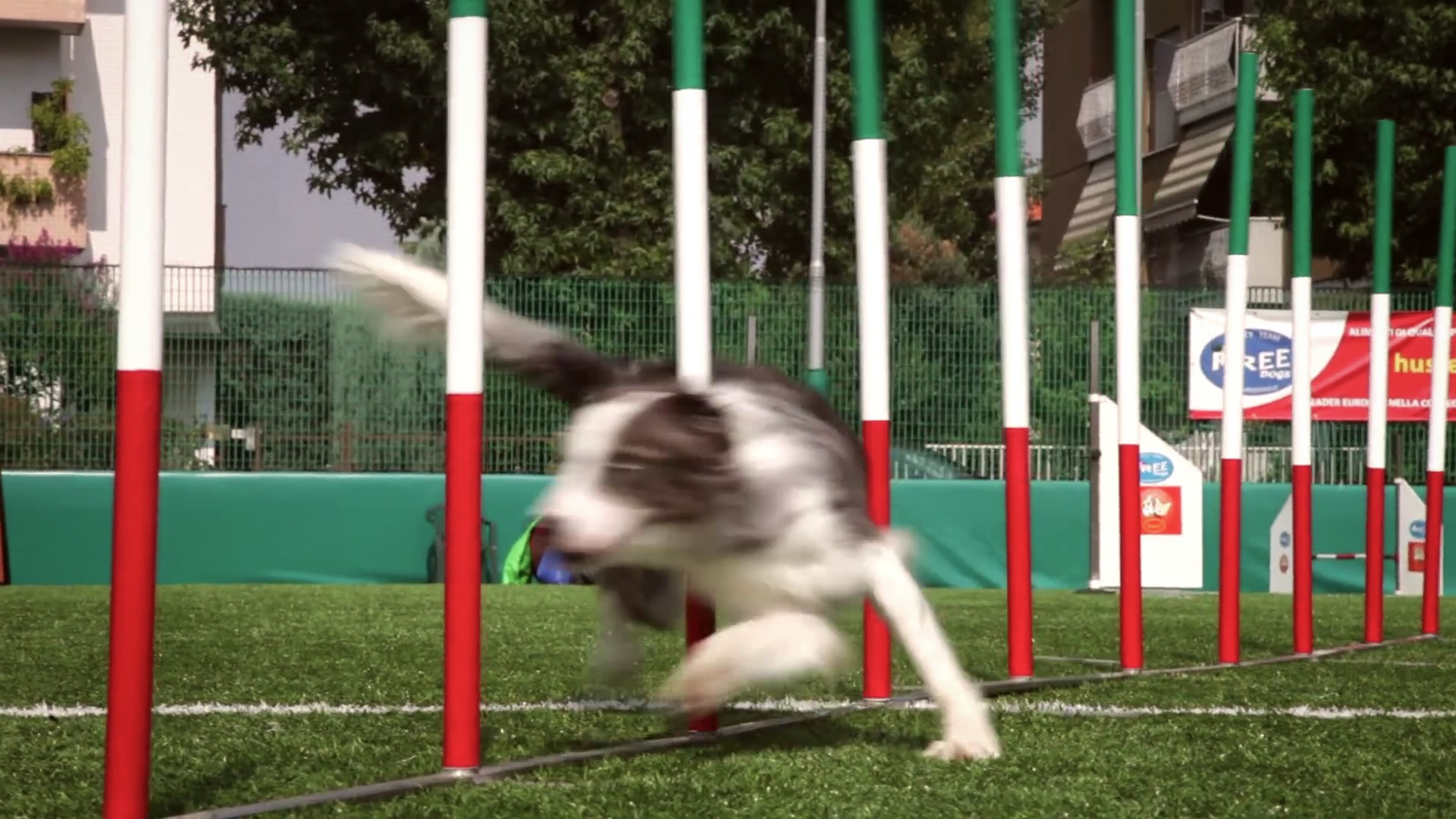}
        }
        {
            \includegraphics[width=\linewidth,height=\sideInputHeight]{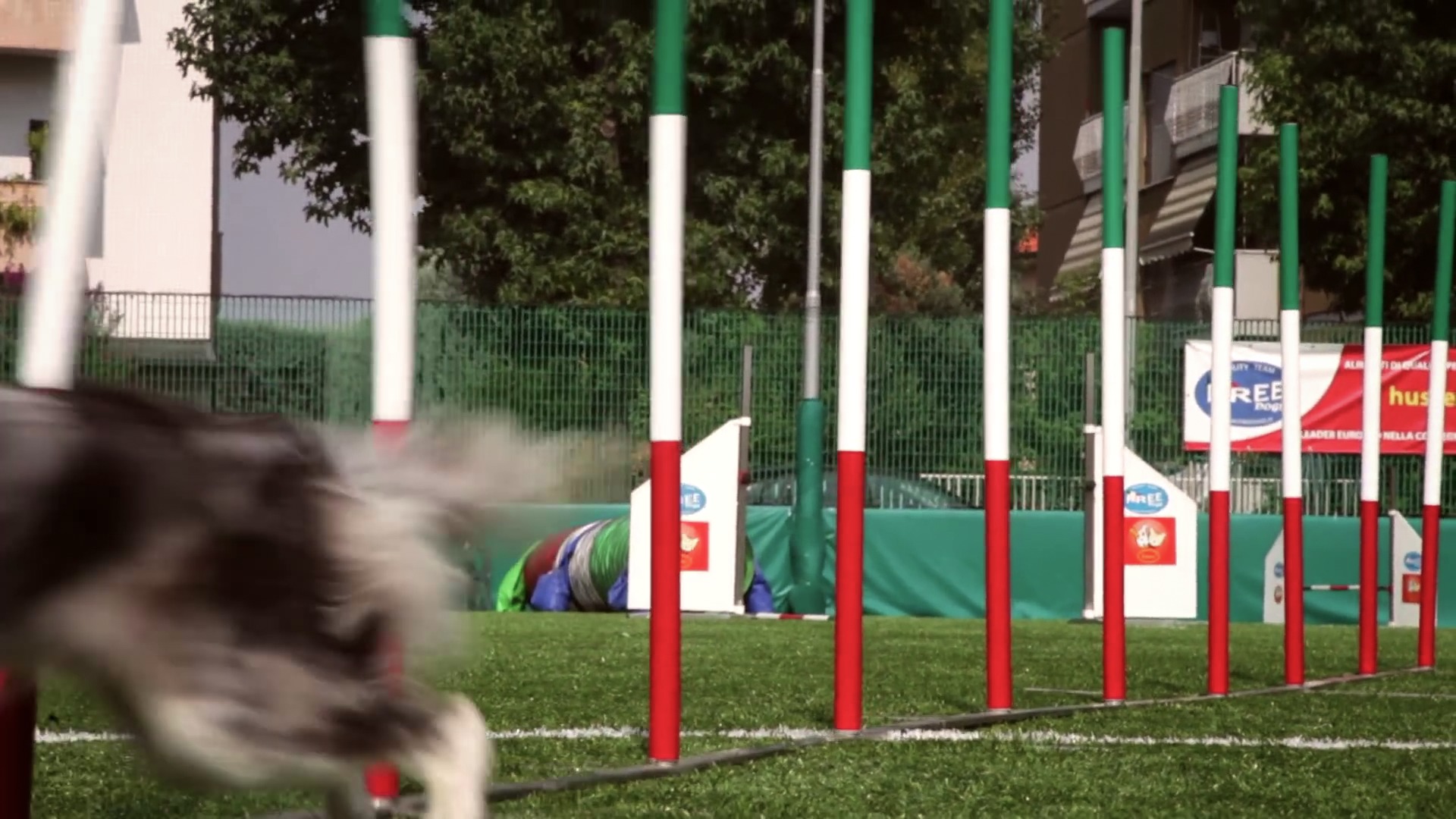}
        }
        {
            \includegraphics[width=\linewidth,trim=50mm 20mm 50mm 50mm,clip]{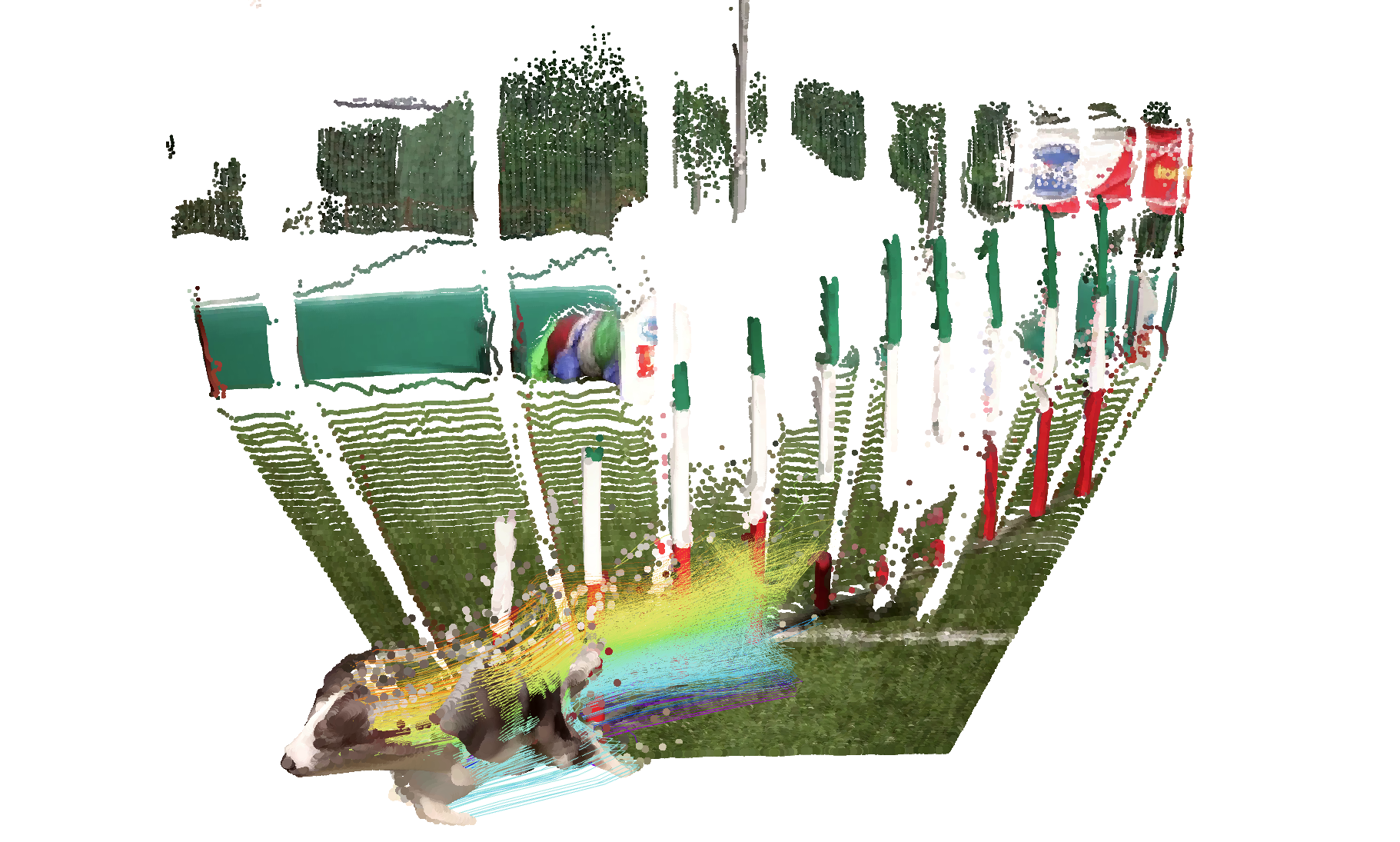}
        }

        \\[4pt]

        \sideexampleblock
        {
            \includegraphics[width=\linewidth,height=\sideInputHeight]{results/appendix_qualitative/input/goldfish/00000.jpg}
        }
        {
            \includegraphics[width=\linewidth,height=\sideInputHeight]{results/appendix_qualitative/input/goldfish/00012.jpg}
        }
        {
            \includegraphics[width=\linewidth,height=\sideInputHeight]{results/appendix_qualitative/input/goldfish/00023.jpg}
        }
        {
            \includegraphics[width=\linewidth,trim=10mm 10mm 10mm 10mm,clip]{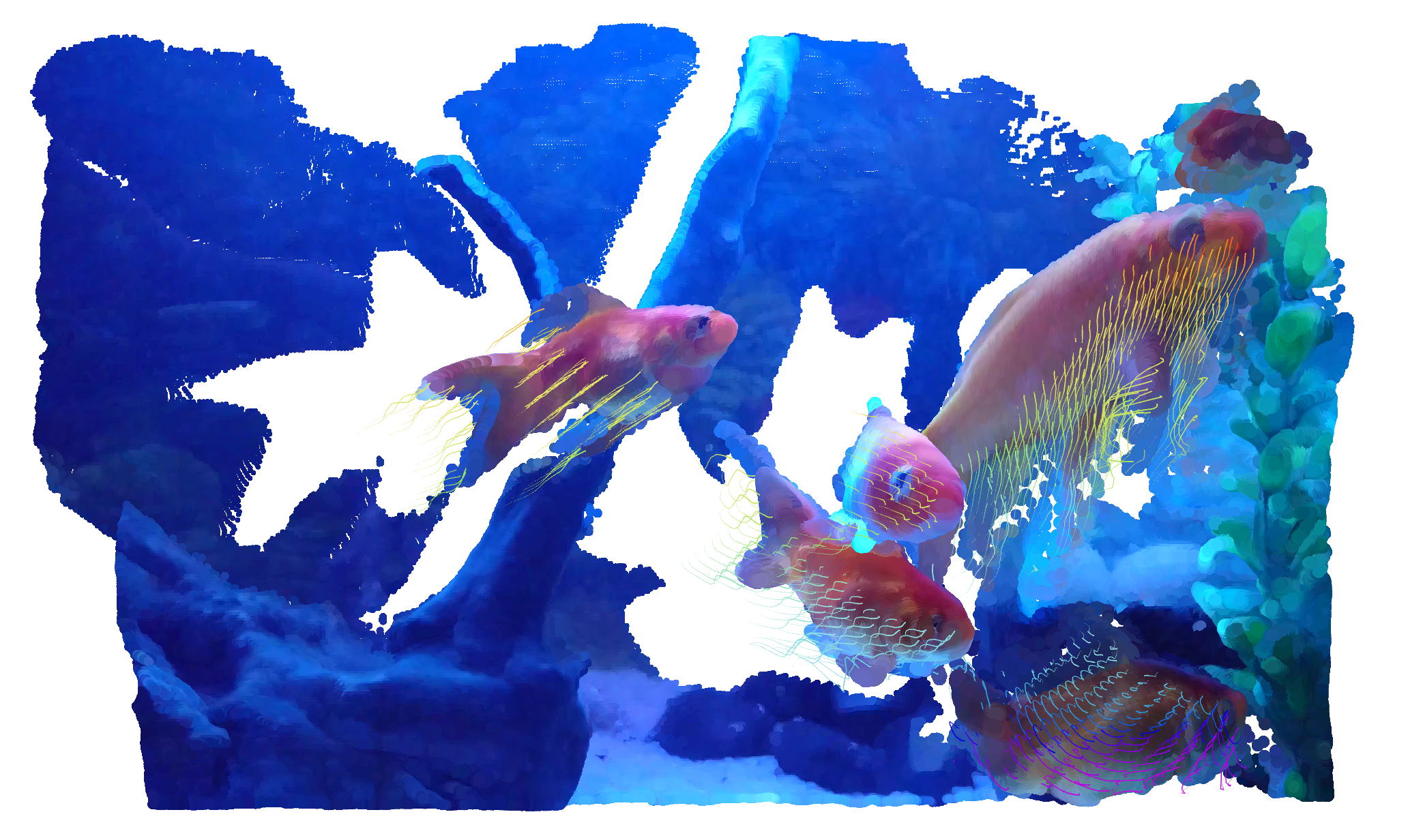}
        }

        &

        \sideexampleblock
        {
            \includegraphics[width=\linewidth,height=\sideInputHeight]{results/appendix_qualitative/input/camel/00000.jpg}
        }
        {
            \includegraphics[width=\linewidth,height=\sideInputHeight]{results/appendix_qualitative/input/camel/00012.jpg}
        }
        {
            \includegraphics[width=\linewidth,height=\sideInputHeight]{results/appendix_qualitative/input/camel/00019.jpg}
        }
        {
            \includegraphics[width=\linewidth]{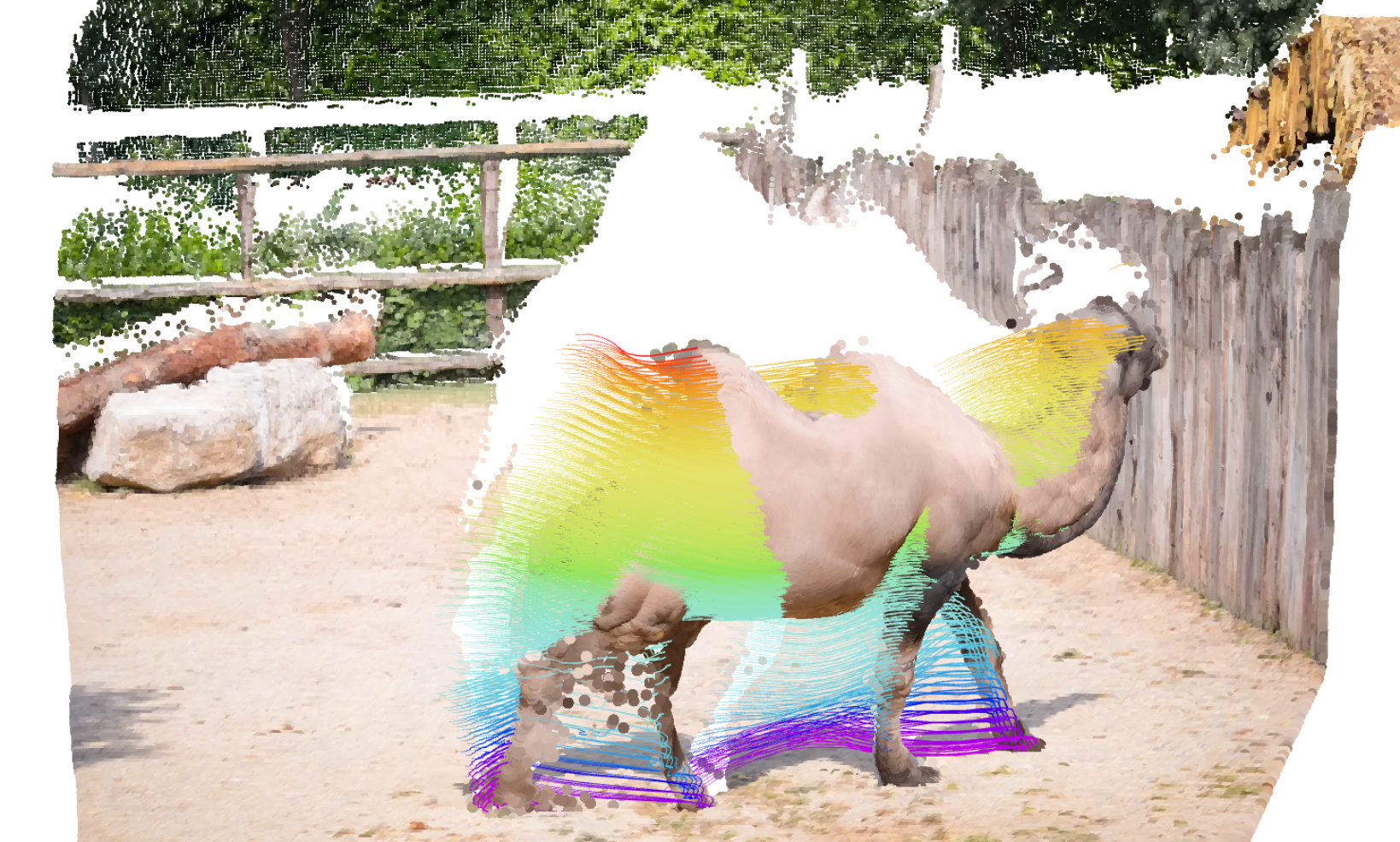}
        }

        \\

    \end{tabular}

    \caption{
        \textbf{Additional Qualitative Results} of motion tracking, point cloud reconstruction and camera pose estimation.
    }
    \label{fig:appendix_additional_vis}
\end{figure*}

In Fig.~\ref{fig:appendix_additional_vis}, we show additional qualitative results, including merged point clouds, tracking and camera pose estimation, and additional 3D tracking showcases.

\subsection{Confidence Filtering}
We use the $\mathrm{expp1}$ confidence activation, i.e., $c = 1+\exp(\hat{c})$ with $\hat{c}$ denoting the raw confidence output, following VGGT~\citep{wang2025vggt}. This allows us to apply percentile-based confidence filtering to remove points that the model predicts with low confidence, typically distributed at the far-away background or foreground boundaries.

\begin{figure*}[htbp]
    \centering
    \setlength{\tabcolsep}{2pt}
    \renewcommand{\arraystretch}{0}

    \newcommand{\inputpair}[2]{
        \begin{minipage}[t]{\linewidth}
            \vspace{0pt}
            \centering
            #1\\[0pt]
            #2
        \end{minipage}
    }

    \newcommand{\resultcell}[1]{
        \begin{minipage}[t]{\linewidth}
            \vspace{0pt}
            \centering
            \makebox[\linewidth][c]{#1}
        \end{minipage}
    }

    \begin{tabular}{
        @{}
        p{0.19\textwidth}
        p{0.38\textwidth}
        p{0.38\textwidth}
        @{}
    }

    &
    \centering\bfseries Raw
    &
    \centering\arraybackslash\bfseries Confidence Filtered
    \\[4pt]

    \inputpair
    {
        \includegraphics[width=\linewidth]{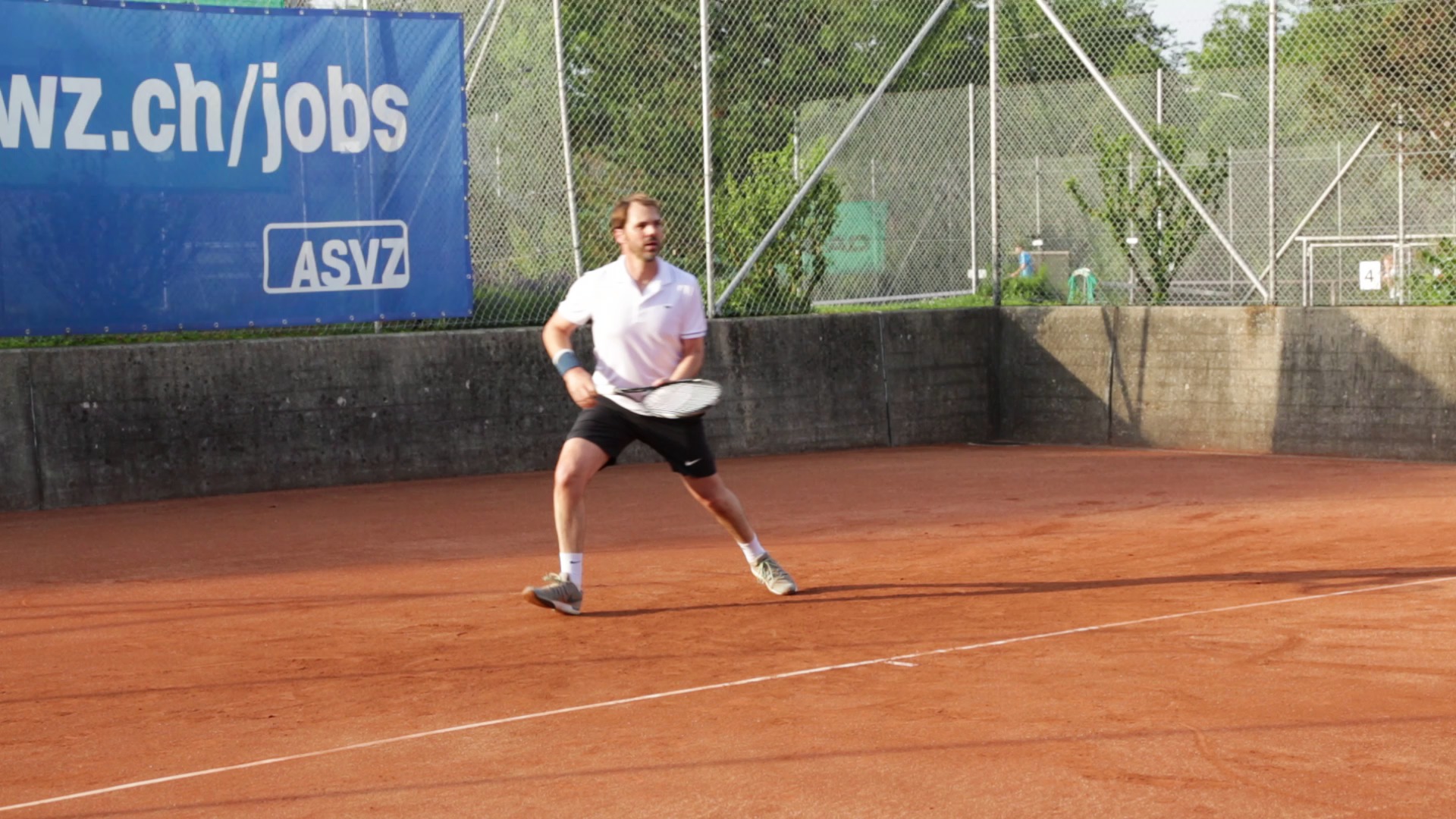}
    }
    {
        \includegraphics[width=\linewidth]{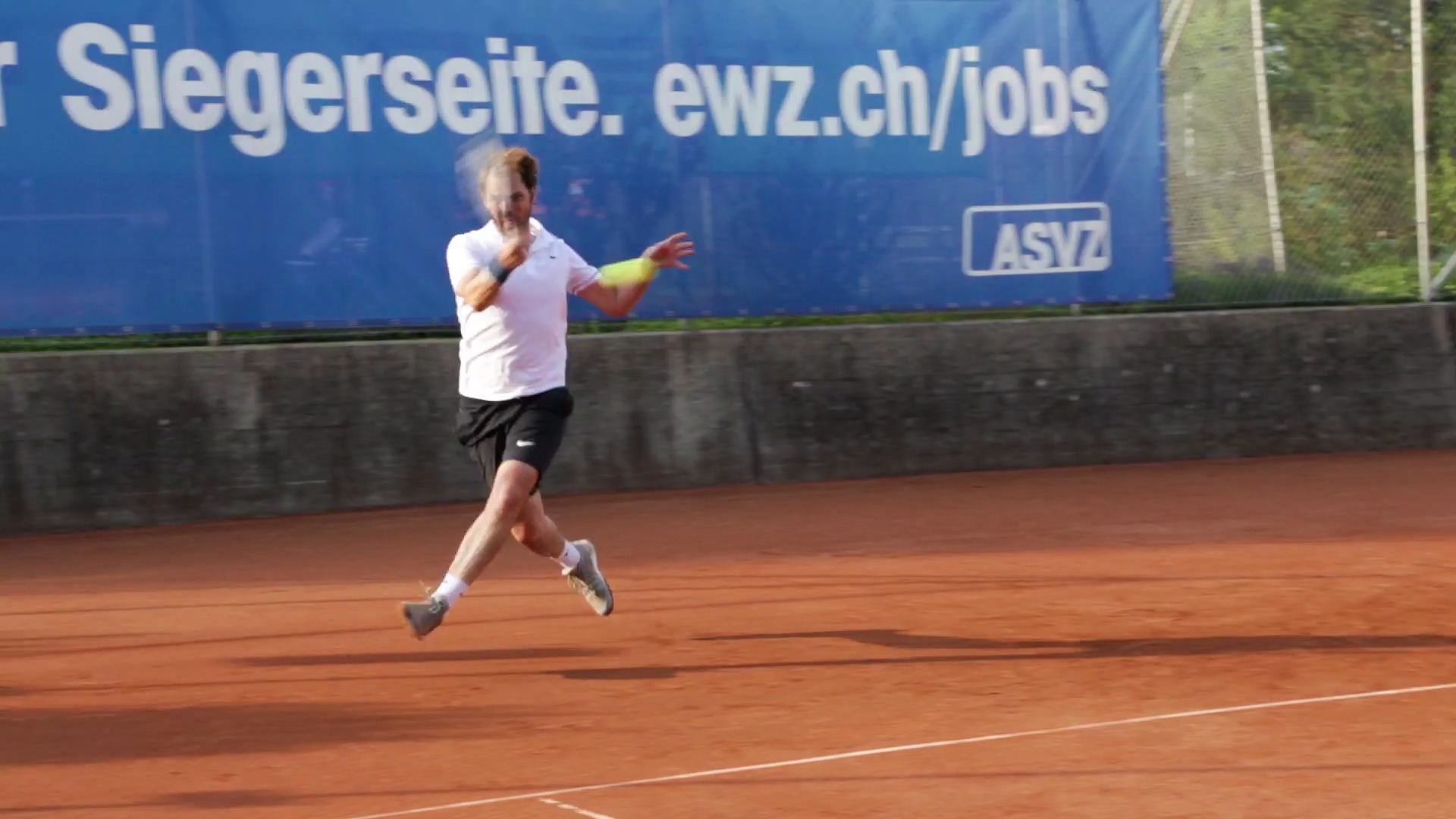}
    }
    &
    \resultcell{
        \includegraphics[width=\linewidth,trim=50mm 20mm 100mm 50mm,clip]{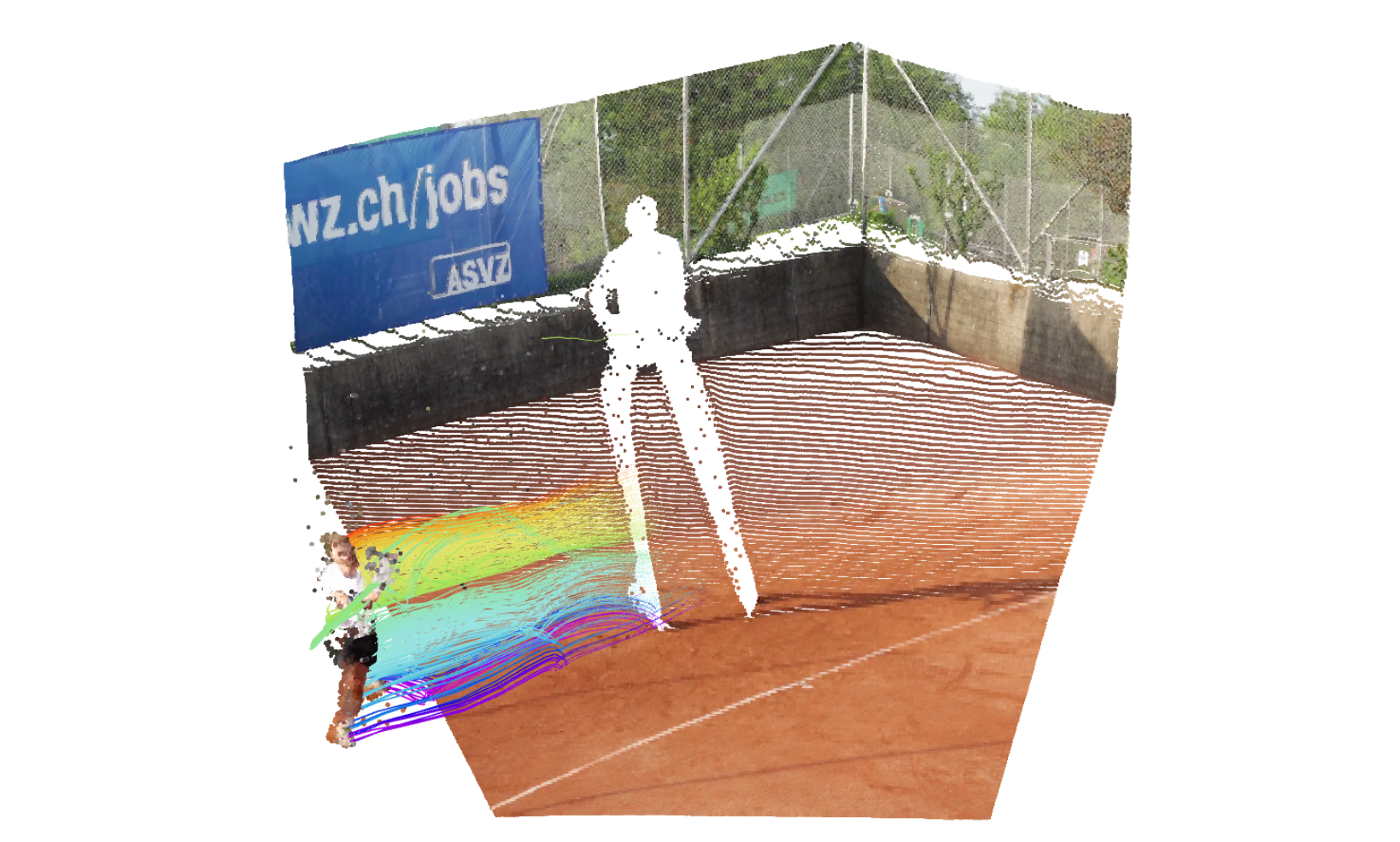}
    }
    &
    \resultcell{
        \includegraphics[width=\linewidth,trim=50mm 20mm 100mm 50mm,clip]{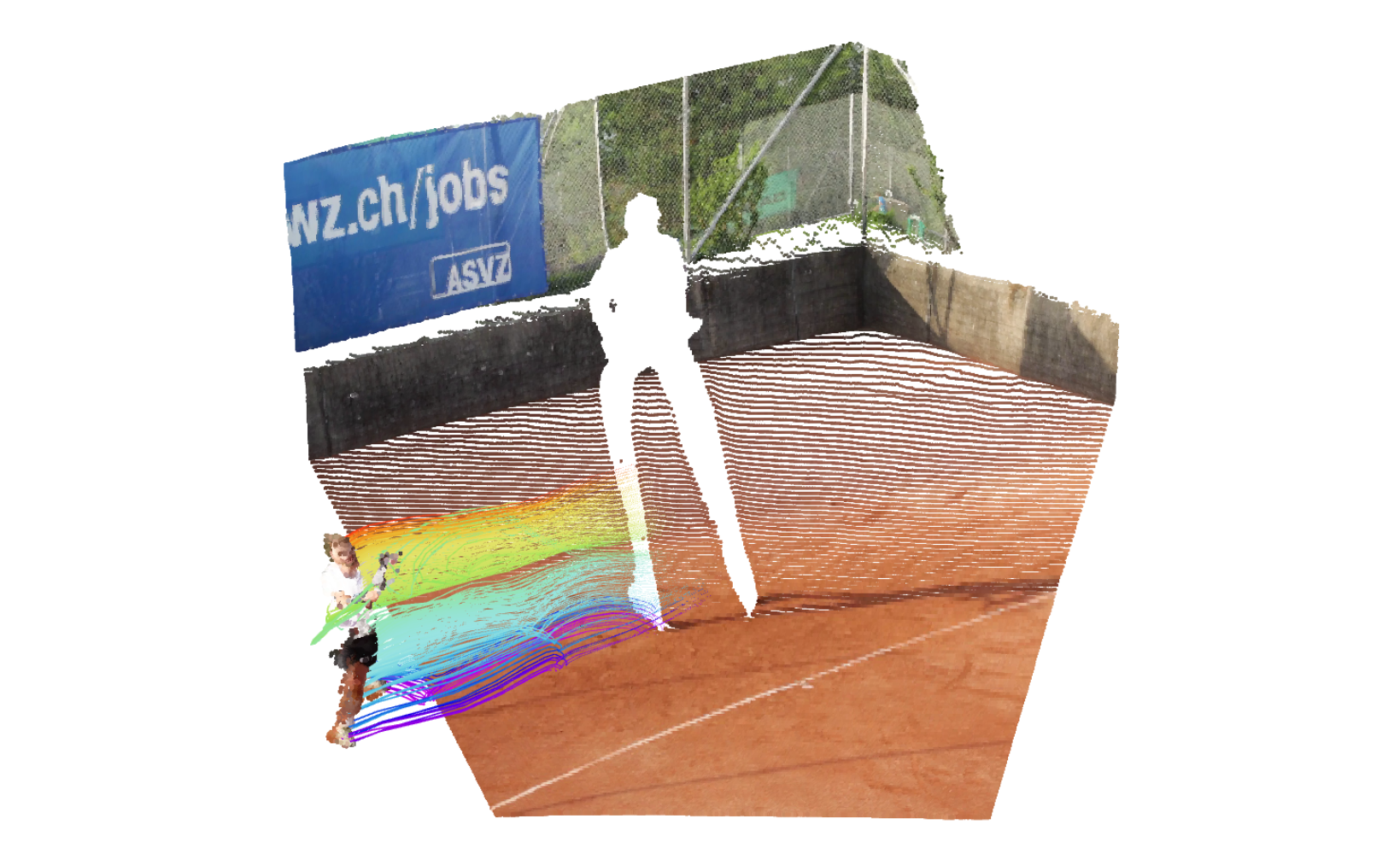}
    }
    \\[-5pt]

    \inputpair
    {
        \includegraphics[width=\linewidth]{results/appendix_qualitative/input/judo/00000.jpg}
    }
    {
        \includegraphics[width=\linewidth]{results/appendix_qualitative/input/judo/00023.jpg}
    }
    &
    \resultcell{
        \includegraphics[width=\linewidth]{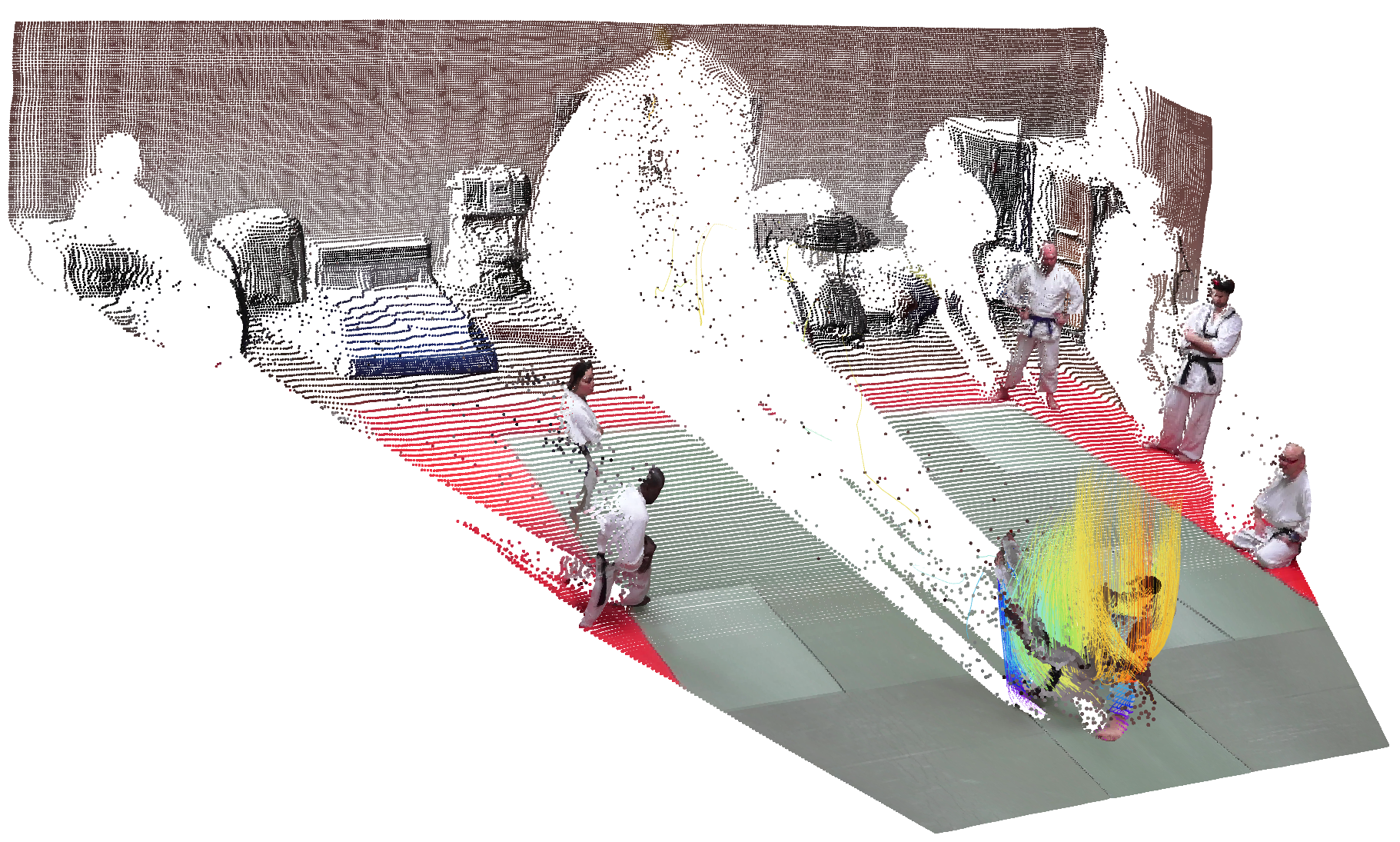}
    }
    &
    \resultcell{
        \includegraphics[width=\linewidth]{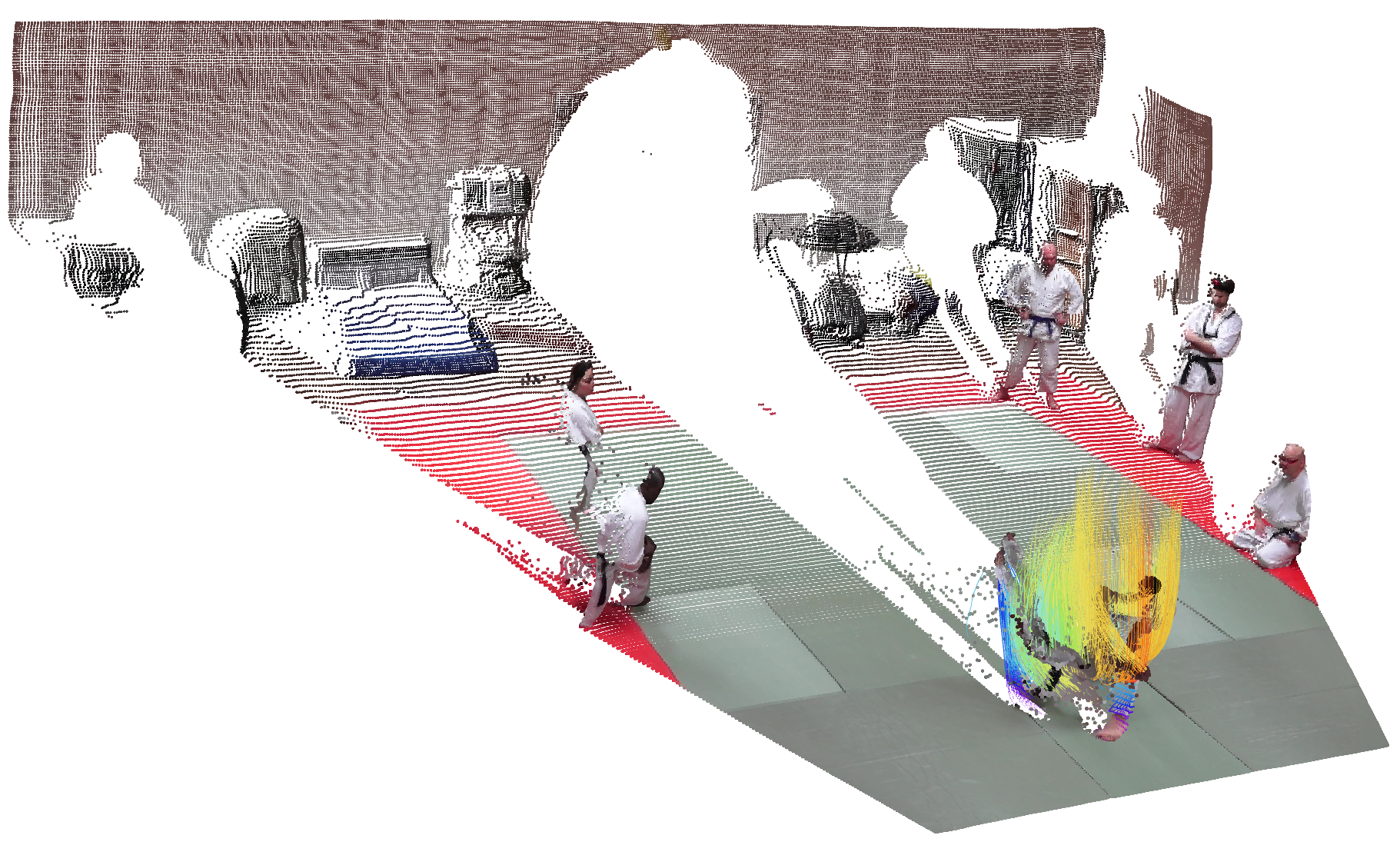}
    }
    \\

    \end{tabular}
    
    \caption{
        \textbf{Confidence Filtering} for flying point suppression.
    }
    \label{fig:appendix_conf_filter}
\end{figure*}

In Fig.~\ref{fig:appendix_conf_filter}, we demonstrate an application of confidence filtering. As shown in the Raw column, the direct tracking outputs are already visually clean, but may contain a small number of flying points. By applying confidence filtering, with a percentile as low as $2\%$, these flying points can be effectively suppressed, resulting in cleaner foreground and background content.

We hereby clarify that for fair comparisons, the visualization results for any method in Fig.~\ref{fig:track_visualization} and Fig.~\ref{fig:dispmap_visualization} are \textbf{NOT} applied with confidence-based suppression.

\section{Training Preparation}
\subsection{Datasets}
\label{sec:appendix_dataset}
We distinguish two types of input frames: \emph{encoder frames}, which are fed to the encoder to extract scene tokens, and \emph{query frames}, from which local RGB patches are extracted to construct decoder queries. The two share the same aspect ratio but may differ in spatial resolution.

For encoder frames, each raw video is maximally center-cropped to one of five predefined aspect ratios. The cropped frames are then uniformly resized to the largest resolution within a $480\times832$ bound, resulting in five encoder frame sizes: $480\times480$, $480\times640$, $480\times832$, $352\times832$, and $256\times832$. The corresponding query frames use the same cropped content before resizing, and their resolutions therefore depend on the native resolution of each dataset. 

The resulting encoder frame and query frame resolutions are reported in Tab.~\ref{tab:dataset_preprocessing}. 
We preprocess each dataset under these spatial configurations and report the approximate number of resulting subclips for each configuration in the ``Est. Subclip Count Per Size'' column. During training, datasets are sampled according to the probabilities reported in the final column.  In practice, we tolerate a slight distortion between them.

To vary the effective field of view of query patches, we apply downsizing-only augmentation to the query frames. For each training clip, we first resize them within a $960\times 1664$ bound, then sample a side length scale factor from $[0.5,1.0]$, and resize all query frames by the same factor. This augmentation is applied only to the query frames for the decoder, while the pre-encoded latents remain unchanged.

We don't apply random content cropping, color jittering or Gaussian blurring.

\begin{table*}[t]
\centering
\captionsetup{skip=4pt}
\caption{
\textbf{Dataset statistics. }
The middle five columns are indexed by `Encoder Frame Size', with each entry recording the corresponding query frame size with the same aspect ratio.}
\label{tab:dataset_preprocessing}
\setlength{\tabcolsep}{4pt}
\renewcommand{\arraystretch}{1.1}
\resizebox{\textwidth}{!}{%
\begin{tabular}{@{}lccccccccc@{}}
\toprule
\multirow{2}{*}{\textbf{Dataset}}
& \multirow{2}{*}{\textbf{Native Res.}}
& \multirow{2}{*}{\textbf{Native AR.}}
& \multicolumn{5}{c}{\textbf{Encoder Frame Size}}
& \multirow{2}{*}{\shortstack{\textbf{Est. Subclip Count}\\\textbf{Per Size}}}
& \multirow{2}{*}{\shortstack{\textbf{Sampling}\\\textbf{Prob.}}} \\
\cmidrule(lr){4-8}
&
&
&
$\mathbf{480\times480}$
& $\mathbf{480\times640}$
& $\mathbf{480\times832}$
& $\mathbf{352\times832}$
& $\mathbf{256\times832}$
&
& \\
\midrule

Syn4D
& $720\times1280$
& $9{:}16$
& $720\times720$
& $720\times960$
& $720\times1248$
& $528\times1248$
& $384\times1248$
& 25k
& 16.064\% \\

DynamicReplica
& $720\times1280$
& $9{:}16$
& $720\times720$
& $720\times960$
& $720\times1248$
& $528\times1248$
& $384\times1248$
& 13k
& 16.064\% \\

PointOdyssey
& $540\times960$
& $9{:}16$
& $540\times540$
& $540\times720$
& $540\times936$
& $396\times936$
& $288\times936$
& 20k
& 16.064\% \\

Kubric Cotracker3
& $512\times512$
& $1{:}1$
& $512\times512$
& -
& $285\times494$
& -
& -
& 45k
& 12.851\% \\

DL3DV
& $528\text{--}539\times927\text{--}972$
& $\sim4{:}7$
& $525\times525$
& $525\times700$
& $525\times910$
& $385\times910$
& $280\times910$
& 18K
& 3.213\% \\

ScanNet++
& $1168\times1752$
& $2{:}3$
& $1005\times1005$
& $1005\times1340$
& $1005\times1742$
& $737\times1742$
& $536\times1742$
& 28k
& 8.032\% \\

VKitti2
& $375\times1242$
& 1:3.3
& $368\times368$
& $368\times491$
& $368\times638$
& $368\times870$
& $368\times1200$
& 4k
& 8.032\% \\

MVS-Synth
& $1080\times1920$
& $9{:}16$
& $1080\times1080$
& $1080\times1440$
& $1080\times1872$
& $792\times1872$
& $576\times1872$
& 0.3k
& 1.606\% \\

Spring
& $1080\times1920$
& $9{:}16$
& $1080\times1080$
& $1080\times1440$
& $1080\times1872$
& $792\times1872$
& $576\times1872$
& 0.1k
& 0.402\% \\

Waymo
& $1280\times1920$
& $2{:}3$
& $1095\times1095$
& $1095\times1460$
& $1095\times1898$
& $803\times1898$
& $584\times1898$
& 12k
& 8.032\% \\

Aria Digital Twin
& $1480\times1480$
& $1{:}1$
& $960\times960$
& -
& $554\times960$
& -
& -
& 35k
& 9.639\% \\

\bottomrule
\end{tabular}%
}
\end{table*}

\subsection{Preprocessing and Sampling}
\label{sec:appendix_preprocess_and_sampling}
\textbf{Preprocessing. }
Each raw video is subsampled into 49-frame clips with frame intervals ranging from 1 to 4. We then pre-encode each clip with the Wan-VAE~\citep{wan2025} and store the normalized latents.

For each clip, camera poses are defined in the first-frame relative world coordinate. The 3D position GT is expressed in the $t_{\mathrm{cam}}$ camera coordinate. We adopt mean normalization as in MotionCrafter~\citep{zhu2026motioncrafter}: valid visible 3D GT points are centered and normalized by their mean distance to the origin. Then they are transformed to $\tcam$'s camera coordinate system. The displacement and normal GTs are represented in the same camera coordinate system.

The 2D position and visibility GTs in practice are both defined in the observed $\ttgt$ frame, regardless of $\tcam$. While a 3D point can be projected into any camera view to obtain a 2D position, visibility in that cross-time view is not directly available from the dataset annotations. Defining both targets in the $\ttgt$ view keeps their physical definitions consistent, and we empirically find that they work well as auxiliary losses, consistent with the design in \citet{li2026opend4rt}.


\textbf{Frame and Query Sampling. } 
Since the Wan-VAE uses causal temporal convolutions, the latent at each timestamp doesn't encode future RGB frames and thus can be safely head-truncated. Its encoding is first-frame anchored: the first latent corresponds to the first RGB frame, while each subsequent latent encodes four additional RGB frames. Thus, for $T=4N+1$, we slice the first $T$ query frames and the first $N+1$ latent frames. In practice, we randomly sample the query clip length from $\{21, 25,\ldots,49\}$. A too-short clip will fail to provide adequate context for motion learning and lead to a performance drop.

During training, we sample at most 2048 queries per training clip, with $\tsrc$ independently sampled for each query. Temporally, we enforce $\ttgt = \tcam$ for $40\%$ of the queries, while the rest are randomly sampled. Spatially, $40\%$ is sampled from the foreground and $10\%$ is sampled from the foreground boundaries marked by Sobel-filtered depth maps.

\subsection{Regarding Aria Digital Twin (ADT)}
We include ADT~\citep{pan2023aria} in our training data because recent Geometry Foundation Models (GFMs) such as VGGT~\citep{wang2025vggt} and DA3~\citep{lin2025da3} include ADT for geometry training, while subsequent GFM-based 4D perception methods~\citep{sucar2026vdpm, luo20264rc, lin2026sm4rt, chen2026uniquery4r} build upon these pretrained models. We therefore include ADT to provide comparable geometric supervision during training.

Since ADT does not provide an official train-test split like DR or PO, to ensure a legal training, we strictly \textbf{EXCLUDE} all sequences used by the WorldTrack-ADT benchmark from our training set, as shown in Tab.~\ref{tab:appendix_adt_test_sequence_exclusion}.

And as shown in Tab.~\ref{tab:appendix_additional_tracking}, including ADT in training overall improves tracking performance on the held-out ADT and PO, while performance on the remaining benchmarks stays consistent.

\begin{table*}[htbp]
    \centering
    \caption{\textbf{
        Excluded ADT test sequences during data preprocessing.} One of the 50 evaluation clips contains no visible frame-0 query points, leaving 49 valid evaluation results.
    }
    \label{tab:appendix_adt_test_sequence_exclusion}
    \small

    \resizebox{\textwidth}{!}{%
        \begin{tabular}{@{}l p{0.34\textwidth} rrr@{}}
            \toprule
            \textbf{Scene Group}
            & \textbf{Excluded IDs}
            & \textbf{\#Raw Sequences}
            & \textbf{\#WT Clips}
            & \textbf{\#Valid Clips} \\
            \midrule

            \texttt{Apartment\_release\_clean}
            & 131, 133--136, 138, 140--150
            & 17 & 19 & 19 \\

            \texttt{Apartment\_release\_decoration}
            & 131--140
            & 10 & 11 & 11 \\

            \texttt{Apartment\_release\_decoration\_skeleton}
            & 131--140
            & 10 & 11 & 10 \\

            \texttt{Apartment\_release\_golden\_skeleton}
            & 100
            & 1 & 1 & 1 \\

            \texttt{Apartment\_release\_meal}
            & 131--133, 136, 138--140
            & 7 & 8 & 8 \\

            \midrule
            \textbf{Total}
            & \textbf{WorldTrack-ADT Test Set}
            & \textbf{45}
            & \textbf{50}
            & \textbf{49} \\

            \bottomrule
        \end{tabular}%
    }
\end{table*}
\begin{table*}[htbp]
\centering
\captionsetup{skip=2pt}
\caption{\textbf{Additional quantitative results on world-coordinate tracking.}}
\label{tab:appendix_additional_tracking}
\renewcommand{\arraystretch}{1.1}
\resizebox{\textwidth}{!}{%
    \begin{tabular}{@{}l|cc|cc|cc|cc@{}}
    \toprule
    \multirow{3}{*}{\textbf{Setting}}
    & \multicolumn{2}{c|}{\textbf{ADT}}
    & \multicolumn{2}{c|}{\textbf{DR}}
    & \multicolumn{2}{c|}{\textbf{PO}}
    & \multicolumn{2}{c}{\textbf{PStudio}} \\
    
    & {APD}$\uparrow$ & {EPE}$\downarrow$
    & {APD}$\uparrow$ & {EPE}$\downarrow$
    & {APD}$\uparrow$ & {EPE}$\downarrow$
    & {APD}$\uparrow$ & {EPE}$\downarrow$ \\
    
    & {All / Dyn.}
    & {All / Dyn.}
    & {All / Dyn.}
    & {All / Dyn.}
    & {All / Dyn.}
    & {All / Dyn.}
    & {All}
    & {All} \\
    \midrule

    Ours-\textbf{II}
    & \textbf{91.93} / \textbf{86.82}
    & \textbf{0.112} / 0.160
    & \textbf{88.18} / 92.52
    & \textbf{0.130} / \textbf{0.090}
    & \textbf{78.11} / \textbf{84.92}
    & \textbf{0.196} / \textbf{0.138}
    & 82.11
    & 0.149 \\

    - w/o ADT
    & 74.69 / 84.79
    & 0.261 / \textbf{0.148}
    & 87.50 / \textbf{92.57}
    & 0.136 / \textbf{0.090}
    & 76.07 / 83.60
    & 0.214 / 0.149
    & \textbf{82.38}
    & \textbf{0.147} \\
    
    \bottomrule
    \end{tabular}%
}
\end{table*}

\section{Training Objectives}
\label{sec:appendix_losses}
For each query, our method jointly predicts its 3D position, confidence, 2D position, visibility, 3D displacement, and surface normal.
We reference the general loss design of D4RT~\citep{zhang2026d4rt} and open-sourced repositories~\citep{jiang2026d4rtpytorch,li2026opend4rt}.

For each loss type
$k\in\{3D,\mathrm{conf},2D,\mathrm{vis},\mathrm{disp},\mathrm{normal}\}$,
we use $m_i^k\in\{0,1\}$ to indicate whether query $i$ has valid GT supervision for that loss.
For a per-query scalar $a_i$, we define the masked average as
\begin{equation}
    \left\langle a_i \right\rangle_{m^k}
    =
    \frac{\sum_{i=1}^{N_q}m_i^k a_i}
         {\sum_{i=1}^{N_q}m_i^k+\epsilon},
\end{equation}
where $N_q$ is the number of queries and $\epsilon>0$ prevents division by zero.
Each loss below is averaged using its corresponding validity mask.

The complete training objective is
\begin{equation}
\label{eq:appendix_loss}
    \mathcal{L}
    ={}
    \lambda_{3D}\mathcal{L}_{3D}
    +
    \lambda_{\mathrm{conf}}\mathcal{L}_{\mathrm{conf}}
    +
    \lambda_{2D}\mathcal{L}_{2D}
    +
    \lambda_{\mathrm{vis}}\mathcal{L}_{\mathrm{vis}}
    +
    \lambda_{\mathrm{disp}}\mathcal{L}_{\mathrm{disp}}
    +
    \lambda_{\mathrm{normal}}\mathcal{L}_{\mathrm{normal}}.
\end{equation}
We use $\lambda_{3D}=1.0$, $\lambda_{\mathrm{conf}}=0.2$, $\lambda_{2D}=0.1$, $\lambda_{\mathrm{vis}}=0.1$, $\lambda_{\mathrm{disp}}=0.1$, and $\lambda_{\mathrm{normal}}=0.5$ in all experiments.

\paragraph{3D Position and Confidence.}
Let $\mathbf{P}_i^{\mathrm{GT}}$ and $\hat{\mathbf{P}}_i$ denote the GT and predicted 3D positions of query $i$, respectively.
We compute scene-wise normalization statistics from all valid and visible trajectory points expressed in the camera coordinate system of the first frame.
After mean normalization, the target position at $t_{\mathrm{tgt}}$ is expressed in the camera frame specified by $t_{\mathrm{cam}}$.

We independently normalize the predicted and GT positions by their respective mean absolute depths over valid queries, denoted by $\hat{s}$ and $s^{\mathrm{GT}}$.
We then apply the signed logarithmic transform
$\phi(x)=\operatorname{sign}(x)\log(1+|x|)$
element-wise to reduce the influence of distant points.
The per-query 3D error is
\begin{equation}
    \ell_{3D,i}
    =
    \left\|
        \phi\left(
            \frac{\hat{\mathbf{P}}_i}{\hat{s}}
        \right)
        -
        \phi\left(
            \frac{\mathbf{P}_i^{\mathrm{GT}}}{s^{\mathrm{GT}}}
        \right)
    \right\|_1.
\end{equation}

Given the raw logit $\hat{c}_i$, we parameterize the confidence as
$c_i=1+\exp(\hat{c}_i)$.
The 3D position and confidence losses are
\begin{equation}
\begin{aligned}
    \mathcal{L}_{3D}
    &=
    \left\langle
        c_i\ell_{3D,i}
    \right\rangle_{m^{3D}},
    \\
    \mathcal{L}_{\mathrm{conf}}
    &=
    \left\langle
        -\log c_i
    \right\rangle_{m^{3D}}.
\end{aligned}
\end{equation}
The confidence-weighted 3D error and logarithmic regularizer jointly learn query-wise confidence while preventing trivial collapse.

\paragraph{2D Position and Visibility.}
Let $\mathbf{p}_i^{\mathrm{GT}}$ and $\hat{\mathbf{p}}_i$ denote the GT and predicted 2D positions in image-normalized coordinates $(x/W,y/H)$.
We apply an element-wise Huber loss with unit transition point:
\begin{equation}
    \mathcal{L}_{2D}
    =
    \left\langle
        \sum_{d\in\{x,y\}}
        \operatorname{Huber}_{1}
        \left(
            \hat{p}_{i,d}
            -
            p_{i,d}^{\mathrm{GT}}
        \right)
    \right\rangle_{m^{2D}}.
\end{equation}

Let $v_i^{\mathrm{GT}}\in\{0,1\}$ denote the GT visibility and $\hat{v}_i$ the predicted visibility logit.
We supervise visibility using binary cross-entropy:
\begin{equation}
    \mathcal{L}_{\mathrm{vis}}
    =
    \left\langle
        \operatorname{BCE}
        \left(
            \hat{v}_i,
            v_i^{\mathrm{GT}}
        \right)
    \right\rangle_{m^{\mathrm{vis}}}.
\end{equation}
Here, $v_i^{\mathrm{GT}}=1$ only when the point is physically visible, has positive depth, and lies inside the target image.

\paragraph{3D Displacement.}
The displacement target describes motion from $t_{\mathrm{src}}$ to $t_{\mathrm{tgt}}$ in the camera frame specified by $t_{\mathrm{cam}}$.
Let $\Delta\mathbf{P}_i^{\mathrm{GT}}$ and $\Delta\hat{\mathbf{P}}_i$ denote the GT and predicted displacements, respectively.
We use
\begin{equation}
    \mathcal{L}_{\mathrm{disp}}
    =
    \left\langle
        \left\|
            \Delta\hat{\mathbf{P}}_i
            -
            \Delta\mathbf{P}_i^{\mathrm{GT}}
        \right\|_1
    \right\rangle_{m^{\mathrm{disp}}}.
\end{equation}

\paragraph{Surface Normal.}
We normalize the predicted and GT normals to unit length and supervise their angular discrepancy using cosine distance:
\begin{equation}
    \mathcal{L}_{\mathrm{normal}}
    =
    \left\langle
        1
        -
        \left(
            \frac{\hat{\mathbf{n}}_i}
                 {\|\hat{\mathbf{n}}_i\|_2}
        \right)^{\!\top}
        \left(
            \frac{\mathbf{n}_i^{\mathrm{GT}}}
                 {\|\mathbf{n}_i^{\mathrm{GT}}\|_2}
        \right)
    \right\rangle_{m^{\mathrm{normal}}}.
\end{equation}
The normals are expressed in the camera frame specified by $t_{\mathrm{cam}}$.

\section{Evaluation Protocols}
\subsection{World-coordinate Tracking}
\label{sec:appendix_world_coord_tracking}
\textbf{Motion Threshold.}
\label{sec:appendix_motion_thres}
ADT, DR and PO contain both static and dynamic tracks, whereas PS contains only dynamic tracks. To reliably distinguish dynamic points from static ones for evaluation, we need to select an appropriate motion threshold that suppresses numerical noise in static GT tracks while preserving dynamic tracks with small motions.

We measure the motion of each GT track by accumulating adjacent-frame 3D displacements over 64 frames in meters. As shown in Tab.~\ref{tab:wt-threshold-static}, increasing the threshold from $5\times10^{-3} \to 1\times10^{-2}$ changes the average static ratio only marginally from $56.82\% \to 57.29\%$, indicating that $5\times10^{-3}$ is adequate for removing the majority of the numerical noise. 

In contrast, increasing the threshold from $5\times10^{-3} \to 5\times10^{-2}$ may lead to noticeably more wrongly classified static tracks. The static ratio on ADT shows $67.23\% \to 74.30\%$, a non-negligible $7.07\%$, while on PStudio, despite being fully dynamic, increases to $2.00\%$.

We therefore use $5\times10^{-3}$ as the dynamic-motion threshold throughout our world-coordinate tracking evaluation.

\begin{table*}[htbp]
    \centering
    \captionsetup{skip=4pt}
    \caption{Percentage of frame-0 visible tracks classified as static under different motion thresholds.}
    \label{tab:wt-threshold-static}
    \small
    \begin{tabular}{lccc}
        \toprule
        Track / \%  & $5\times10^{-3}$ m & $1\times10^{-2}$ m & $5\times10^{-2}$ m \\
        \midrule
        ADT     & 67.23 & 68.45 & 74.30 \\
        DR      & 71.81 & 72.33 & 73.62 \\
        PO      & 47.07 & 47.25 & 48.49 \\
        PStudio &  0.00 &  0.00 &  2.00 \\
        \midrule
        Average & 56.82 & 57.29 & 59.34 \\
        \bottomrule
    \end{tabular}
\end{table*}

\textbf{Metric Evaluation.} We follow the protocol of MonST3R~\citep{zhang2025monst3r} using the single global median scale alignment to the GT tracks in the frame-0 world coordinate over the 64-frame tracks. 

Let the predicted tracks be
$\hat{\mathbf{X}}^{\mathrm{world}}\in\mathbb{R}^{T\times N\times 3}$,
where $\hat{\mathbf{X}}^{\mathrm{world}}_{t,n}\in\mathbb{R}^{3}$ denotes the 3D position of the $n$-th track at frame $t$.

The global scale factor $s$ is computed as:
\begin{equation}
    s =
    \frac{
        \operatorname*{median}_{t,n}
        \left\|\mathbf{X}^{\mathrm{GT,world}}_{t,n}\right\|_2
    }{
        \operatorname*{median}_{t,n}
        \left\|\hat{\mathbf{X}}^{\mathrm{world}}_{t,n}\right\|_2
    }.
\end{equation}

The per-point error and End-Point Error (EPE) are computed as:
\begin{equation}
    \begin{aligned}
        e_{t,n}
        &= \left\|
        s\hat{\mathbf{X}}^{\mathrm{world}}_{t,n}
        - \mathbf{X}^{\mathrm{GT,world}}_{t,n}
        \right\|_2, \\
        \mathrm{EPE}
        &= \operatorname*{mean}_{t,n} e_{t,n}.
    \end{aligned}
\end{equation}

Average Percent of Points within Delta (APD) measures the average percentage of points whose errors fall below multiple distance thresholds.
Specifically, given thresholds
$\Delta=\{0.1,\,0.3,\,0.5,\,1.0\}$\,m,
we compute:
\begin{equation}
    \mathrm{APD}
    = 100\times \frac{1}{|\Delta|}
    \sum_{\delta\in \Delta}
    \operatorname*{mean}_{t,n}
    \mathbb{I}\!\left[e_{t,n}<\delta\right],
\end{equation}
where $\mathbb{I}[*]$ is the indicator function.

\textbf{Setups.}
For our method, to satisfy the VAE temporal constraint, we infer 65 frames and discard the last frame before aligning to and evaluating on GT, yielding $T=64$. We keep the query frame in its raw resolution and set the encoder frame to have the same aspect ratio with resolution bounded within $480\times832$.

For SpatialTrackerV2 (STv2)~\citep{xiao2025spatialtrackerv2}, we choose their Type I - Monocular RGB model, i.e., RGB inputs go through its own Front model, which predicts the initial depth and cameras, and Offline model, which jointly refines the point tracks and camera poses, without GT intrinsics or extrinsics. 

For OpenD4RT~\citep{li2026opend4rt}, we choose their `OpenD4RT\_48CLIP\_9Mix\_NoCropAUG' checkpoint, which can infer at most 48 frames at a time. 
For the frames beyond 48, following its published anchor-clip evaluation code, we fix the first frame in every at-most-48-frame input clip and shift the remaining frames to include each target frame.

For the remaining methods, we directly follow their official inference guidelines.

\subsection{Video Depth Estimation}
\label{sec:appendix_video_depth}
\textbf{Metric Evaluation. } We follow the protocol of MonST3R~\citep{zhang2025monst3r} by applying dataset-specific validity ranges when computing depth metrics: \\
For Sintel, evaluation is applied to pixels with GT depth $0 < d_{\mathrm{GT}} < 70\,\mathrm{m}$. After aligning the predicted depth to the GT, predicted depths are additionally clipped to a maximum of $70\,\mathrm{m}$ before metric computation. For Bonn, evaluation is also applied to pixels with GT depth $0 < d_{\mathrm{GT}} < 70\,\mathrm{m}$, but we do not clip the aligned predictions. For KITTI, no maximum-depth threshold is applied, and all valid GT depth pixels are used for evaluation.

We distinguish two types of alignment. For merging chunks of multi-chunk methods, we globally optimize several (scale, shift) pairs over overlapped frames while fixing the first chunk to have (scale=$1$, shift=$0$). 
Specifically, we predict $K$ overlapping chunks of video depth maps. Let
$\mathcal{T}_k$ denote the global frames for which chunk $k$ actually predicts
depth, and let $\hat{D}^{(k)}_{t,p}$ be its prediction at frame $t$ and pixel
$p$. We align each chunk using a scale-shift pair $(s_k,b_k)$:
\begin{equation}
  \tilde{D}^{(k)}_{t,p}=s_k\hat{D}^{(k)}_{t,p}+b_k.
\end{equation}

For adjacent chunks, let
$\mathcal{O}_k=\mathcal{T}_k\cap\mathcal{T}_{k+1}$ denote their overlapped
predicted frame indices. Let
$\mathcal{V}_{k,t}$ denote the pixels at frame $t$ whose depth predictions
are valid in both chunks. We globally estimate the alignments by:
\begin{equation}
  \{s_k^*,b_k^*\}_{k=1}^{K} = \min_{\{s_k,b_k\}}
  \frac{1}{K-1}\sum_{k=1}^{K-1}
  \frac{
      \displaystyle\sum_{t\in\mathcal{O}_k}
      \sum_{p\in\mathcal{V}_{k,t}}
      \left|\tilde{D}^{(k)}_{t,p}
      -\tilde{D}^{(k+1)}_{t,p}\right|
  }{
      \displaystyle\sum_{t\in\mathcal{O}_k}
      |\mathcal{V}_{k,t}|
  },
  \; \text{s.t.}\;s_1^*=s_1=1,\; b_1^*=b_1=0.
\end{equation}

For prediction-GT alignment, we resize the complete depth predictions $\doubletilde{{D}}_{t,p}$ to the GT size using bicubic interpolation. We separately fit a single (scale, shift) pair over pixels valid in GT. Then we compute $\mathrm{AbsRel}$ and $\delta_{<1.25}$:

\begin{align}
  s^\ast,b^\ast = \arg\min_{s,b} &\sum_{(t,p)\in \mathcal{V}} \left| s\doubletilde{D}_{t,p} + b - D_{t,p}^{\mathrm{GT}} \right|, 
  \mathrm{where}\; \mathcal{V} = \left\{(t,p)\ \middle|\ D_{t,p}^{\mathrm{GT}}\text{ is valid}\right\},\\
  D_{t,p}^{\mathrm{pred}} &= s^\ast \doubletilde{D}_{t,p}+b^\ast,\\
  \operatorname{AbsRel} &= \frac{1}{|\mathcal{V}|} \sum_{(t,p)\in \mathcal{V}} \frac{|D_{t,p}^{\mathrm{pred}} - D_{t,p}^{\mathrm{GT}}|}{D_{t,p}^{\mathrm{GT}}},\\
  \delta_{<1.25} &= 100 \times \frac{1}{| \mathcal{V} |} \sum_{(t,p)\in \mathcal{V}} \mathbb{I}\!\left[ \max\!\left( \frac{D_{t,p}^{\mathrm{pred}}}{D_{t,p}^{\mathrm{GT}}}, \frac{D_{t,p}^{\mathrm{GT}}}{D_{t,p}^{\mathrm{pred}}} \right)<1.25 \right].
\end{align}

For each benchmark, scene-level metrics are averaged, with the number of valid evaluation pixels in each scene used as the weight, rather than a direct averaging across scenes.

\textbf{Setups.} For MonST3R, VGGT, DepthAnythingv3, MapAnything and 4RC, we infer all required frames in a single run. 
For Spann3R, Any4D, V-DPM and OpenD4RT, we infer in multiple chunks of 48 frames with an overlap of 24 frames, i.e., $|\mathcal{T}_k|=48,\; |\mathcal{O}_k| = 24$.

Our method also uses chunked inference, with $|\mathcal{T}_k|=49$ and $|\mathcal{O}_k| = 24$ to maintain a comparable inference setting.
For a few Sintel videos that are shorter than the chunk length, we pad the sequence with the repeated last frame until its length reaches the next valid length of $4N+1$, as required by the Wan-VAE. The padded frames are used only for encoding and are discarded before aligning to and evaluating on GT.

Our method naturally supports high-resolution depth estimation. This can be achieved simply by querying on high-resolution query frames.
However, for a fair comparison in Tab.~\ref{tab:video_depth}, we use query frames with a maximum side length of 512 pixels, following a resolution setting comparable to other methods.
The raw query frames are resized while preserving their aspect ratio, and the corresponding encoder frames are resized to the encoder input resolution with the same aspect ratio.

\subsection{Camera Pose Estimation}
\label{sec:appendix_camera_pose}
\textbf{Metric Evaluation. }
For aligning to GT cameras, we estimate a single scene-level $\mathrm{Sim}(3)$ transformation from the corresponding estimated and GT camera centers at all valid frames using Umeyama alignment~\citep{umeyama1991umeyama}, and apply it to the estimated camera-to-world poses. We compute the trajectory metrics using the \texttt{evo} library while following the evaluation protocol of MonST3R~\citep{zhang2025monst3r}.

We report the root-mean-square Absolute Trajectory Error (ATE), Relative Pose Error in translation (\(\mathrm{RPE_T}\)), and Relative Pose Error in rotation (\(\mathrm{RPE_R}\)), measured in meters, meters, and degrees, respectively. Each metric is first computed per scene and then averaged over the per-scene values with equal weight.

Let $\hat{\mathbf T}_t,\mathbf T_t^{\mathrm{GT}}\in SE(3)$ denote the aligned estimated and GT camera-to-world poses at frame $t$, respectively. For each valid adjacent frame pair $(t,t+1)$, we compute
\begin{equation}
\Delta\hat{\mathbf T}_t
=
\hat{\mathbf T}_t^{-1}\hat{\mathbf T}_{t+1},
\qquad
\Delta\mathbf T_t^{\mathrm{GT}}
=
\left(\mathbf T_t^{\mathrm{GT}}\right)^{-1}
\mathbf T_{t+1}^{\mathrm{GT}},
\end{equation}
and define the relative-pose error transformation as
\begin{equation}
\mathbf E_t
=
\left(\Delta\mathbf T_t^{\mathrm{GT}}\right)^{-1}
\Delta\hat{\mathbf T}_t
=
\begin{bmatrix}
\mathbf R_t^{E} & \mathbf t_t^{E}\\
\mathbf 0^\top & 1
\end{bmatrix}.
\end{equation}

Let $\mathcal P$ denote the set of originally adjacent frame pairs $(t,t+1)$ for which both GT poses are valid. The relative translation and rotation errors are:
\begin{equation}
\mathrm{RPE_T}
=
\sqrt{
\frac{1}{|\mathcal P|}
\sum_{t\in\mathcal P}
\left\|\mathbf t_t^{E}\right\|_2^2
},
\qquad
\mathrm{RPE_R}
=
\sqrt{
\frac{1}{|\mathcal P|}
\sum_{t\in\mathcal P}
\theta\!\left(\mathbf R_t^{E}\right)^2
},
\end{equation}
where
\begin{equation}
\theta(\mathbf R)
= \frac{180}{\pi}
\cos^{-1}\!\left(
\operatorname{clamp}
\left(
\frac{\operatorname{tr}(\mathbf R)-1}{2},
-1,1
\right)
\right)
\end{equation}
is the rotation angle in degrees.   

Invalid frames are not skipped to form additional relative-pose pairs. GT alignment is used exclusively for evaluation.

\textbf{Inference Setup.}
Our method estimates the camera trajectory in three stages: cross-camera relative pose estimation, within-chunk pose graph optimization, and global chunk alignment and fusion.

\textit{\textbf{Cross-camera relative pose estimation.}}
We divide the input video into overlapping temporal chunks and encode each chunk once to obtain a shared scene representation $\mathbf F_b$. Using chunk-local frame indices, we construct sliding three-frame windows
\(
\mathcal W_s=\{s,s+1,s+2\}
\)
within each chunk $b$ and sparsely sample queries $(u_k,v_k)$ from the first frame $s$ of each window.

Using the query formulation introduced in Sec.~\ref{sec:preliminary}, we fix both the source and target times to $s$ and vary only the camera time:
\begin{equation}
\hat{\mathbf X}_{k,i}^{3D,(s)}
\equiv
\hat{\mathbf X}^{3D}(u_k,v_k,s,s,i)
=
\mathcal H_{3D}\!\left(
    \mathcal D\!\left(
        \mathbf q(u_k,v_k,s,s,i),\mathbf F_b
    \right)
\right),
\qquad i\in\mathcal W_s,
\label{eq:camera_cross_view_query}
\end{equation}
where the arguments of $\mathbf q$ follow the order
$(u,v,t_{\mathrm{src}},t_{\mathrm{tgt}},t_{\mathrm{cam}})$,
and $\mathcal H_{3D}$ denotes the 3D-position component of the prediction head. Since $t_{\mathrm{src}}=t_{\mathrm{tgt}}=s$, different values of \(i\) represent the same instantaneous geometry in different camera coordinate systems rather than tracking the point across time.

For each camera pair $i<j$ in $\mathcal W_s$, we score each correspondence by the minimum of its two predicted confidences and retain the top $50\%$ of valid correspondences. We then estimate the rigid transformation from camera $j$ to camera $i$ using an unweighted Kabsch fit:
\begin{equation}
\left(
\hat{\mathbf R}_{ij}^{(s)},
\hat{\mathbf t}_{ij}^{(s)}
\right)
=
\arg\min_{\mathbf R\in SO(3),\,\mathbf t\in\mathbb R^3}
\sum_{k\in\mathcal K_{ij}^{(s)}}
\left\|
\hat{\mathbf X}_{k,i}^{3D,(s)}
-
\left(
\mathbf R\hat{\mathbf X}_{k,j}^{3D,(s)}
+\mathbf t
\right)
\right\|_2^2,
\label{eq:camera_kabsch}
\end{equation}
where $\mathcal K_{ij}^{(s)}$ denotes the retained correspondence set obtained from the query window with source frame $s$. We write the resulting relative transformation as
$\hat{\mathbf T}_{ij}^{(s)}\in SE(3)$.

\textit{\textbf{Intra-chunk pose graph optimization.}}
We construct a camera pose graph by adding an edge for every camera pair within each window $\mathcal W_s$. Measurements of the same camera pair obtained from different query windows are retained as separate constraints.

Let $\mathbf T_i\in SE(3)$ denote the chunk-local camera-to-world pose of camera $i$, mapping points from camera $i$ to the reference coordinate system of chunk $b$:
\begin{equation}
\overline{\mathbf X}_{\mathrm{ref}}
=
\mathbf T_i\overline{\mathbf X}_{\mathrm{cam},i},
\qquad
\overline{\mathbf X}
=
\begin{bmatrix}
\mathbf X^\top & 1
\end{bmatrix}^{\top}.
\label{eq:camera_pose_convention}
\end{equation}
Under this left-multiplication convention, the transformation from camera $j$ to camera $i$ is
$\mathbf T_i^{-1}\mathbf T_j$. Therefore, each relative-pose measurement obtained from query window $s$ provides the constraint:
\begin{equation}
\hat{\mathbf T}_{ij}^{(s)}
\simeq
\mathbf T_i^{-1}\mathbf T_j,
\qquad
\mathbf E_{ij}^{(s)}
=
\left(\hat{\mathbf T}_{ij}^{(s)}\right)^{-1}
\mathbf T_i^{-1}\mathbf T_j.
\label{eq:camera_pose_graph_constraint}
\end{equation}

We fix the first camera pose to $\mathbf T_0=\mathbf I$, initialize the remaining poses by traversing the pose graph, and jointly refine them through robust nonlinear least squares while keeping $\mathbf T_0$ fixed. Specifically, we apply a soft-$L_1$ loss with equal weights to the axis--angle rotation and translation residuals extracted from each $\mathbf E_{ij}^{(s)}$.

\textit{\textbf{Global chunk alignment and trajectory fusion. }}
Camera trajectories inferred independently from different chunks are expressed in separate reference frames and may differ in scale. 
Let $\mathbf R_{b,t}$ and $\mathbf c_{b,t}$ denote the camera-to-reference orientation and camera center of frame $t$ estimated within chunk $b$. We assign each chunk $b$ a $\mathrm{Sim}(3)$ transformation
\(
\mathbf S_b=(a_b,\mathbf Q_b,\mathbf d_b)
\),
which maps its local reference frame to the global reference frame.

The globally aligned orientation and camera center are:
\begin{equation}
\widetilde{\mathbf R}_{b,t}
=
\mathbf Q_b\mathbf R_{b,t},
\qquad
\widetilde{\mathbf c}_{b,t}
=
a_b\mathbf Q_b\mathbf c_{b,t}
+\mathbf d_b.
\label{eq:camera_chunk_sim3}
\end{equation}

We initialize these transformations sequentially by estimating pairwise $\mathrm{Sim}(3)$ alignments from the orientations and camera centers of overlapping frames. We then fix the first chunk transformation and jointly optimize the remaining chunk transformations using orientation and center consistency over all overlapping chunk pairs.

Finally, let $\mathcal B_t$ denote the set of chunks containing frame $t$. We fuse their predictions by averaging the aligned camera centers and projecting the sum of the aligned rotation matrices onto $SO(3)$:
\begin{equation}
\mathbf c_t
=
\frac{1}{|\mathcal B_t|}
\sum_{b\in\mathcal B_t}
\widetilde{\mathbf c}_{b,t},
\qquad
\mathbf R_t
=
\operatorname{Proj}_{SO(3)}
\left(
\sum_{b\in\mathcal B_t}
\widetilde{\mathbf R}_{b,t}
\right).
\label{eq:camera_trajectory_fusion}
\end{equation}
This produces a single camera trajectory covering the complete input sequence.

\textit{\textbf{Implementation. }}
We use 49-frame chunks with an overlap of 34 frames. The final chunk is shifted backward when necessary to cover the end of the sequence. Within each chunk, the query windows $\mathcal W_s$ advance with stride one. Image-plane queries are sampled on an 8-pixel grid.

For the query-based OpenD4RT, we adopt the same inference implementation for fair comparison.

\end{document}